\pdfoutput=1

\documentclass[11pt]{article}

\usepackage[preprint]{acl}

\usepackage{times}
\usepackage{latexsym}

\usepackage[T1]{fontenc}

\usepackage[utf8]{inputenc}

\usepackage{microtype}

\usepackage{inconsolata}

\usepackage{graphicx}

\definecolor{todo}{rgb}{1,0,0}

\newcommand{\ignore}[1]{}
\definecolor{ik}{rgb}{0,0,1}

\newcommand{\ik}[1]{}

\definecolor{dz}{rgb}{0.7,0.2,0.9}

\definecolor{editred}{RGB}{220,0,0}

\definecolor{ar}{rgb}{0.7,0.2,0.9}

\usepackage{enumitem}

\usepackage{svg}
\usepackage{booktabs,longtable,tabularx,pifont,hyperref, makecell}
\usepackage[most]{tcolorbox}
\usepackage{array}
\usepackage{cuted}
\usepackage{xcolor}
\usepackage{soul}

\tcbset{
  promptpanelbase/.style={
    enhanced,
    unbreakable,
    boxrule=0.6pt,
    arc=2mm,
    outer arc=2mm,
    frame style={line cap=butt, line join=round},
    left=3mm,right=3mm,top=3mm,bottom=3mm,
    before skip=6pt, after skip=6pt,
    title={#1},
    fonttitle=\bfseries,
    coltitle=white,
  }
}

\newtcolorbox{systempanel}[1]{%
  promptpanelbase={#1},
  breakable,
  colframe=blue!55!black,
  colback=blue!3,
  colbacktitle=blue!60!black
}

\newtcolorbox{userpanel}[1]{%
  promptpanelbase={#1},
  colframe=green!45!black,
  colback=green!3,
  colbacktitle=green!50!black
}

\newtcolorbox{examplepanel}[2][]{%
  promptpanelbase={#2},
  colframe=purple!55!black,
  colback=purple!3,
  colbacktitle=purple!65!black,
  left=2mm,
  right=2mm,
  top=1mm,
  bottom=1mm,
  #1
}

\newtcolorbox{wideexamplebox}[1]{%
  promptpanelbase={#1},
  colframe=purple!55!black,
  colback=purple!3,
  colbacktitle=purple!65!black,
  width=\textwidth,
  left=2mm,
  right=2mm,
  top=1mm,
  bottom=1mm,
}

\newenvironment{wideexamplefloat}[2][!t]
{%
  \begin{figure*}[#1]
  \centering
  \begin{minipage}{\textwidth}
  \begin{examplepanel}{#2}
}
{%
  \end{examplepanel}
  \end{minipage}
  \end{figure*}
}

\colorlet{reviewredbg}{red!22}
\sethlcolor{reviewredbg}

\newcommand{\reviewhl}[1]{\hl{#1}}
\newcommand{\reviewnote}[1]{\textcolor{red!70!black}{\,[#1]}}

\newcommand{\feeduphl}[1]{{\sethlcolor{green!20}\hl{#1}}}
\newcommand{\feedbackhl}[1]{{\sethlcolor{yellow!30}\hl{#1}}}
\newcommand{\feedforwardhl}[1]{{\sethlcolor{cyan!20}\hl{#1}}}

\newtcolorbox{templatepanel}[2][]{%
  promptpanelbase={#2},
  breakable,
  colframe=orange!60!black,
  colback=orange!3,
  colbacktitle=orange!65!black,
  float*,
  floatplacement=tbp,
  width=\textwidth,
  #1
}

\newcommand{\Yes}{\ding{51}}
\newcommand{\No}{\ding{55}}

\usepackage{pgfplots}
\pgfplotsset{compat=1.18}
\usepackage{multirow}
\usepackage{subcaption}

\usepackage{xspace}
\newcommand{\nameDataset}{Exposía\xspace}
\newcommand{\expose}{exposé\xspace}

\title{\nameDataset: Teaching and Assessment of Academic Writing Skills for Research Project Proposals and Peer Feedback}
\author{
  \textbf{Dennis Zyska\textsuperscript{1}},
  \textbf{Alla Rozovskaya\textsuperscript{2}},
  \textbf{Ilia Kuznetsov\textsuperscript{1}},
  \textbf{Iryna Gurevych\textsuperscript{1}}
\\
  \textsuperscript{1}Ubiquitous Knowledge Processing Lab (UKP Lab), Department of Computer Science \\and Hessian Center for AI (hessian.AI), Technical University of Darmstadt\\
 \textsuperscript{2}Department of Computer Science at Queens College, \\City University of New York (CUNY) 
\\
  \small{
    \textsuperscript{1}\href{www.ukp.tu-darmstadt.de}{www.ukp.tu-darmstadt.de}
  }
}

\usepackage{adjustbox}

\begin{document}
\maketitle
\begin{abstract}
We present \nameDataset, the first public dataset\footnote{\url{https://tudatalib.ulb.tu-darmstadt.de/handle/tudatalib/4978}} that connects writing and feedback in higher education, enabling research on educationally grounded computational approaches to teaching and evaluating academic writing. \nameDataset includes student research project proposals and peer and instructor feedback consisting of comments and free-text reviews. 
The dataset was collected in the ``Introduction to Scientific Work'' course of the Computer Science program that focuses on teaching \emph{academic writing skills} and \emph{providing peer feedback on academic writing}. \nameDataset reflects the multi-stage nature of the academic writing process that includes drafting, receiving feedback, and revising the writing based on the feedback received. Both the project proposals and peer feedback are accompanied by human assessment scores based on a fine-grained, pedagogically-grounded schema for writing and feedback assessment that we develop.

We use \nameDataset to benchmark state-of-the-art large language models (LLMs) on two tasks: automated scoring of (1) the proposals and (2) the student reviews.
We find that the two tasks are best served by different LLMs. Furthermore, closed-source models consistently outperform open-weight models, motivating further research on improving the performance of open-weight models preferred in classroom settings. Finally, we establish that a prompting strategy that scores multiple aspects of the writing together is the most effective, paving the way for more effective classroom deployment of modern LLMs.
\end{abstract}

\section{Introduction}
\label{sec:introduction}
\begin{figure}[ht]
    \centering
    \includegraphics[width=.45\textwidth]{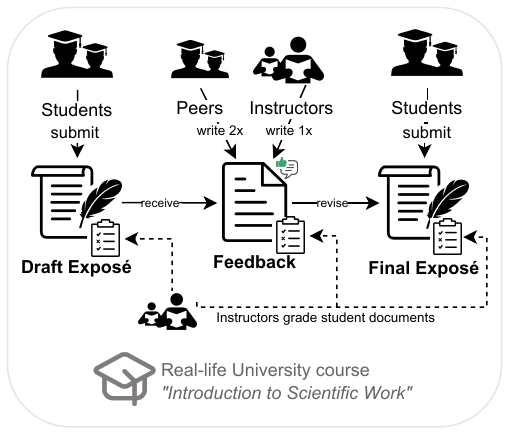}
    \caption{Overview of \nameDataset. 
    Students submit a \emph{draft exposé}, receive \emph{feedback} from instructors and peers in the form of \emph{comments} and free-text \emph{reviews}, revise and submit a \emph{final exposé}. Instructors grade draft and final \expose{}s \emph{and} student feedback.} 
    \label{fig:eyecatcher}
    \vspace{-0.7cm}

\end{figure}

Writing a research project proposal or \expose{}\footnote{The term \expose{} is used in many universities to refer to a \emph{research project proposal}, a structured document that outlines a planned research project. We will use \emph{research project proposal}, \emph{(research project) \expose{}} and \expose{} interchangeably.} is a key challenge in academic education and represents a pressing concern across the academic community~\cite{Muneer2020}. It is also an essential skill required for conducting research and for achieving academic success~\cite{kivunja2016,GRECH201896}. 
However, many students lack the necessary competencies to develop a research proposal and its core components, including identifying a research question, formulating appropriate research methodology, and defining hypotheses and methods to solve the problem~\cite{rastri2023}.

At the same time, higher education faces another pressing challenge: the peer review crisis~\cite{lu-etal-2025-identifying}. Peer feedback is a powerful pedagogical tool that improves the writing performance of both authors and reviewers~\cite{Wu2012,jiang2024}. Yet, students are rarely taught this skill in a classroom. Notably, research shows \emph{a strong link} between academic writing skills and the ability to provide effective feedback~\cite{Huisman2018},  underscoring the need for instructional approaches that \emph{simultaneously support research project proposal writing and the development of effective peer review skills.}

Teaching students how to write a good \expose{} and how to provide high-quality feedback is very challenging. \emph{Formative assessment}, which is used during the learning process, is a fundamental component of effective teaching~\cite{Black1998, Gass2012}. However, human evaluation of \expose{}s and peer feedback requires substantial effort~\cite{Grabau1995}. This applies to \emph{holistic scoring of \expose{}s and feedback}, \emph{fine-grained aspect-based assessment}, and especially to \emph{the provision of textual feedback in the form of comments and reviews}. These limitations necessitate the development of \emph{computational approaches} -- not only for automated scoring of writing and feedback quality, but also for generating detailed text-level feedback on academic writing that evaluates and supports learning. In addition to the lack of appropriate benchmarks both for holistic and fine-grained scoring, providing textual feedback requires modeling complex cross-document relationships between proposal drafts, feedback, and resulting revisions. Curating such data is a time-consuming process. Finally, collecting data in a live course comes with ethical and legal challenges (see \emph{Limitations}). As a result, existing resources for computational writing assistance and assessment are fragmented across subtasks such as automated essay scoring~\cite{Ke2019}, peer feedback analysis~\cite{staudinger-etal-2024-analysis}, or revision modeling~\cite{ruan-etal-2024-large}.

We present \nameDataset, the first corpus that integrates drafting, feedback, revision, and fine-grained \expose{} and feedback assessment scores within a single dataset (Figure~\ref{fig:eyecatcher}). \nameDataset is \emph{continuously} collected in a newly developed university Computer Science course. The course focuses on teaching students (1) \emph{how to write a research project \expose{}} and (2) \emph{how to write peer feedback}. While this is an undergraduate Computer Science course, the skills it develops generalize across disciplines (Section~\ref{sec:dataset}) and are equally applicable to graduate students and other early-career researchers.

\begin{figure*}[ht]
    \centering 
    \includegraphics[width=.92\textwidth]{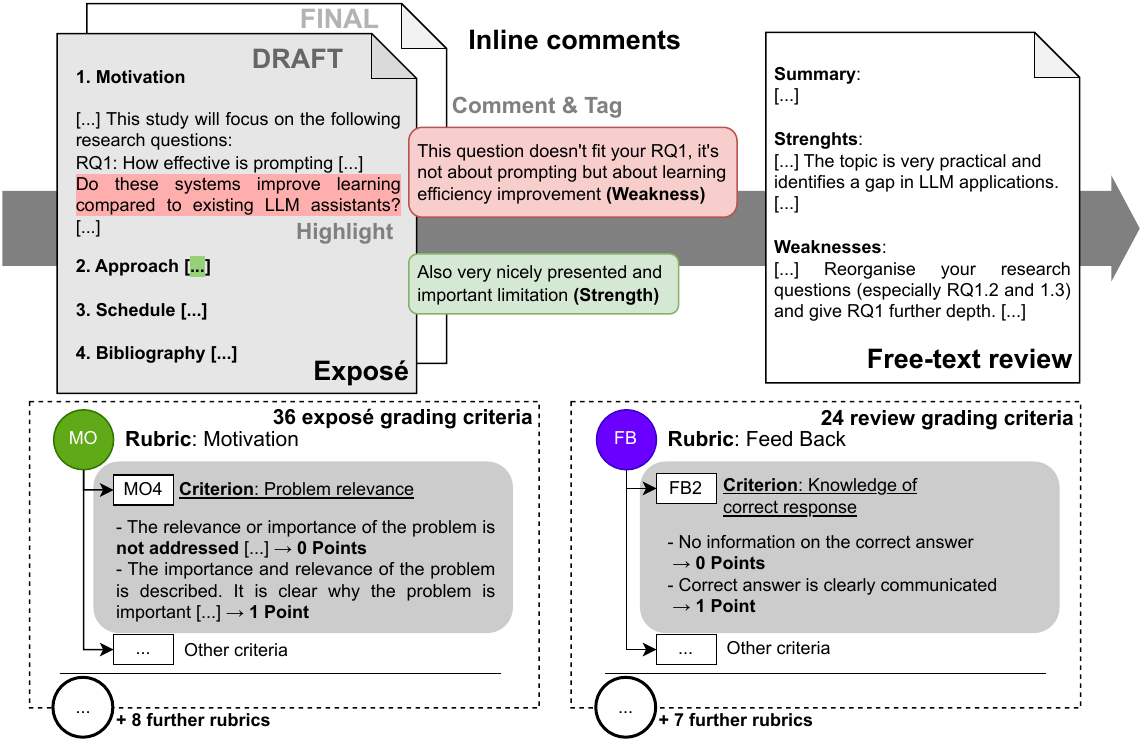} \caption{Example instance from \nameDataset. Top: A student \textbf{draft expos\'e} (top-left) is reviewed by an instructor or a student peer and receives \textbf{feedback}. The feedback consists of (1) passage-anchored \textbf{inline comments} (shown in the top middle in red and green), each associated with a  \textbf{tag} (e.g., Weakness, Strength);  and (2) a free-text \textbf{review} (top-right). Bottom: Expos\'es and reviews are graded by instructors and receive scores on multiple aspects (\textbf{criteria}) organized into high-level topics (\textbf{rubrics}). The \textbf{exposé grading criteria} and the \textbf{review grading criteria} are shown in the bottom left and bottom right of the figure, respectively.}
    \label{fig:motivating-example} 
    \vspace{-3mm}
\end{figure*}

\textbf{\nameDataset} captures a complete research-proposal writing workflow. As illustrated in Figure~\ref{fig:eyecatcher}, each student first produces an expos\'e \textbf{draft}. The draft receives (i) \textbf{peer feedback} from fellow students and (ii) \textbf{instructor feedback}, both provided as (a) \textbf{inline comments} anchored to draft passages and (b) a \textbf{free-text review}. The student revises the draft based on the feedback to produce the \textbf{final} \expose{} version. Figure~\ref{fig:motivating-example} (top) details each component.

We further design fine-grained grading criteria that target the key aspects of \expose{} and feedback writing (Figure~\ref{fig:motivating-example}, bottom). These criteria are grounded in established research on \expose{} writing and feedback assessment and are tailored to the academic disciplinary context of \nameDataset. The exposé drafts, the final \expose{}s, and the student feedback are graded by course instructors, who provide scores on each of the criteria.

To the best of our knowledge, \nameDataset is the first dataset, collected in an educational setting, that features this unique class of cross-document relations, \emph{enabling new applications and tasks} for teaching the skills for writing research proposals and providing effective feedback (see Section~\ref{sec:discussion}). We further establish baseline models with LLMs for two tasks in this space: \textbf{ expos\'e scoring} and \textbf{feedback scoring}, based on the fine-grained grading schema for writing and feedback assessment.

\vspace{-2mm}
\paragraph{Contributions}

The main contribution of this work is the introduction of a novel problem definition for teaching and assessment of \expose{} and feedback writing skills in an educational setting. The result is the first \emph{unique} public dataset of student research \expose{}s and feedback, both accompanied by human assessment scores, based on the fine-grained grading schema that we develop. We present in-depth analyses of the dataset, and models for the corresponding tasks of \expose{} and feedback scoring, establishing baselines that the community can build on.

\section{Related Work}
\label{sec:related-work}
We focus on \expose{} and feedback writing skills, which are distinct from other (academic) writing skills. Below, we discuss related research directions and the gaps in existing datasets. 

\noindent \textbf{Automated essay scoring (AES)} AES~\cite{Page1966} is an established NLP task that focuses on writing assessment.
However, AES benchmarks consist of persuasive essays in general-domain language, manually annotated with a holistic score, complemented by scores for a small set of assessment criteria (e.g., content, organization, language) \cite{mathias-bhattacharyya-2018-asap,gaudeau-2025-beyond, li-ng-2024-icle, Xue2021, Crossley2023, Ishikawa2023}. Importantly, these benchmarks do not include human feedback, revisions, or feedback assessment scores. 

\noindent \textbf{Peer feedback} Because \nameDataset{} links each \expose{} draft with feedback text and assessment scores, it is related to scholarly peer-review datasets~\cite{peerread,nlpeer}. However, these datasets provide only coarse-grained annotations. Crucially, they serve fundamentally different purposes, as they focus on scientific communication in a professional decision-making context, whereas \nameDataset{} is situated in an educational environment with a focus on learning and assessment. 

\noindent \textbf{Connecting writing, feedback, and revisions} A small number of resources connect student writing with peer or instructor feedback and/or revision~\cite{zhang-etal-2017-corpus, Kashefi2022,pilan-etal-2020-dataset,rietsche-etal-2022-specificity}, enabling analyses of how feedback relates to revision outcomes. 
However, existing datasets still focus on persuasive essays in general-domain language. Furthermore, they lack fine-grained annotations that allow for evaluating the impact of feedback on revisions and writing outcomes, and that are provided in \nameDataset{}: (i) passage-anchored inline comments tied to draft spans, (ii) role-separated \emph{peer and instructor} feedback, and (iii) criterion-level assessment scores on the text to be assessed and the feedback. 

\section{Course Setting and Study Protocol}

\label{subsec:course-protocol}
\nameDataset was collected across two iterations of the newly introduced
Bachelor-level Computer Science course \emph{Introduction to Scientific Work}:
the first cohort in the winter term 2024/25 and the second cohort in the winter
term 2025/26.\footnote{The institutional IRB has approved the related proposal.}

\noindent \textbf{Course structure} In the first cohort, the students wrote an exposé on one of three topics, covering various areas of Computer Science and Artificial Intelligence (AI):\footnote{Topics were evenly assigned based on preference voting.} \textit{Holographic AI Assistant} (Human-Computer Interaction), \textit{Interactive LLM-based Teaching Agents} (Natural Language Processing), or \textit{Generative AI in Healthcare} (Data Mining). In the second cohort, the students worked on three new topics:
\textit{Generative AI for Creative Collaboration} (Natural Language
Processing), \textit{Interactive Service Agents} (Human-Computer Interaction),
or \textit{Ethical and Transparent AI Systems} (Trustworthy AI). In addition,
the second cohort introduced a German-language track alongside the
English-language track.

In the beginning of the course,\footnote{The syllabus is provided in Appendix~\ref{appendix:syllabus}.} alongside the lectures, the students received guidelines for writing an \expose{} (Appendix~\ref{appendix:writing_guidelines}) and the \expose{} grading criteria (Section~\ref{subsec:expose-criteria}). Each student produced an exposé draft. This was followed by a peer-review phase, where each student wrote double-blind reviews for two \expose{}s written by their peers, selected uniformly at random. The students were provided with review grading criteria (Section~\ref{subsec:review-criteria}). The instructional staff also provided a review for each \expose{}. This was followed by a revision phase to produce the final \expose{}, based on the feedback. Both the drafts and final versions, and the student reviews were graded by the instructional staff. The final grade was calculated as follows: draft exposé 30\%, peer reviews 30\%, and final exposé 40\%.\footnote{Grades for each component are derived from criterion-based scores (see Section~\ref{sec:dataset} and Appendix~\ref{appendix:subsec:expose-revision-stats:points}).} Reviewing was facilitated through the open-access CARE platform~\cite{zyska-etal-2023-care,Zyska2025}.

In the first cohort, out of 193 students enrolled in the course,\footnote{Enrollment was limited because the course was offered for the first time as a pilot.} 159 completed all requirements. The second cohort was
substantially larger: 445 students submitted a draft exposé, and 436 completed
all requirements. The \nameDataset dataset refers to the \emph{open data} that participants agreed to contribute for public release. All 12 instructors in the first cohort and 23 instructors in the second cohort
consented to share their materials. Among students who completed the course, 34\% in the first cohort and 39\% in the second cohort consented
to public release. Consent was requested independently of grading and had no effect on course assessment.

\textbf{Grade distributions of consenting and non-consenting students}
Since Exposía only includes data from consenting students, we compare the grade distribution of consenting students with that of the full course population. We find that in both cohorts, the distributions between consenting students and the full course population are similar, and the average grade is the same, indicating that the released subset is not primarily composed of high-achieving students. 

\noindent \textbf{Continuous large-scale data collection}
\nameDataset is collected continuously and will be released incrementally.
The \emph{first release}, covering the winter term 2024/25, contains 55 \expose{}s,
306 reviews, and 2,253 inline comments. The first cohort serves as the basis for the analyses and experimental results reported in the main paper. 
The \emph{second release}, covering the winter term 2025/26, was finalized shortly before camera-ready and adds 171 \expose{}s, 767 reviews, and 4,773 inline comments. Appendix~\ref{appendix:second-cohort} documents the second cohort
and provides a cross-cohort analysis. The second cohort substantially extends the resource, by providing a larger,
bilingual collection, criterion-level scores accompanied by instructor justifications, and information about whether human assessments were conducted with or without LLM support.
These additions enable the scaling of peer-feedback processes through AI support, as well as future research on cross-lingual writing assessment and the effects of AI-supported criterion-based assessment. The course runs annually, and the dataset will continue
to grow, following similar collection efforts~\cite{nlpeer}.

\textbf{Consistency protocol for future cohorts} We develop a protocol to ensure  a consistent data collection and assessment procedure across changes in staff and \expose{} topics. Specifically, all instructors use the same exposé and review criteria documented in the next section and Appendix~\ref{appendix:dataset}. Instructors are introduced to the criteria before grading begins, and discuss difficult cases in calibration sessions. For future cohorts, a recorded calibration video and representative examples are provided to support onboarding of new instructors. 

\section{The \nameDataset Dataset}
\label{sec:dataset}
\subsection{\nameDataset{} Dataset Overview}
\vspace{-1mm}
\label{subsec:dataset-overview}
Figure~\ref{fig:motivating-example} illustrates two main components of the \nameDataset{} dataset: \emph{\expose{}s} and \emph{feedback}.

\noindent \textbf{Exposés} Each exposé is available in two versions: a \emph{draft} and a revised \emph{final version} produced after receiving feedback. 
The exposés follow a standard Computer Science research proposal template with the following sections: (1) \emph{Motivation}; (2) \emph{Approach} divided into three parts -- \emph{Related Work}, \emph{Theoretical Framework}, and \emph{Methodology}; (3) a \emph{Schedule} outlining the week-by-week plan; and (4) a \emph{Bibliography} (Appendix~\ref{appendix:writing_guidelines}).

\noindent \textbf{Feedback: inline comments and reviews}  Each draft \expose{} received feedback from instructors and other students. Each feedback consists of \emph{inline comments} and a free-text \emph{review}. An \emph{inline comment} consists of a span-level \emph{highlight} on the draft text and a free-text \emph{comment} connected to it (Figure~\ref{fig:motivating-example}). Each inline comment is also associated with a \emph{tag} from a fixed taxonomy
 $\mathcal{Y} =
\{\textit{Strength}, \textit{Weakness}, \textit{Highlight}, \textit{Other}\}$. A reviewer first adds inline comments. Then the reviewer uses them as the basis for composing a \emph{review}, a document-level free-text feedback on the draft. Appendix~\ref{appendix:dataset} provides more detail.

\noindent \textbf{Grading of the \expose{}s and student reviews} The \expose{}s and student reviews are graded by instructors on 36 and 24 assessment criteria, respectively. The same criteria and scoring scales
apply to both cohorts. The second cohort additionally includes
written justifications for most criterion scores. The criteria are designed to reflect traits that could impact an overall quality of the \expose{} and review, respectively, and were developed in consultation with the instructors of the course. The criteria are grounded in established research on \expose{} and feedback writing, as detailed below (see Appendix~\ref{appendix:criteria} for annotation guidelines). Before the course was run, the criteria were assessed by the instructors and their definitions were adjusted for clarity (for more detail, see Appendix~\ref{appendix:iaa-setup}).

\subsection{Exposé Assessment Criteria and Rubrics}
\vspace{-1mm}
\label{subsec:expose-criteria}
Both the draft and the final \expose{} are graded independently by an instructional staff member and include 36 per-criterion \emph{assessment scores} (see Table~\ref{tab:expose_rubric} for guidelines on grading). Each criterion is rated on an ordinal scale: [0,1], [0,1,2], or [0,-1,-2].

\noindent \textbf{Designing \expose{} assessment criteria and rubrics} The criteria were derived using 18 \emph{cross-disciplinary guidance documents} for writing thesis exposés from major universities in Germany and Austria (Appendix~\ref{appendix:expose-collection}), supplemented by observations from instructors in our institution, based on over 100 \expose{}s previously submitted in the Computer Science Department. The 36 criteria cover different aspects of writing a scientific
proposal, with a focus on \emph{content} assessment. Importantly, the criteria are \emph{topic-independent} (rather than specific to the research topic of the proposal). This approach is in line with aspect-based assessment used in automated essay scoring ~\cite{li-ng-2024-icle} and focuses assessment on \emph{transferable proposal-writing skills} (e.g., motivation and methodology). Accordingly, the same 36 criteria are applied uniformly across the proposal topics.

Guided by prior research in rubric construction and fairness-oriented assessment \cite{JONSSON2007130,Mayring_2000, Kuckartz2023}, we organize the criteria into nine high-level categories referred to as \emph{rubrics}\footnote{Rubrics are an effective assessment tool and are constructed to ensure \emph{consistency in evaluation across a diverse student body and multiple evaluators}~\cite{niu2012rubrics}. See Appendix~\ref{appendix:expose-criteria} for more details.} (Table~\ref{tab:iia-expose-review-criteria-rubrics-only}), by grouping together criteria that assess related aspects of the \expose{} writing.

\subsection{Review Assessment Criteria and Rubrics}
\vspace{-1mm}
\label{subsec:review-criteria}
 Each \emph{student review} was independently graded by an instructional staff member on 24 criteria (see Table~\ref{tab:review_rubric} for the guidelines on grading) organized into eight rubrics (Table~\ref{tab:iia-expose-review-criteria-rubrics-only}), with each criterion rated on an ordinal scale of [0,1], [0,1,2], or [-1,1]. 

\noindent \textbf{Designing review assessment criteria and rubrics} The review assessment criteria are grounded in established pedagogical theory and, similar to the \expose{} criteria, focus on feedback skills that are not tied to the proposal topic (e.g., argumentation, actionable suggestions).  The criteria operationalize key educational principles of effective feedback and are primarily informed by the feedback model in~\newcite{hattie2007power}.

Drawing on \newcite{narciss2008feedback}, we integrate the components of feedback as criteria within the eight rubrics. We further incorporate insights from~\newcite{darcy-etal-2024-aries},
which analyzes revisions in response to peer reviews in scientific writing
and shows that \emph{actionable comments} offering concrete suggestions are
more likely to lead to substantive text revisions. As such, our rubrics emphasize argumentation, accuracy, and clear suggestions for improvement as indicators of pedagogically effective feedback~\cite{Shute2008}.
Appendix~\ref{appendix:review-criteria} provides more detail. 

\section{\nameDataset{} Dataset Analysis}
\label{sec:dataset_analysis}

\subsection{\nameDataset{} Dataset Statistics}
\label{sec:dataset_statistics}

\begin{table}[t]
    \centering
    \begingroup
    \setlength{\tabcolsep}{6pt}
    \begin{tabular}{l r r r}
    \toprule
    \textbf{Topic} & \textbf{Exposés} & \textbf{IC} & \textbf{Reviews} \\
    \midrule
    T1 & 19 & 798 & 43 \\
    T2 & 18 & 785 & 44 \\
    T3 & 18 & 670 & 35 \\
    \multicolumn{3}{l}{\emph{Orphaned reviews\textsuperscript{*}}} & 184 \\
    \midrule
    Total \textbf{$\Sigma$} & \textbf{55} & \textbf{2{,}253} & \textbf{306} \\
    \bottomrule
    \end{tabular}
    \endgroup
    \caption{Statistics on the \nameDataset{} corpus for the first cohort
(winter term 2024/25). Topics: T1=\emph{The Holographic AI Assistant}, T2=\emph{Interactive LLM-based teaching agents}, T3=\emph{Generative AI in Healthcare}. IC=Inline Comments. \textsuperscript{*}Orphaned reviews  are reviews for which no corresponding exposé exists.} 
    \label{tab:dataset-overview}
      \vspace{-4mm}

\end{table}

The dataset statistics for the first cohort are in Table~\ref{tab:dataset-overview}.
The 55 exposés are evenly distributed across three CS/AI topics: Human--Computer Interaction, Natural Language Processing, and Data Mining / AI in Healthcare, with 19, 18, and 18 exposés, respectively. This provides topical variation while keeping the task, rubric, and feedback protocol fixed.
As the dataset comprises only materials for which consent was granted, some exposés lack the full set of three reviews (Table~\ref{tab:appendix:reviews_per_expose}), and there exist 184 \emph{orphaned} reviews because the corresponding exposés could not be made public.
All released exposés nevertheless include the complete draft--feedback--revision structure with at least one review. The orphaned reviews further contribute student-written feedback and remain fully usable for review-level analyses and tasks, including review scoring, review-quality assessment, and inter-annotator agreement analysis on student feedback.
The number of student reviews is 110, including 76 that are orphaned. Each draft \expose{} includes on average 41 inline comments. 
\emph{Weakness} is the most frequent tag (59\%) of all comments, followed by \emph{Strength} (18\%), \emph{Other} (14\%) and \emph{Highlight} (9\%).

\begin{figure}[t]
  \centering
  \includegraphics[width=.48\textwidth]
  {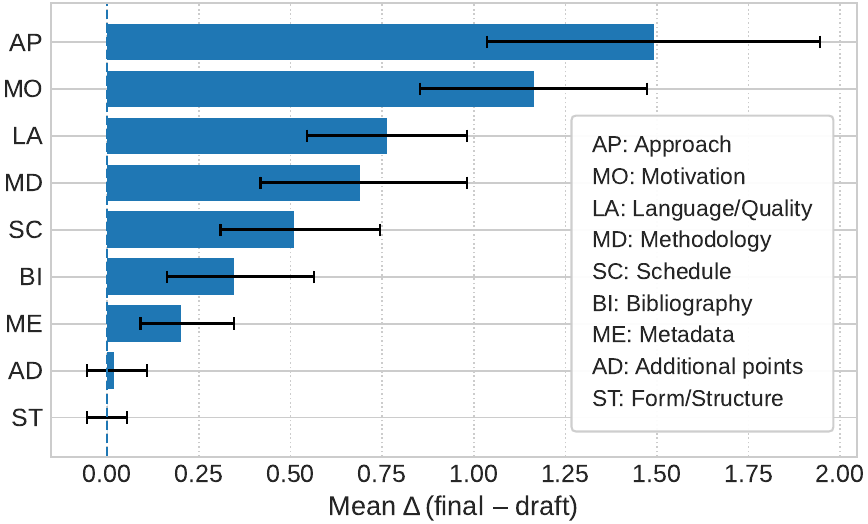}
  \caption{Per-rubric score improvements from draft to final \expose{} in the first cohort. Each bar shows the mean change in score. Error bars indicate nonparametric bootstrap 95\% confidence intervals for each item.}
  \label{fig:rubric_deltas}
  \vspace{-4mm}
\end{figure}

\noindent \textbf{Exposé revision analysis} We analyze the impact of \emph{feedback} by comparing how the final \expose{} changes relative to the draft: final versions increase in length from 528 to 649 words, on average.  
We also analyze revisions with respect to \expose{} criterion-based scores, averaged by rubric. The largest gains occur in \emph{Approach}, \emph{Motivation}, \emph{Language/Quality} and \emph{Methodology} (Figure~\ref{fig:rubric_deltas}). The most substantial revisions focus on conceptual and argumentative elements (see Appendix~\ref{appendix:subsec:expose-revision-stats}).

\noindent \textbf{Review analysis} We compare reviews written by instructors vs. students (Appendix~\ref{appendix:subsec:review-stats}): instructor reviews are almost twice as short as student reviews, but have higher lexical density and higher lexical diversity. Appendix~\ref{appendix:subsec:review-stats} provides more analyses. 

\subsection{Inter-Annotator Agreement}

\label{sec:iaa}
\begin{table}[ht]
\centering
\begingroup
\setlength{\tabcolsep}{4pt} 
\begin{tabularx}{\columnwidth}{>{\raggedright\arraybackslash}X r r r}
\toprule
\textbf{Rubric} & \textbf{QWA} & $\boldsymbol{\alpha}$ & $\mathbf{r}$ \\
\midrule
\multicolumn{4}{@{}l}{\textbf{Exposé rubrics}} \\
\midrule
Language/Quality (\#2)   & 0.85 & 0.10 & 0.19 \\
Metadata (\#2)           & 0.83 & 0.00 & 0.31 \\
Form/Structure (\#2)     & 0.97 & 0.33 & 0.34 \\
Motivation (\#7)         & 0.76 & 0.11 & 0.17 \\
Approach (\#8)           & 0.77 & 0.22 & 0.26 \\
Methodology (\#7)        & 0.62 & 0.18 & 0.22 \\
Schedule (\#4)           & 0.93 & 0.36 & 0.44 \\
Bibliography (\#2)       & 0.78 & 0.19 & 0.42 \\
Addt'l points (\#2)      & 0.98 & 0.32 & 0.34 \\
\midrule
All (avg.)               & 0.79 & 0.20 & 0.27 \\
\midrule
\multicolumn{4}{@{}l}{\textbf{Review rubrics}} \\
\midrule
Structure/Clarity (\#2)  & 0.86 & 0.27 & 0.38 \\
Language (\#2)           & 0.93 & 0.09 & 0.17 \\
Feed Up (\#2)            & 0.81 & 0.43 & 0.43 \\
Feed Back (\#6)          & 0.86 & 0.04 & 0.08 \\
Feed Forward (\#2)       & 0.78 & 0.02 & 0.04 \\
Content Quality (\#3)    & 0.94 & 0.22 & 0.24 \\
Errors (\#5)             & 0.93 & 0.10 & 0.12 \\
Addt'l points (\#2)      & 0.98 & -0.01 & \textemdash \\
\midrule
All (avg.)               & 0.89 & 0.13 & 0.18 \\
\bottomrule
\end{tabularx}
\endgroup
\caption{IAA on \expose{} and review grading in the first cohort. Quadratic weighted agreement (QWA), Krippendorff's ($\boldsymbol{\alpha}$), and Pearson $r$. $r$ for Addt'l points (review) is undefined due to zero variance and excluded from the average. Result for each rubric is averaged over all criteria in the rubric. The number of criteria per rubric is next to rubric name. Rubric definitions are in Tables~\ref{tab:expose_rubric} and~\ref{tab:review_rubric}.}
\label{tab:iia-expose-review-criteria-rubrics-only}
  \vspace{-4mm}
\end{table}

We compute inter-annotator agreement (IAA) on \expose{} and review criterion-based scores: each draft \expose{} and student review are independently graded by two instructors on 36 \expose{} and 24 review criteria, respectively (using two raters to compute IAA is common practice in similar settings such as AES~\cite{li-ng-2024-icle}), yielding 3{,}960 exposé assessment scores and 6{,}720 review assessment scores. This yields a substantial number of criterion-level judgments for reliability analysis.

\noindent \textbf{Rater training and annotation effort} During the semester, each draft \expose{} and student review were graded by an instructor only once; after the course, each item was graded again to calculate IAA. This annotation effort is substantial: grading one exposé requires assessing it on 36 criteria and providing feedback (inline comments and a review), taking 1–2 hours in total. Grading one review involves assessing it on 24 criteria and takes 20--40 minutes (see Appendix~\ref{subsec:annotation-effort} for more detail).
Calibration was conducted before and during grading: Group~1 jointly graded initial items, attended weekly calibration sessions, and discussed difficult cases throughout the semester, while Group~2 received the same guidelines, a recorded calibration video, and an in-person discussion with experienced Group~1 instructors.
See Appendix~\ref{appendix:iaa-setup} for more detail on rater training.

\noindent \textbf{IAA metrics} We report Krippendorff’s $\alpha$~\cite{krippendorff04},\footnote{Krippendorff’s $\alpha$ can be used in a setup such as ours, where we are comparing multiple raters where raters also belong to one of the two groups.}  Pearson correlation coefficient $r$, and raw quadratic weighted percent agreement  (QWA).
We include all three metrics because $\alpha$ and $r$ support comparison with prior work in automated writing assessment, while QWA is more informative under prevalence skew.
We have several criteria with skewed distributions where the majority label accounts for over 90\% cases (Tables~\ref{tab:iia-expose-criteria} and ~\ref{tab:iia-review_criteria}). In this case, both $\alpha$ and $r$ can be misleading~\cite{Jeni2013-yl, Zhao2022-xu}, and QWA is the most informative metric. Appendix~\ref{appendix:iaa-human} and~\ref{subsec:failure} provide more detail.

\noindent \textbf{IAA on exposé assessment scores} Table~\ref{tab:iia-expose-review-criteria-rubrics-only} (top) summarizes IAA results by \expose{} rubric. Overall agreement yields an average QWA of $0.787$, with $\alpha = 0.198$, $r = 0.270$. The observed correlation levels are comparable to those reported on automated essay scoring \cite{persing-ng-2014-modeling,persing-ng-2013-modeling} and reflect the subjectivity of the task. The rubrics that attain the lowest agreement levels are those related to content (Motivation, Approach, Methodology). Table~\ref{tab:iia-expose-criteria} reports IAA for all 36 criteria. 

\noindent \textbf{IAA on review assessment scores} Table~\ref{tab:iia-expose-review-criteria-rubrics-only} (bottom) summarizes IAA results by review rubric, computed over student reviews.  While QWA is higher for review rubrics than for \expose{}, $\alpha$ and Pearson coefficient are lower, due to the skewed distributions of many criteria (Appendix~\ref{subsec:failure}). For criterion-level agreement and discussion, see Appendix~\ref{appendix:iaa-human} and 
Appendix Table~\ref{tab:iia-review_criteria}. 

\section{Experiments}
\label{sec:baseline}
\subsection{Experimental Setup}
\label{subsec:baseline-setup}

\paragraph{Tasks}
We use \nameDataset{} to benchmark LLM performance on two tasks:
(i) criterion-based exposé assessment and (ii) criterion-based assessment of student reviews. 
We focus on zero-shot prompting to establish reproducible baselines for Exposía. This setting also reflects realistic classroom deployment scenarios, where annotated examples for a new cohort, topic, or criterion may not yet be available.

\begin{figure}[t]
  \centering

  \begin{subfigure}[b]{\linewidth}
    \centering
    \includegraphics[width=\linewidth]{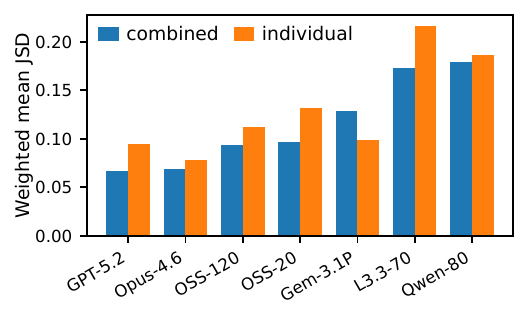}
    \caption{Exposé scoring}
    \vspace{-1mm}
    \label{fig:jsd_expose}
  \end{subfigure}
\vspace{-3mm}
  \medskip

  \begin{subfigure}[b]{\linewidth}
    \centering
    \includegraphics[width=\linewidth]{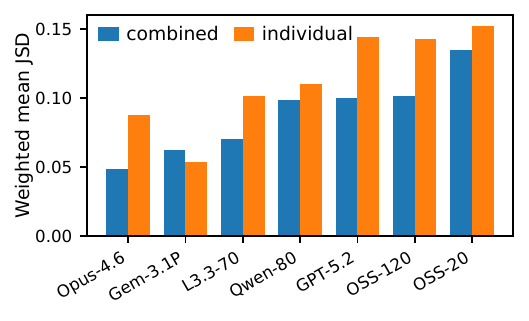}
    \vspace{-6mm}
    \caption{Review scoring}
    \vspace{-2mm}
    \label{fig:jsd_review}
  \end{subfigure}

  \caption{Weighted-mean binned JSD between human and LLM score distributions (\emph{lower is better}), by prompting variant. Models: \textit{Opus-4.6}=Claude Opus 4.6; \textit{Gem-3.1P}=Gemini 3.1 Pro; \textit{GPT-5.2}=GPT 5.2; \textit{OSS-120}=GPT OSS (120B); \textit{OSS-20}=GPT OSS (20B); \textit{L3.3-70}=Llama 3.3 (70B); \textit{Qwen-80}=Qwen 3 (80B).}
\vspace{-4mm}
  \label{fig:jsd_combined}
\end{figure}

\noindent \textbf{LLM prompting} We assess LLM performance in a zero-shot setting across both open-weight and closed-source models (Appendix~\ref{appendix:performance_models}). For both tasks, we use a unified prompting strategy and compare two prompting variants: (i) \emph{single-criterion prompts}, which present and score one criterion at a time, and (ii) \emph{combined prompts}, which list all criteria for the respective task in a single call. Cross-criterion synergy could potentially benefit the model, with a lower execution cost. Appendix~\ref{appendix:prompt_templates} provides more detail about the prompts.

\noindent \textbf{Evaluation} Following studies on criterion-based scoring~\cite{PACK2024100234, huang-wilson-2025-evaluating}, we compare predicted criterion scores to human ratings using the protocol of \citet{elangovan2025beyond} who recommend using the binned Jensen--Shannon divergence (JSD) for capturing human--machine alignment, instead of agreement metrics such as Krippendorff, Pearson, or QWA, which are designed to account for errors in human annotation, but are not appropriate for computing human--machine agreement. For each criterion, we bin items by the median human score and aggregate the results over agreement and disagreement bins. We compute JSD between the empirical human score distribution and the model's predicted score distribution within each bin. We report the mean binned JSD (\emph{lower is better}), aggregated over criteria (we also report QWA between models and human raters for completeness). Results are reported on 55 \expose{} drafts and 110 student reviews. For each \expose{} and review, the corresponding model predicts 31 and 22 scores, respectively,\footnote{5 \expose{} and 2 review criteria are excluded (Appendix~\ref{appendix:experimental-details}).} for a total of 1705 \expose{} and 2420 review criterion scores.

\subsection{Results}
\label{subsec:results}
Figure~\ref{fig:jsd_combined} shows key results on \expose{} and review assessment (averaged over all criteria). Also shown in Table~\ref{tab:jsd-domain-allmodels}, and Table~\ref{tab:qwa-expert-allmodels} tabulates QWA results. We emphasize the JSD results, however, especially since many of our criteria exhibit relatively high human variation (Table~\ref{tab:iia-expose-review-criteria-rubrics-only}), and when the proportion of samples with uncertainty in human-assigned labels is relatively high, machine labels may superficially appear to
have similar or better correlation with the human majority label compared to
the human-to-human correlation. This can create the illusion that labels
from automatic evaluation approximate the human majority label~\cite{elangovan2025beyond}, which is exactly what we observe when evaluating using QWA. Our key findings are as follows:

\noindent \textbf{Combined prompting outperforms single-criterion prompting}
Across tasks and models, combined prompting yields higher alignments with human raters. Combined prompting is also more efficient computationally (Appendix~\ref{appendix:results} and~\ref{appendix:performance}). This is important since student activity is deadline-driven (Figure~\ref{fig:appendix:behaviour:daily_activity}). Thus, \emph{combined prompting is recommended for classroom deployment}. 

\noindent \textbf{Closed-source models outperform open-weight models}
Across tasks, closed-source models tend to achieve \emph{lower} binned JSD than open-weight models, indicating closer alignment with human score distributions. 
At the same time, educational deployments impose strict privacy constraints: students may not consent to their work being transmitted to public-facing model APIs. This creates a practical need for better-aligned, locally deployable open-weight models that support privacy-preserving educational applications.

\noindent \textbf{Different models are better suited for \expose{} and review scoring} We find that Claude exhibits consistently high alignment with humans on both tasks. However, other models are less consistent. For instance, \emph{GPT 5.2} attains the best alignment on \expose{} scoring but is among the poorest models on review scoring, whereas \emph{Llama 3.3} exhibits the opposite behavior. This suggests that different models may be better suited to the two tasks. In Appendix~\ref{appendix:results}, we identify criteria on which LLMs align most closely with human raters.

\noindent \textbf{Model performance is stable on review scoring but not \expose{} scoring} Finally, we observe that all models exhibit similar alignment levels on review scoring, whereas on \expose{} scoring the models exhibit less stability, and the difference between the top-performing model (\emph{GPT 5.2}) and the most poorly-aligned \emph{Qwen 3} is substantial. 

\noindent \textbf{Comparison with zero-shot prompting on automated essay scoring} Recent zero-shot AES results provide a useful reference point. ~\newcite{lee-etal-2024-unleashing} report average quadratic weighted kappa scores of 0.550 and 0.587 on ASAP and TOEFL11, respectively, with Mistral-7B-Instruct. These numbers are not directly comparable to the results on Exposía because the tasks, rubrics, and evaluation metrics differ. Nevertheless, they show that zero-shot LLM-based writing assessment remains challenging.

\subsection{Illustrative Examples from \nameDataset}

Appendix~\ref{appendix:examples} presents two  examples from \nameDataset. The examples illustrate how an expos\'e draft,  reviewer feedback, and a revised \expose{} are connected in the dataset.
We briefly discuss one example that focuses on the \expose{} criterion ``Teaser'' (Research Question). This example shows how the \emph{Teaser} criterion is used to assess whether a proposal states a clear research question. The relevant excerpt from the draft proposal, including an inline comment from an instructor, is below:

\mbox{[...]} \reviewhl{This research addresses the critical gap in AI-driven cancer research by integrating the impact of the tumor microenvironment into data analysis workflows for personalized treatment planning. We will \mbox{[...]} address the limitations of current AI models in incorporating TME data and explore the anticipated improvement of personalized treatment prediction.}\reviewnote{\textsc{Weakness}: The goals remain broad; the main research question is not clearly stated.}\mbox{[...]}

In the draft, the research direction is understandable, but remains too broad because it describes goals and anticipated contributions rather than a distinct research question. This is reflected in the instructor reviews and the corresponding scores assigned on the \emph{Teaser} criterion (1 and 0 for each instructor, for a maximum of 2, Table~\ref{tab:appendix_expose_grading}). We show a review excerpt from the instructor who gave a 1:

\mbox{[...]} \feeduphl{Having an explicit, clearly defined research question gives a lot of structure to the overall expose...  the exact goals and your main focus are not entirely clear to me. } \mbox{[...]} 

The final \expose{} (Appendix~\ref{appendix:examples}) illustrates how the \expose{} was successfully revised, following the feedback, and assessed at 2 on this criterion. 
 
Lastly, we analyze LLM-generated assessment scores for this example. We find that the models disagree in their scores. The best-performing model in Figure~\ref{fig:jsd_combined} (\emph{GPT 5.2}) assigned a score of 0 with the following justification: \feedbackhl{``No explicit research question is stated.''}. Several models interpreted the passage as an implicit research direction: \emph{Claude Opus} assigned a score of 1 and argued that the work was described in broad terms without a clearly articulated research question. Yet other models gave a score of 2. For example, \emph{Llama 3.3} described the research question as clearly formulated. 

Although the instructors disagree on the score, both identified the missing \emph{Teaser}. In contrast, several LLMs, especially open-source ones, failed to detect the absence of a clearly formulated research question. This example highlights the limitations of LLM-based scoring and motivates future work on alternative approaches and improved prompting based on closer analysis of model justifications.

\section{Tasks Enabled by \nameDataset}
\label{sec:discussion}
As discussed in Section~\ref{sec:dataset}, the annotations in \nameDataset{} are intentionally broad and support a wide range of tasks for scientific writing assistance and assessment, beyond those evaluated in this paper. This richness is a key strength of the dataset.\footnote{Publishing a dataset with additional components that enable new tasks is consistent with other resource papers in CL venues~\cite{baumel-etal-2014-query,qin-etal-2018-automatic}.} In particular, the dataset enables research on modeling draft--feedback--revision links across documents, with numerous applications. More broadly, the dataset supports research on the relationships among feedback, revision, and writing development in higher-education settings.
Because the corpus contains paired drafts and final \expose{}s,
criterion-level assessments of both versions, and the corresponding feedback, it enables analyses of textual revisions and draft-to-final score changes with respect to the feedback received. In this paper, we analyze these patterns for the first cohort (Section~\ref{sec:dataset_analysis}), and find that the largest score increases occur for rubrics related to content and argumentation rather than form. Extending this analysis remains future work.

\nameDataset supports more fine-grained tasks such as linking inline comments to specific revisions, predicting revisions from feedback, and modeling feedback properties, associated with criterion-level improvements in revised \expose{}s. Such models can help identify types of feedback that are most effective for improving specific writing dimensions. 
Exposía also enables feedback-generation tasks, where a model produces localized comments or document-level reviews from a draft \expose{} and the assessment criteria. This is an important application because good feedback can improve students' writing and, at the same time, help students improve their own feedback skills~\cite{Wu2012,jiang2024}. We leave feedback-generation experiments to future work because responsible evaluation of generated feedback in educational settings requires a dedicated study of correctness, helpfulness, actionability, and potential harm. Future modeling work can also extend our zero-shot baselines with few-shot prompting, rationale-based prompting, prompt optimization, and fine-tuning.

Finally, automated scoring of exposés supports scalable assessment for research proposal writing, while automated review scoring supports scalable assessment of student ability to provide effective peer feedback (a key practical need given the substantial human grading effort; Appendix~\ref{subsec:annotation-effort}).

All of these applications directly contribute to addressing the two challenges of (1) research proposal writing and (2) providing feedback on proposal writing. \nameDataset is the first dataset that explicitly connects the two target skills.

\section{Conclusion}

\label{sec:conclusion}
We introduce a novel task definition for teaching and assessment of exposé and feedback writing skills,  and present the first public dataset, \nameDataset, of student exposés and feedback, accompanied by human assessment scores based on the fine-grained grading schema we develop.

By integrating drafting, feedback, and revision into a single resource, \nameDataset enables pedagogically grounded research on writing and feedback in higher education and provides a unified framework that allows for a joint study of feedback and \expose{} writing quality. We also outline a range of tasks enabled by \nameDataset. Although the dataset is collected in the Computer Science domain, the task formulation and the assessment framework are general enough to inform the development of models in other areas. We establish baseline performance with state-of-the-art LLMs on \expose{} and review scoring tasks. Our resource, findings and models pave the path for the holistic study of computational support for academic writing and feedback.

\section*{Limitations}
\label{sec:limitations}
This study has several limitations. One limitation is that \nameDataset{} is shaped by the course consent protocol: since only consented materials are included, not all exposés retain the complete set of three reviews.
For the same reason, some of the reviews in \nameDataset may be \emph{orphaned}, if the author of the corresponding exposé did not provide consent (Section~\ref{sec:dataset_statistics}). As such, the consent structure reduces the number of complete draft–review–revision instances in the released data.

Consent-based data release may also introduce self-selection bias, for example if students who are more confident in their writing or less concerned about data sharing are more likely to contribute. To assess whether the released subset primarily reflects high-performing students, we compared the grade distribution of consenting students with that of the full course population. The average grade is the same in both groups, and the grade distributions are similar. Nevertheless, grade similarity cannot rule out other forms of self-selection, such as differences in privacy attitudes or confidence. Consent was requested independently of grading outcomes and had no effect on course assessment.

Although \nameDataset{} remains modest in size compared with some other NLP datasets, it is substantial for the educational domain, particularly given its complexity and the richness of its annotations. We discuss the time investment required to create the \nameDataset{} components in Appendix~\ref{subsec:annotation-effort}.

Another limitation concerns the generalizability of the dataset and the assessment criteria beyond Computer Science and AI, to other disciplinary contexts. We stress that the course is designed to teach research proposal writing and peer feedback skills, which generalize across disciplines and are applicable to early-career researchers in general. Furthermore, our assessment criteria are grounded in educational assessment research that is not specific to Computer Science (Section~\ref{sec:dataset}) and are derived from cross-disciplinary guidance documents (Appendix~\ref{appendix:expose-collection}). However, empirical validation of cross-disciplinary transferability is beyond the scope of this paper: it would require collecting multiple comparable draft--feedback--revision datasets across disciplines and evaluating whether human agreement patterns and model performance remain stable across domains. 
We therefore expect the proposed framework to generalize to other disciplines, although broader empirical validation remains important future work.

Additionally, we leave for future work the question of empirically validating the assessment criteria that we design for \expose{} and review grading. This is because validating the assessment framework would require a large-scale comparative study in which students are assessed using either the criteria developed in this work, or an alternative assessment scheme, and the outcomes of both groups would need to be compared. Such a study would take more than a year to complete and would require the involvement of several researchers. It is therefore outside the scope of the present work. 

Another, related limitation is that \nameDataset{} focuses specifically on research proposal writing, which differs from other (academic) writing tasks in purpose, structure, and style (see Section~\ref{sec:related-work}). Similarly, \expose{} feedback differs from other forms of feedback studied in the NLP community, in particular, peer review for academic publishing. As a result, the assessment criteria and experimental findings reported in our study may not transfer directly to other writing tasks and require further validation. At the same time, we expect some aspects to generalize beyond this setting: several criteria, such as language quality and organization, overlap with other areas, for example, automated essay scoring. In contrast, other dimensions, including motivation, methodology, and approach, are more closely tied to the goals of proposal writing. 

A further limitation is that open-source models do not yet perform on par with the strongest closed-source systems across the tasks considered in this work. This is especially relevant in educational settings, where student consent and privacy constraints may limit the use of external proprietary models. Future work should therefore focus on developing stronger open-source assessment models and privacy-preserving deployment strategies that better fit the practical and ethical constraints of educational data.

Finally, and most importantly, although LLM-based scoring can be attractive for reducing instructional workload, our results should not be interpreted as support for fully automated grading: the models exhibit systematic scoring tendencies and may miss substantive issues that are crucial for guiding revision, and automated grading can incentivize strategic ``gaming'' rather than genuine improvement~\cite{ALQAHTANI20231236}; we therefore view LLMs as decision-support tools, and as common in AES, recommend a human-in-the-loop workflow when using them for assessment.

\section*{Ethical Considerations}
\label{sec:ethical-considerations}

Collecting data in an educational setting is challenging due to the high sensitivity of the data, which raises serious privacy and ethical issues~\cite{guan2023}. The data collection was approved by the University’s Institutional Review Board and Data Protection Office under IRB proposal \textbf{EK~77/2024}.

Data was collected based on informed consent (all instructors and 55 students), following a three-tier structure:  (1) general consent for data processing within the review platform, (2) optional consent for contributing textual and highlight data, and (3) optional consent for contributing interaction data. Participation in the study was strictly voluntary and had no effect on grading for the students. No monetary compensation was provided to students, as participation was part of the course setting; importantly, contributing research data was optional. Instructors were compensated via regular University employment contracts, following institutional pay scales and policies. The compensation was not contingent on participation in the study.

We release \nameDataset{} to enable research on academic writing, peer feedback, and assessment in educational settings. We stress that our findings are not to be interpreted as justification for fully automatic grading: all assessment-oriented applications of AI should be restricted to decision-support roles with human oversight, in line with the limitations outlined above.

\section*{Acknowledgments}
This work has been funded by the German Research Foundation (DFG) as part of the PEER project (grant GU 798/28-1), Hessian.AI Connectom Fund (LLMentor: Expert-AI Coteaching of “Introduction to Scientific Work”) and by the European Union (ERC, InterText, 101054961). Views and opinions expressed are, however, those of the author(s) only and do not necessarily reflect those of the European Union or the European Research Council. Neither the European Union nor the granting authority can be held responsible for them.
We thank Dr. Ekaterina Kochmar (Mohamed bin Zayed University of Artificial Intelligence) and Dr.~Martin Greisel (University of Augsburg) for their valuable feedback and discussions.

\bibliography{custom}

\appendix

\section{Course Syllabus}
\label{appendix:syllabus}
This appendix documents the syllabus of the undergraduate computer
science course \textit{Introduction to Scientific Work}, which
provided the instructional setting for the data collection described
in Section~\ref{subsec:course-protocol}.

The course covers foundational concepts and practical skills
required for conducting scientific research.
Here we present the weekly lectures:

\begin{enumerate}[itemsep=2pt, topsep=2pt, parsep=0pt, partopsep=0pt]
    \item What is science? Course overview and introduction
    \item Empirical methods and research design
    \item Literature review and academic search strategies
    \item Project proposals (exposés); \textit{start writing exposé draft}
    \item Scientific writing 
    \item \LaTeX{} for academic writing
    \item Research Questions and practical exercise on academic literature search
    \item Introduction to peer review
    \item Feedback; \textit{start writing peer feedback}
    \item Generative AI in scientific work
    \item Research ethics I
    \item Research ethics II
    \item Workshop: Software development 
    \item Workshop: Research and project management
\end{enumerate}

\section{Exposé Template and Writing Guidelines}
\label{appendix:writing_guidelines}
Panel \hyperref[panel:template]{\textit{Exposé template}} shows the content of the \LaTeX{} \expose{} template that students received as part of the course and used to create their \expose{}s. 
While the complete instructions were provided in an extensive slide-based lecture (Appendix~\ref{appendix:syllabus}), the excerpt reproduced here is the concise reference sheet embedded directly in the template. 
Because this template guided how students organized and formatted their drafts, we include it to ensure transparency and to document the pedagogical context in which the dataset was created.

\begin{templatepanel}{Exposé template}
\phantomsection\label{panel:template}

\subsection*{1\quad Motivation}

\textbf{Hook} – Introduce the topic and catch the reader’s interest.
\begin{itemize}
    \item \emph{Concrete}: Quote, example, real-world event.
    \item \emph{Helpful}: Directly useful or valuable information.
    \item \emph{Convincing}: Strong argument or empirical fact.
    \item \emph{Shocking}: Unexpected or counterintuitive finding.
    \item \emph{Personal}: Research journey or applied experience.
    \item \emph{Understandable}: Link complex ideas to everyday life.
\end{itemize}

\textbf{Anchor} – Situate the problem concisely.
\begin{itemize}
    \item Discipline, field, or domain.
    \item How the problem has been addressed so far.
    \item Current academic relevance.
    \item Deficits or criticism in existing literature.
\end{itemize}

\textbf{Teaser} – Identify the gap and position your research.

\textbf{Research Questions}
\begin{itemize}
    \item Break the main RQ into meaningful sub-questions.
    \item Clarify terminology; define if needed.
    \item Set clear boundaries and scope.
\end{itemize}

\vspace{0.8em}
\subsection*{2\quad Approach}

\textbf{State of the Art (SOTA)}
\begin{itemize}
    \item Refer to relevant existing work.
    \item Highlight research deficits and criticism.
    \item Combine different traditions or perspectives.
\end{itemize}

\textbf{Theoretical Framework}
\begin{itemize}
    \item Introduce theories used in the study.
    \item Explain key concepts and definitions.
    \item Describe how the theory informs the RQs.
\end{itemize}
\emph{Note:} Theories need not be SOTA but should be connected to it.

\textbf{Methodology}
\begin{itemize}
    \item Specify method (quantitative, qualitative, mixed, source research).
    \item Define target population and sampling.
    \item State data sources and availability.
    \item Clarify agreements or permissions.
    \item Anticipate potential difficulties.
    \item Discuss methodological limitations.
    \item Explain relevance and scope.
    \item Describe variables and their relationships.
    \item Outline practical implementation (software, documents, helpers).
    \item Clarify epistemic goals (knowledge to be gained).
    \item If building on prior work, state progress so far.
\end{itemize}

\vspace{0.8em}
\subsection*{3\quad Schedule}

Provide a chronological work plan with realistic milestones.

\textbf{Tips:}
\begin{itemize}
    \item Break tasks into concrete steps.
    \item Include checkpoints and buffer time.
    \item Use weekly or monthly units depending on project size.
    \item Seek advice from experienced researchers.
\end{itemize}

\textbf{Example Schedule}

\small
\setlength{\tabcolsep}{4pt}
\renewcommand{\arraystretch}{1.2}

\begin{tabularx}{\linewidth}{|c|c|>{\raggedright\arraybackslash}X|>{\raggedright\arraybackslash}X|}
\hline
\textbf{Month} & \textbf{Week} & \textbf{Task} & \textbf{Description} \\ \hline
1 & 1 & Planning & Work plan, coordination \\
1 & 2--3 & Literature Review & Search and initial analysis \\
1 & 4 & Literature Evaluation & Critical review and selection \\ \hline
2 & 5--6 & Theoretical Framework & Develop theory and methodology \\
2 & 7 & Concept Development & Draft interview/ analysis guide \\
2 & 8 & Scenario Planning & Design simulations or experiments \\ \hline
3 & 9--10 & Recruitment & Acquire participants \\
3 & 11 & Data Collection & Conduct interviews/ experiments \\
3 & 12 & Data Preparation & Clean and pre-analyze data \\ \hline
4 & 13--14 & Qualitative Analysis & In-depth qualitative analysis \\
4 & 15--16 & Quantitative Analysis & Statistical analysis \\ \hline
5 & 17 & Writing Results & Compose results section \\
5 & 18 & Discussion & Interpret findings \\
5 & 19 & Methodology Review & Finalize methodology \\
5 & 20 & Full Revision & Edit and review all chapters \\ \hline
6 & 21 & Final Editing & Final proofreading, formatting \\
6 & 22--24 & Finalization & Buffer, printing, submission \\ \hline
\end{tabularx}

\vspace{1em}

\textbf{Reference}

[1] Helmut Balzert, Christian Schäfer, Marion Schröder, Uwe Kern, Roman Bendisch, and Klaus Zeppenfeld. \emph{Wissenschaftliches Arbeiten: Ethik, Inhalt und Form wissenschaftlicher Arbeiten, Handwerkszeug, Quellen, Projektmanagement, Präsentation}. W3L-Verlag, Herdecke/Witten, 2011. ISBN: 978-3-937137-59-9.

\end{templatepanel}

\section{The Second Cohort: Winter Term 2025/26}
\label{appendix:second-cohort}
The second \nameDataset{} cohort became available after the paper had completed review and was released shortly before camera-ready. Consequently, the analyses and experiments in the main paper remain based on the first cohort. We nevertheless include the second cohort in the present dataset release and document it here because it constitutes a substantial extension of the resource.

\noindent\textbf{Course and participation}
The course was offered for a second time in the winter term 2025/26, following the same draft--feedback--revision structure, the same \expose{} and review criteria, and the same reviewing workflow as in the first cohort.
The students worked on three new research topics: \emph{Generative AI for Creative Collaboration} (Natural Language Processing), \emph{Interactive Service Agents} (Human--Computer Interaction), and \emph{Ethical and Transparent AI Systems} (Trustworthy AI).
The third topic introduces Trustworthy AI, an area not represented among the three topics in the first cohort.

Of the 445 students who submitted a draft \expose{}, 436 completed all course requirements. Among these students, 39\% consented to the public release of their materials, compared with 34\% in the first cohort. Consent rates were similar across the three topics, ranging from 32\% to 37\%. The resulting release contains 171 \expose{}s, 767 reviews, and 4,773 inline comments.
Table~\ref{tab:dataset-overview-ws2526} reports the distribution across topics using the same definitions as the dataset overview in the main paper.
\begin{table}[t]
    \centering
    \begingroup
    \setlength{\tabcolsep}{4pt}
    \begin{tabular}{lrrr}
        \toprule
        \textbf{Topic}
        & \textbf{Exposés}
        & \textbf{IC}
        & \textbf{Reviews} \\
        \midrule
        T1 & 63 & 1{,}511 & 102 \\
        T2 & 46 & 1{,}276 & 81 \\
        T3 & 46 & 1{,}510 & 86 \\
        \emph{not recorded} & 16 & 476 & 25 \\
        \midrule
        \emph{Orphaned reviews}$^*$ & -- & -- & 473 \\
        \midrule
        \textbf{$\Sigma$}
        & \textbf{171}
        & \textbf{4{,}773}
        & \textbf{767} \\
        \bottomrule
    \end{tabular}
    \endgroup
    \caption{Statistics for the second \nameDataset{} cohort. Topics:
    T1=\emph{Generative AI for Creative Collaboration},
    T2=\emph{Interactive Service Agents}, and
    T3=\emph{Ethical and Transparent AI Systems}. IC=Inline Comments.
    $^*$Orphaned reviews are reviews for which the corresponding \expose{} is not part of the public release.}
    \label{tab:dataset-overview-ws2526}
\end{table}

\noindent\textbf{Bilingual course tracks}
The second release additionally contains German-language \expose{}s and reviews from participants who consented to the public release of their materials. 
The \expose{} and review assessment criteria are introduced in
Sections~\ref{subsec:expose-criteria} and~\ref{subsec:review-criteria} and documented in detail in Appendix~\ref{appendix:expose-criteria}.
The two language versions use the same rubric codes, criterion order, scoring levels, and point ranges.
Both language versions of the \expose{} and review criteria are included in the release, and each record identifies the language in which it was produced.

Of the 171 released \expose{}s, 149 are associated with the English-language track and 22 with the German-language track. The review data comprise 673 English-language reviews and 96 German-language reviews.

\noindent\textbf{LLM-assisted grading}
The second cohort additionally included an LLM-assisted condition for the instructor assessment of draft \expose{}s. 
As described in Section~\ref{sec:dataset}, instructors assess each \expose{} using the criterion-based grading scheme introduced in Section~\ref{subsec:expose-criteria} and documented in detail in Appendix~\ref{appendix:expose-criteria}.

Before an instructor assessed a draft \expose{}, an on-premise Qwen3-14B model processed its \LaTeX{} source together with the corresponding assessment criterion and its scoring descriptors. For each criterion, the model was instructed to assign an integer score and provide a one--two sentence justification grounded in specific evidence from the \expose{}. The model additionally generated draft textual feedback. Criterion-level assessments were generated separately but batched into a single inference call per submission. We used deterministic decoding (temperature~0), and the criterion definitions were matched to the language of the submission.

Instructors were assigned either to an assisted or a control condition. 
In the assisted condition, the model-generated scores, justifications, and draft feedback were shown pre-filled in the grading interface; instructors could accept or modify them before submitting their assessment. 
In the control condition, instructors used the same grading interface, but the model-generated pre-assessment was not shown. The condition was assigned at the instructor level rather than independently for each \expose{}. 
The use of LLM-assisted grading in the course was approved by the university's ethics board under approval EK~56/2025.

In the second cohort, instructors also provided short written justifications for most criterion-level scores. 
These justifications are included in the dataset together with their provenance, distinguishing instructor-written justifications from model-generated suggestions that were accepted verbatim.
For example, for the criterion \emph{Hook existing}, one model-generated justification reads: ``The exposé opens with a compelling statistic: `Only 13\% of germans trust news media which were mainly created by artificial intelligence (AI).' This directly addresses the topic's relevance and establishes a clear problem statement, making it a suitable hook.''

The release contains 84 assisted draft gradings. 
In these assessments, 1,048 of 2,694 recorded justifications were submitted unchanged from the model-generated suggestion, and 2,278 of 2,602 criterion scores (88\%) were identical to the model's proposed score.
In the control condition, where the same model pre-assessment was generated but not shown to instructors, 77\% of the human-assigned scores were identical to the model proposal.

\noindent\textbf{Comparison of the releases}
Table~\ref{tab:cohort-comparison} compares the two released cohorts under consistent counting rules. 
The second release is substantially larger in absolute terms and introduces written criterion-level justifications, which are available for 16,160 of its 19,464 numeric criterion scores (83\%). 

In the first cohort, a dedicated second grading pass was collected after the course specifically to estimate inter-annotator agreement (IAA). 
Since the second cohort used the same assessment criteria and annotation protocol, and the reliability of this protocol had already been evaluated on the first cohort, we did not consider a second grading pass necessary for the purposes of the present release.

The larger cohort did not produce proportionally more feedback per document. Inline comments decreased from 41.0 to 27.9 per released \expose{}, while reviews per \expose{} decreased from 2.22 to 1.72. The composition of the inline feedback also shifted: the proportion labelled as a weakness decreased from 59\% to 50\%, while the proportion labelled as a strength increased from 18\% to 26\%. 

\begin{table}[t]
    \centering
    \begingroup
    \setlength{\tabcolsep}{3pt}
    \begin{tabular}{@{}lrr@{}}
    \toprule
    & \textbf{WS 24/25} & \textbf{WS 25/26} \\
    \midrule
    Exposés & 55 & 171 \\
    Reviews & 306 & 767 \\
    \quad orphaned & 184 & 473 \\
    Inline comments & 2{,}253 & 4{,}773 \\
    \quad per exposé & 41.0 & 27.9 \\
    \quad Weakness & 59\% & 50\% \\
    \quad Strength & 18\% & 26\% \\
    \quad Other & 14\% & 12\% \\
    \quad Highlight & 8\% & 10\% \\
    Annotations & 5{,}765 & 11{,}584 \\
    Comments & 5{,}797 & 11{,}782 \\
    Comment votes & 592 & 105 \\
    Display states$^\dagger$ & -- & 562 \\
    Gradings & 450 & 649 \\
    \quad raw & 538 & 778 \\
    \quad drafts & 110 & 162 \\
    \quad finals & 55 & 162 \\
    \quad reviews & 285 & 325 \\
    Criterion scores & 12{,}779 & 19{,}464 \\
    \quad justified & -- & 16{,}160 \\
    Behaviour records & 586{,}505 & 1{,}991{,}894 \\
    Languages & EN & EN, DE \\
    Second grading pass (IAA) & yes & -- \\
    \bottomrule
    \end{tabular}
    \endgroup
    \caption{Comparison of the two \nameDataset{} cohorts using the same counting procedure. Gradings are deduplicated by author, grader, type, review, and criterion; raw counts include duplicated storage locations. Behaviour records cover both the course and the subsequent assessment. $^\dagger$Display states indicate whether a comment card was expanded or collapsed. EN = English; DE = German.}
    \label{tab:cohort-comparison}
\end{table}

\noindent\textbf{Cross-cohort LLM baseline}
To examine whether the first-cohort baseline transfers to the second cohort, we repeated a reduced version of the automated assessment experiment from Section~\ref{sec:baseline}. We evaluated GPT-5.2 under zero-shot combined prompting, using the same prompt construction and the same 31 \expose{} and 22 review criteria as in the first-cohort experiment. The \expose{} evaluation was restricted to the control condition, whose graders were not shown the LLM pre-assessment. The review evaluation used all graded reviews available in the experimental input, since no LLM-generated criterion scores were displayed during review grading. The English-language rubric descriptions were used in the prompts for both English- and German-language records.
Table~\ref{tab:appendix-cohort-qwa} reports quadratic-weighted agreement between the model and the first human grading group.

\begin{table}[t]
\centering
\begingroup
\setlength{\tabcolsep}{7pt}
\renewcommand{\arraystretch}{1.00}

\begin{tabular}{@{}lrrr@{}}
\toprule
\textbf{Rubric}
& \shortstack{\textbf{WS}\\\textbf{24/25}}
& \shortstack{\textbf{WS}\\\textbf{25/26}}
& $\boldsymbol{\Delta}$ \\
\midrule

\multicolumn{4}{@{}l}{\emph{Exposé drafts}} \\
\multicolumn{4}{@{}l}{%
  \hspace{1em}($n=55$ vs.\ $87$, control condition)
} \\
Language & 0.88 & 0.86 & $-0.02$ \\
Metadata & 0.90 & 0.92 & $+0.02$ \\
Motivation & 0.83 & 0.87 & $+0.04$ \\
Approach & 0.86 & 0.87 & $+0.01$ \\
Methodology & 0.72 & 0.76 & $+0.04$ \\
Schedule & 0.94 & 0.91 & $-0.03$ \\
Bibliography & 0.57 & 0.56 & $-0.01$ \\
\textbf{All criteria}
& \textbf{0.82}
& \textbf{0.83}
& $\mathbf{+0.01}$ \\
\midrule

\multicolumn{4}{@{}l}{%
  \emph{Reviews} ($n=106$ vs.\ $337$)
} \\
Structure and Clarity & 0.84 & 0.88 & $+0.04$ \\
Language & 0.94 & 0.92 & $-0.02$ \\
Feed Up & 0.83 & 0.80 & $-0.03$ \\
Feed Back & 0.88 & 0.85 & $-0.03$ \\
Feed Forward & 0.87 & 0.60 & $-0.27$ \\
Content Quality & 0.91 & 0.90 & $-0.01$ \\
Errors & 0.85 & 0.91 & $+0.06$ \\
\textbf{All criteria}
& \textbf{0.87}
& \textbf{0.85}
& $\mathbf{-0.02}$ \\
\bottomrule
\end{tabular}

\endgroup
\caption{Quadratic-weighted agreement between GPT-5.2 under combined prompting and human criterion scores in WS~2024/25 and WS~2025/26. The second-cohort \expose{} evaluation is restricted to the control condition, whose graders were not shown the LLM pre-assessment. The second cohort contains only one human grading pass, so no within-cohort human--human agreement reference can be computed.}
\label{tab:appendix-cohort-qwa}
\end{table}

The binned Jensen--Shannon divergence analysis from Section~\ref{sec:baseline} cannot be extended to the second cohort. Its agreement and disagreement bins depend on whether at least two independent human raters assigned the same score to an item. Duplicating the single available rating would place every item in the agreement bin by construction and would therefore not approximate the missing quantity.

Overall model--human agreement on \expose{} drafts remains similar across cohorts, increasing from 0.82 to 0.83. Agreement on reviews decreases slightly, from 0.87 to 0.85. The largest criterion-level change occurs for Feed Forward, where agreement decreases from 0.87 to 0.60. This is also one of the criteria for which the mean human-assigned score decreases most strongly in Table~\ref{tab:cohort-scores}. Thus, the score comparison and the independently computed model-agreement comparison both identify Feed Forward as a principal difference between the two cohorts.

\noindent\textbf{Score comparison}
Table~\ref{tab:cohort-scores} reports descriptive per-rubric scores for the two cohorts. 
For the first cohort, we use the in-semester grading pass (Group~1), which is the basis of the first-cohort descriptive score analysis and is directly comparable to the single grading pass available for the second cohort.

Restricting the second-cohort drafts to graders who were not shown the LLM pre-assessment (\emph{control}, $n=78$), the mean total draft score increases from 33.00 to 35.14. When the 84 assisted gradings are included (\emph{all}, $n=162$), the mean increases further to 35.49. 
Mean review scores decrease from 22.33 to 21.27. The largest changes occur for the Feed Up and Feed Forward rubrics. Mean final-\expose{} scores increase from 38.20 to 39.52.

\noindent\textbf{Limits on direct comparison}
The cross-cohort experiment should be interpreted as a transfer check rather than a full replication of the experiments in Section~\ref{sec:baseline}. It evaluates only GPT-5.2 under combined prompting, and the \expose{} evaluation is restricted to the control condition, in which graders were not shown the LLM pre-assessment. Because the second cohort contains only one human grading pass, no second-cohort human--human agreement reference can be computed. For the same reason, the binned Jensen--Shannon divergence analysis from Section~\ref{sec:baseline} cannot be extended to this cohort, since its agreement and disagreement bins require at least two independent human ratings.

More generally, the cohorts should not be pooled without accounting for differences in language, research topics, grader composition, and assessment procedure. The second cohort includes both English- and German-language records and an LLM-assisted draft-grading condition. Because this condition was assigned at the grader level, LLM assistance is confounded with grader identity, preventing a clean estimate of the effect of visible model support. Moreover, the cross-cohort experiment uses English-language rubric descriptions in the prompts for both English- and German-language records and therefore does not isolate language effects.

The first cohort consequently remains the basis for the agreement analyses and model benchmarks in the main paper. The second cohort instead provides a transfer check together with a substantially larger bilingual collection, criterion-level written justifications, and explicit provenance for assessments produced with and without visible LLM support.

\section{Additional Details on the \nameDataset Dataset}
\label{appendix:dataset}
\paragraph{Exposés} Each exposé is available in two versions: a \emph{draft} and a revised \emph{final version} produced after receiving feedback. 

Each version is provided as (i) a compiled PDF file and (ii) the original \LaTeX{} source, including the separate bibliography file (\texttt{.bib}) and all textual content. 

\paragraph{Reviews}  Every \expose{} draft is accompanied by two \emph{student reviews}  written by fellow students taking the course and at least one \emph{instructor review}. Instructional staff included nine instructors and consisted of PhD students, postdoctoral researchers, and teaching assistants holding a Master’s degree (Appendix~\ref{appendix:iaa-setup} (Group~1)). 
The feedback was double-blind among students, but not with respect to the instructional staff. During the revision process, students could identify whether a specific feedback came from an expert or a student.

Each review is released as (i) an HTML document that preserves the original formatting used on the review platform and (ii) a plain-text rendering. For all reviews, we provide basic metadata including a synthetic reviewer name, reviewer role, document size, the associated exposé (if available), and an anonymized identifier.

For all reviews for which the author provided additional consent for behavioral logging, we release the full edit history of the review text as stored by the platform, capturing the successive revisions made by the reviewer while composing their feedback. This results in sequences of review drafts that can be aligned chronologically via timestamps.

\paragraph{Inline comments and voting}
Beyond document-level feedback, the platform supports fine-grained, span-based comments on the exposé draft. For each exposé, we collect a set of inline comments created by students and instructors while reading the draft. Each comment is attached to a character-level span in the PDF-rendered text of the \expose{} and is associated with (i) optional free-text content, (ii) a categorical tag from a fixed taxonomy
$\mathcal{Y} =
\{\textit{Strength}, \textit{Weakness}, \textit{Highlight}, \textit{Other}\}$
and (iii) metadata about the author role (student or instructor) and the underlying review document it belongs to.

The platform further allows other participants, but not the comment author, to cast binary votes (\emph{thumbs up} / \emph{thumbs down}) on each comment. These votes were intended as judgments of comment helpfulness. However, interviews with participants indicate that student authors sometimes used the voting mechanism as a personal \textit{checklist} during revision (marking comments as \textit{done} rather than as helpful), despite explicit instructions from the instructors to use it purely as a helpfulness signal. We therefore provide the raw voting data, together with information about voter role, but caution that votes do not form a clean ground truth for perceived usefulness.

\paragraph{Behavioral interaction logs}
For users who provided explicit additional consent, we release interaction logs from the review platform, covering 45 students and 9 instructors. Each record contains an \texttt{action} type, a JSON-encoded \texttt{data} payload, a \texttt{timestamp}, and an anonymized \texttt{user} identifier. Logged events are timestamped UI interactions (Table~\ref{tab:behaviour-actions-by-group}); an overview of activity over the semester is shown in Figure~\ref{fig:appendix:behaviour:daily_activity} and discussed in Appendix~\ref{appendix:behavior} . Actions including navigation and visibility changes (e.g., \texttt{Navigation step}, \texttt{PDF page show/hide}, \texttt{Browser Tab show/hide}), scrolling and resizing (e.g., \texttt{Annotation Page scroll}, \texttt{PDF scroll}, \texttt{Sidebar scroll}, \texttt{PDF (Browser) resize}), interface controls and modal events (e.g., \texttt{Sidebar click}, \texttt{Topbar click}, \texttt{Modal show}/\texttt{Modal hide}, \texttt{Modal Button click}), and lightweight editing interactions (e.g., \texttt{Text select}, \texttt{Copy text}, \texttt{Text paste}).

\begin{figure*}[t]
  \centering
  \includegraphics[page=1,width=\textwidth]{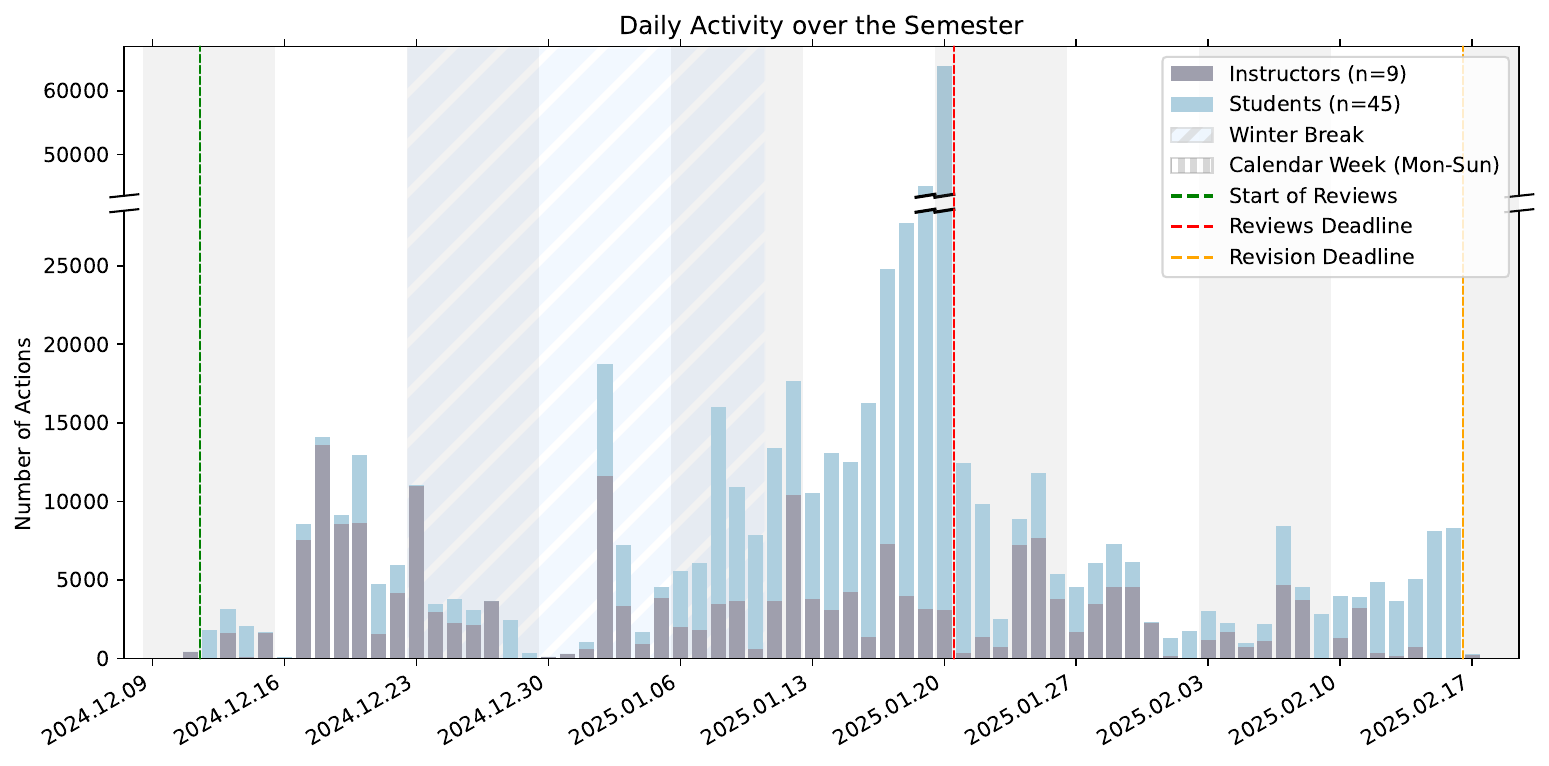}
  \caption{Daily number of actions across the semester for instructors (n=9) and students (n=45). Bars show per-day activity (instructors in gray, students in blue). Alternating vertical shading indicates calendar weeks (Mon--Sun), and the hatched region marks the winter break. Dashed lines mark the start of reviews (green), the review deadline (red), and the revision deadline (orange). The y-axis is broken to accommodate the peak activity before the review deadline.}
  \label{fig:appendix:behaviour:daily_activity}
\end{figure*}

\begin{table}[t]
\centering
\setlength{\tabcolsep}{3.5pt}
\resizebox{\linewidth}{!}{%
\begin{tabular}{lrrr}
\toprule
Action & \shortstack[r]{Students\\(n=45)} & \shortstack[r]{Instructors\\(n=9)} & Total \\
\midrule
Mouse move & 73,811 & 131,272 & 205,083 \\
PDF page show/hide & 53,378 & 102,120 & 155,498 \\
Annotation Page scroll & 18,944 & 31,080 & 50,024 \\
Browser Tab show/hide & 16,810 & 31,657 & 48,467 \\
PDF (Browser) resize & 13,377 & 25,475 & 38,852 \\
Text select & 7,403 & 8,043 & 15,446 \\
Sidebar scroll & 2,445 & 7,248 & 9,693 \\
Navigation step & 2,705 & 3,742 & 6,447 \\
Sidebar click & 2,882 & 3,382 & 6,264 \\
Topbar click & 1,826 & 3,512 & 5,338 \\
PDF scroll & 2,206 & 2,179 & 4,385 \\
Copy text & 1,403 & 2,371 & 3,774 \\
Table click & 1,811 & 1,164 & 2,975 \\
Text paste & 1,208 & 1,435 & 2,643 \\
Browser load & 715 & 1,364 & 2,079 \\
Modal show & 889 & 324 & 1,213 \\
Modal hide & 396 & 304 & 700 \\
Search used & 102 & 369 & 471 \\
Modal Button click & 259 & 176 & 435 \\
\midrule
\textbf{Total} & 202,570 & 357,217 & 559,787 \\
\bottomrule
\end{tabular}
}
\caption{Number of behaviour data rows per action, split by user group for the first cohort.}
\label{tab:behaviour-actions-by-group}
\end{table}

\paragraph{Authorship metadata and anonymization}
Each exposé, review, inline comment, vote, and interaction event is associated with an anonymized author identifier and a role label (\textit{student}, \textit{junior}, \textit{senior}). To support more interpretable examples and qualitative analysis, we replace real names with consistently applied synthetic names generated with \texttt{Faker}.\footnote{\url{https://faker.readthedocs.io/en/master}} No raw platform identifiers, email addresses, or other personally identifying information are released. This setup enables the analysis of role-specific behavior and bias while preserving participant privacy.

\begin{table*}[t]
    \centering
    \begingroup
    \setlength{\tabcolsep}{8pt}
    \renewcommand{\arraystretch}{1}

    \begin{tabular}{@{}lccc@{}}
        \toprule
        & \textbf{WS 2024/25}
        & \multicolumn{2}{c}{\textbf{WS 2025/26}} \\
        \cmidrule(lr){2-2}
        \cmidrule(lr){3-4}
        \textbf{Rubric}
        & \emph{Group 1}
        & \emph{Control}
        & \emph{All} \\
        \midrule

        \multicolumn{4}{@{}l}{\emph{Exposé drafts} ($n=55$, $78$, and $162$, respectively)} \\
        Language / Quality & 2.69 ($\pm$1.03) & 3.22 ($\pm$0.80) & 3.33 ($\pm$0.75) \\
        Metadata & 2.75 ($\pm$0.52) & 2.73 ($\pm$0.45) & 2.73 ($\pm$0.44) \\
        Form / Structure & 1.96 ($\pm$0.19) & 1.92 ($\pm$0.27) & 1.94 ($\pm$0.24) \\
        Motivation & 6.93 ($\pm$1.48) & 7.81 ($\pm$1.46) & 7.84 ($\pm$1.41) \\
        Approach & 7.91 ($\pm$2.27) & 8.90 ($\pm$2.57) & 8.88 ($\pm$2.38) \\
        Methodology & 4.07 ($\pm$1.33) & 4.05 ($\pm$1.48) & 4.01 ($\pm$1.34) \\
        Schedule & 4.15 ($\pm$0.85) & 4.29 ($\pm$1.34) & 4.46 ($\pm$1.07) \\
        Bibliography & 2.56 ($\pm$0.86) & 2.44 ($\pm$0.86) & 2.47 ($\pm$0.78) \\
        Additional points & $-0.02$ ($\pm$0.36) & $-0.22$ ($\pm$0.64) & $-0.18$ ($\pm$0.58) \\
        \textbf{Total} & \textbf{33.00 ($\pm$6.20)} & \textbf{35.14 ($\pm$6.74)} & \textbf{35.49 ($\pm$5.91)} \\

        \midrule
        \multicolumn{4}{@{}l}{\emph{Exposé finals} ($n=55$ and $162$, respectively)} \\
        Language / Quality & 3.45 ($\pm$0.79) & \multicolumn{2}{c}{3.75 ($\pm$0.50)} \\
        Metadata & 2.95 ($\pm$0.23) & \multicolumn{2}{c}{2.85 ($\pm$0.36)} \\
        Form / Structure & 1.96 ($\pm$0.19) & \multicolumn{2}{c}{1.98 ($\pm$0.16)} \\
        Motivation & 8.09 ($\pm$1.02) & \multicolumn{2}{c}{8.52 ($\pm$0.91)} \\
        Approach & 9.42 ($\pm$2.07) & \multicolumn{2}{c}{10.20 ($\pm$1.44)} \\
        Methodology & 4.76 ($\pm$0.58) & \multicolumn{2}{c}{4.72 ($\pm$0.74)} \\
        Schedule & 4.65 ($\pm$0.55) & \multicolumn{2}{c}{4.81 ($\pm$0.48)} \\
        Bibliography & 2.91 ($\pm$0.29) & \multicolumn{2}{c}{2.75 ($\pm$0.51)} \\
        Additional points & 0.00 ($\pm$0.27) & \multicolumn{2}{c}{$-0.05$ ($\pm$0.44)} \\
        \textbf{Total} & \textbf{38.20 ($\pm$3.87)} & \multicolumn{2}{c}{\textbf{39.52 ($\pm$3.24)}} \\

        \midrule
        \multicolumn{4}{@{}l}{\emph{Reviews} ($n=112$ and $325$, respectively)} \\
        Structure and Clarity & 3.26 ($\pm$1.04) & \multicolumn{2}{c}{3.67 ($\pm$0.73)} \\
        Language & 2.82 ($\pm$0.47) & \multicolumn{2}{c}{2.79 ($\pm$0.46)} \\
        Feed Up & 2.37 ($\pm$1.49) & \multicolumn{2}{c}{1.59 ($\pm$1.62)} \\
        Feed Back & 3.96 ($\pm$0.31) & \multicolumn{2}{c}{3.93 ($\pm$0.35)} \\
        Feed Forward & 1.74 ($\pm$0.50) & \multicolumn{2}{c}{1.14 ($\pm$0.73)} \\
        Content Quality & 4.51 ($\pm$0.79) & \multicolumn{2}{c}{4.46 ($\pm$0.88)} \\
        Errors & 3.71 ($\pm$0.62) & \multicolumn{2}{c}{3.78 ($\pm$0.60)} \\
        Additional points & $-0.04$ ($\pm$0.19) & \multicolumn{2}{c}{$-0.08$ ($\pm$0.29)} \\
        \textbf{Total} & \textbf{22.33 ($\pm$3.46)} & \multicolumn{2}{c}{\textbf{21.27 ($\pm$3.48)}} \\
        \bottomrule
    \end{tabular}

    \endgroup
    \caption{Per-rubric scores by cohort and assessment stage. Values are means ($\pm$SD). \emph{Control} includes gradings by graders who did not see the LLM pre-assessment, whereas \emph{All} also includes LLM-assisted draft gradings, whose scores may be anchored toward the model proposal.}
    \label{tab:cohort-scores}
\end{table*}

\section{Guidelines on Criteria and Rubrics for Exposé and Review Assessment}
\label{appendix:criteria}
This appendix provides an overview of \expose{} and review criteria and the guidelines used by the instructors when grading the \expose{}s and student reviews, respectively.

\paragraph{Expertise levels} Each criterion is assigned an \emph{expertise level}\footnote{Two co-authors inspected all criteria and labeled each for the expertise level required for scoring an \expose{} for a given criterion.}, indicating the type of judgment needed to assess the quality of the \expose{} or review with respect to this criterion. We introduce a three-level expertise scheme (expertise levels by criteria are shown in Tables~\ref{tab:expose_rubric} and \ref{tab:review_rubric} for \expose{} and review criteria, respectively):
\begin{itemize}
    \item \textbf{Level 0 (formal).} No academic or expert knowledge required. Largely objective. Sample \expose{} criteria: \emph{Metadata available}; \emph{Template used}; \emph{Number of pages}. Review criteria: None. 
    \item \textbf{Level 1 (academic).} Comprehension-based judgments of clarity, coherence, and basic research sense-making, but no specific in-domain expertise is required. Sample \expose{} criteria: \emph{Language quality}; \emph{Hook existing}. Sample review criteria: \emph{Summary available}; \emph{Language}.
    \item \textbf{Level 2 (expert).} Expert judgment of the soundness and adequacy of the reasoning and its alignment with disciplinary standards. Sample \expose{} criteria: \emph{Teaser (RQ)}; \emph{Relevance of theoretical framework}. Sample review criteria: \emph{Knowledge of correct response}; \emph{Learning goals}.
\end{itemize}

\subsection{Annotation Guidelines: Exposé Grading Criteria and Rubrics}
\label{appendix:expose-criteria}

Table~\ref{tab:expose_rubric} lists all the 36 \expose{} assessment criteria, grouped into nine high-level rubrics. These criteria define the scoring rules applied to each submitted exposé.

\paragraph*{\textbf{Refinement of criteria and rubrics for \expose{} assessment}} Before the course was run, the designed criteria and rubrics were piloted on exposés previously submitted within the department: two members of the instructional staff assessed together a small sample of old student \expose{}s, and other instructional staff members went over these to give feedback. This feedback was used to verify that every criterion is assessable and the criteria definitions are clear. Criteria definitions were adjusted accordingly for clarity and consistency. 

\subsection{Annotation Guidelines: Review Grading Criteria and Rubrics}
\label{appendix:review-criteria}
Table~\ref{tab:review_rubric} lists all the 24 review assessment criteria grouped into eight high-level rubrics. These criteria define the scoring rules applied to each submitted student review. While actionable feedback promotes learning, ineffective or misleading reviews can have the opposite effect. 
To this end, inspired by \newcite{du-etal-2024-llms}, the criteria include a penalty for unhelpful feedback patterns (e.g., neglect of key elements, unsupported claims, or contradictory advice).

\paragraph*{\textbf{Refinement of criteria and rubric for review assessment}} As with the exposé rubrics, all criteria and rubrics were reviewed by instructors and piloted on sample reviews before course deployment. 

\onecolumn
{
\begin{longtable}{>{\raggedright\arraybackslash}p{1cm}>{\raggedright\arraybackslash}p{3cm}>{\centering\arraybackslash}p{1cm}>{\raggedright\arraybackslash}p{\dimexpr\textwidth-1cm-3cm-1cm-8\tabcolsep\relax}}
\toprule
\textbf{Code} & \textbf{Criterion} & \textbf{Expert} & \textbf{Point descriptors}\\
&&\textbf{level}&\\
\midrule
\endfirsthead

\toprule
\textbf{Code} & \textbf{Criterion} & \textbf{Expert} & \textbf{Point descriptors}\\
&&\textbf{level}&\\
\midrule
\endhead

\midrule
\multicolumn{4}{r}{\textit{(continued on next page)}} \\
\bottomrule
\endfoot

\endlastfoot

\specialrule{1.2pt}{2pt}{2pt}
\multicolumn{4}{l}{\textbf{Rubric:} Language/Quality (LA)} \\
\multicolumn{4}{>{\raggedright\arraybackslash}p{\dimexpr\textwidth-2\tabcolsep\relax}}{\textit{Assessment of language quality and structural coherence of the text.}} \\
\midrule


LA1 & Language quality & 1 &
\textbf{0:} many linguistic errors (e.g. wrong tense, wrong personal pronouns)\\
& & & \textbf{1:} few linguistic errors, clear and comprehensible sentences\\
& & & \textbf{2:} no linguistic errors, clear and comprehensible sentences with consistent terminology\\[0.4em]

LA2 & Common Thread & 1 &
\textbf{0:} No recognizable common thread, the structure is unclear and jumpy\\
& & & \textbf{1:} Partly recognizable, but the structure is not continuous\\
& & & \textbf{2:} Text has a continuous structure that is easy to follow, common thread is clear and recognizable from beginning to end\\[1em]

\specialrule{1.2pt}{2pt}{2pt}
\multicolumn{4}{l}{\textbf{Rubric:} Metadata (ME)} \\
\multicolumn{4}{>{\raggedright\arraybackslash}p{\dimexpr\textwidth-2\tabcolsep\relax}}{\textit{Evaluation of title appropriateness and creativity.}} \\
\midrule


ME1 & Preliminary title & 1 &
\textbf{0:} Title does not match the topic\\
& & & \textbf{1:} Title fits the topic; similar title to the one in the topic suggestions\\
& & & \textbf{2:} Title fits the topic; creative title changed in a novel way and/or fits into the story of the exposé\\[0.4em]

ME2 & Metadata available & 0 &
\textbf{0:} Name, date and matriculation number are not entered correctly/changed\\
& & & \textbf{1:} Name, date, matriculation number are correct and updated\\[1em]

\specialrule{1.2pt}{2pt}{2pt}
\multicolumn{4}{l}{\textbf{Rubric:} Form/Structure (ST)} \\
\multicolumn{4}{>{\raggedright\arraybackslash}p{\dimexpr\textwidth-2\tabcolsep\relax}}{\textit{Evaluation of document formatting and structural requirements.}} \\
\midrule


ST1 & Template used & 0 &
\textbf{0:} Latex template is not used\\
& & & \textbf{1:} Latex template is used\\[0.4em]

ST2 & Number of pages & 0 &
\textbf{0:} > three pages (motivation + approach only)\\
& & & \textbf{1:} <= three sides (motivation + approach only)\\[1em]

\specialrule{1.2pt}{2pt}{2pt}
\multicolumn{4}{l}{\textbf{Rubric:} Motivation (MO)} \\
\multicolumn{4}{>{\raggedright\arraybackslash}p{\dimexpr\textwidth-2\tabcolsep\relax}}{\textit{Assessment of problem identification, context establishment, and research question formulation.}} \\
\midrule


MO1 & Hook existing & 1 &
\textbf{0:} Hook not available\\
& & & \textbf{1:} Hook present and suitable for the topic\\[0.4em]

MO2 & Anker & 1 &
\textbf{0:} The problem is not mentioned at all.\\
& & & \textbf{1:} The problem is stated in general terms without specific details. It is recognizable which problem is being addressed, but important information is omitted.\\
& & & \textbf{2:} The problem is described in detail. All relevant aspects of the problem are mentioned and clarify its scope and possible effects.\\[0.4em]

MO3 & Domain & 1 &
\textbf{0:} The domain or context in which the problem occurs is not mentioned.\\
& & & \textbf{1:} The domain or the relevant expert area of the problem is mentioned. It is clear in which environment the problem exists.\\[0.4em]

MO4 & Problem relevance & 1 &
\textbf{0:} The relevance or importance of the problem is not addressed. It remains unclear why this problem is relevant or worth discussing.\\
& & & \textbf{1:} The importance and relevance of the problem is described. It is made clear why the problem is important and what impact or consequences it could have (e.g. for specific individuals, organisations or society as a whole).\\[0.4em]

MO5 & Problem handling & 1 &
\textbf{0:} current handling of the problem or current criticism is not given\\
& & & \textbf{1:} current handling of the problem or current criticism is given\\[0.4em]

MO6 & Teaser (RQ) & 2 &
\textbf{0:} There is no research question available\\
& & & \textbf{1:} A research question exists, but it has weaknesses, e.g. due to unclear terminology or vague formulation. It is difficult to recognise what is to be investigated and how the question fits the problem description.\\
& & & \textbf{2:} The research question is formulated clearly and precisely. It clearly conveys what is to be investigated, and fits thematically with the problem description.\\[0.4em]

MO7 & RQ Limitation & 2 &
\textbf{0:} The research question is either too general or too broad, so that it does not seem realistic to answer it within the scope of a Bachelor's thesis, to answer it in the context of a Bachelor's thesis.\\
& & & \textbf{1:} The research question is precise and clearly defined so that it can realistically be addressed in the context of a Bachelor's thesis. The question is formulated in concrete terms and describes specifically which aspects are to be investigated without deviating from the topic.\\[1em]

\specialrule{1.2pt}{2pt}{2pt}
\multicolumn{4}{l}{\textbf{Rubric:} Approach (AP)} \\
\multicolumn{4}{>{\raggedright\arraybackslash}p{\dimexpr\textwidth-2\tabcolsep\relax}}{\textit{Assessment of state of the art analysis, theoretical framework.}} \\
\midrule


AP1 & SOTA & 1 &
\textbf{0:} SOTA not available\\
& & & \textbf{1:} SOTA present and correctly cited\\[0.4em]

AP2 & SOTA Relevance & 1 &
\textbf{0:} The relevance of SOTA is not present or unclear.\\
& & & \textbf{1:} The relevance of SOTA is mentioned, but is only partially comprehensible or weakly presented.\\
& & & \textbf{2:} It is clearly shown how SOTA is relevant to the topic of the exposé and why it is relevant to the research.\\[0.4em]

AP3 & SOTA Weaknesses & 1 &
\textbf{0:} SOTA's weak points are not mentioned\\
& & & \textbf{1:} SOTA's weaknesses are mentioned, but only superficially or unclearly described\\
& & & \textbf{2:} SOTA's weaknesses are clearly identified and precisely explained; it is clear which weaknesses are relevant for own research\\[0.4em]

AP4 & SOTA Delimitation & 1 &
\textbf{0:} No differentiation of own performance from the current state of research\\
& & & \textbf{1:} The own achievement is differentiated from SOTA; it is explicitly emphasized, what is new or unique about your own research\\[0.4em]

AP5 & SOTA Combination & 2 &
\textbf{0:} No meaningful combination of existing material (even if reference is made to existing material, but its combination or consolidation is unconvincing or unclear)\\
& & & \textbf{1:} Existing material is clearly and meaningfully combined; the added value of the combination is clear.\\[0.4em]

AP6 & Theoretical Framework & 1 &
\textbf{0:} The theoretical framework does not exist or is completely misleading\\
& & & \textbf{1:} The theoretical framework is present, but unclear, poorly structured or difficult to understand\\
& & & \textbf{2:} The theoretical framework is clearly and logically structured. The theoretical approaches are presented in a understandable and comprehensible\\[0.4em]

AP7 & Relevance of theoretical framework & 2 &
\textbf{0:} The theoretical framework shows no connection to the research topic, or the theory is irrelevant\\
& & & \textbf{1:} The theoretical framework clearly demonstrates how the selected theories directly relate to the research topic and support the research question\\[0.4em]

AP8 & Methodology Availability & 1 &
\textbf{0:} Methodology not available\\
& & & \textbf{1:} Methodology available and suitable for answering the RQ(s)\\[1em]

\specialrule{1.2pt}{2pt}{2pt}
\multicolumn{4}{l}{\textbf{Rubric:} Methodology (MD)} \\
\multicolumn{4}{>{\raggedright\arraybackslash}p{\dimexpr\textwidth-2\tabcolsep\relax}}{\textit{Assessment of the methodology.}} \\
\midrule


MD1 & Methodology Completeness & 1 &
\textbf{0:} Methodology is available, but not all RQ(s) are addressed\\
& & & \textbf{1:} Methodology is available and addresses all points of the research questions\\[0.4em]

MD2 & Methodology Relevance & 1 &
\textbf{0:} The relevance of the methodology for the research project is not clear or is unclear\\
& & & \textbf{1:} It is clearly and precisely demonstrated why the chosen methodology is relevant to the research project and how it contributes to the research questions\\[0.4em]

MD3 & Methodology Target group & 1 &
\textbf{0:} The target group is not specified or there is no precise differentiation\\
& & & \textbf{1:} The target group is clearly and precisely defined and it is clear why it is relevant to the methodology\\[0.4em]

MD4 & Methodology Existing Material & 1 &
\textbf{0:} No reference is made to the existing material or the material is inappropriate\\
& & & \textbf{1:} The existing material is appropriately and accurately related to the topic; it is clear that this material is relevant to the methodology\\[0.4em]

MD5 & Methodology Difficulties & 2 &
\textbf{0:} Potential difficulties in the practical implementation of the methods are not addressed. Potential challenges that could hinder the research process are overlooked.\\
& & & \textbf{1:} Potential difficulties in the practical implementation of the methods are described clearly and precisely. Concrete solutions or strategies to overcome these difficulties are also presented, in order to organize the research process as efficiently and target-oriented as possible.\\[0.4em]

MD6 & Methodology Possibilities/restrictions & 2 &
\textbf{0:} The possibilities or limitations of the methods used are not addressed, or they are only mentioned superficially, without concrete details. It remains unclear how these methods contribute to answering the research question and which limitations could possibly influence the validity of the results.\\
& & & \textbf{1:} The possibilities and limitations of the methods are described clearly and precisely. It is clear what strengths the methods bring to the study and what weaknesses or limitations could possibly influence the results. limitations could possibly influence the results. The choice of methods is critically reflected upon and related to the research project.\\[0.4em]

MD7 & Methodology Details & 2 &
\textbf{0:} No additional explanations are given on the methodology or implementation\\
& & & \textbf{1:} The methodology is explained comprehensively and precisely with additional explanations (e.g. details on implementation),  so that the procedure is clearly comprehensible\\[1em]

\specialrule{1.2pt}{2pt}{2pt}
\multicolumn{4}{l}{\textbf{Rubric:} Schedule (SC)} \\
\multicolumn{4}{>{\raggedright\arraybackslash}p{\dimexpr\textwidth-2\tabcolsep\relax}}{\textit{Assessment of project timeline structure, completeness, and realism.}} \\
\midrule


SC1 & Schedule Availability & 0 &
\textbf{0:} Tabular schedule not available\\
& & & \textbf{1:} Tabular schedule in chronological order available\\[0.4em]

SC2 & Schedule Completeness & 1 &
\textbf{0:} Tabular schedule contains gaps\\
& & & \textbf{1:} Tabular schedule available from start to finish and reasonably divided into blocks\\[0.4em]

SC3 & Schedule Block Description & 0 &
\textbf{0:} not available or only headlines available\\
& & & \textbf{1:} individual blocks described (not only headlines)\\[0.4em]

SC4 & Schedule Realistic Relevance & 1 &
\textbf{0:} The time distribution is obviously unrealistic and important work steps are missing. The research questions are not or only insufficiently addressed\\
& & & \textbf{1:} The timeline contains the most important steps, but some of the timelines are unrealistic or unclear. It is partially unclear how the steps contribute to answering the research questions. The timeline does not appear to be fully aligned with the Bachelor's thesis context.\\
& & & \textbf{2:} The timeline is clearly structured, realistic and aligned with the scope of the Bachelor's thesis. All important work steps are included and organised in a sensible time frame. Each step contributes to answering the research questions, and the entire plan fits coherently into the exposé and the context of the thesis.\\[1em]

\specialrule{1.2pt}{2pt}{2pt}
\multicolumn{4}{l}{\textbf{Rubric:} Bibliography (BI)} \\
\multicolumn{4}{>{\raggedright\arraybackslash}p{\dimexpr\textwidth-2\tabcolsep\relax}}{\textit{Assessment of bibliography consistency and literature relevance.}} \\
\midrule


BI1 & Bibliography Consistency & 0 &
\textbf{0:} Bibliography with different citation styles\\
& & & \textbf{1:} Bibliography is standardized ( consistent citation style) and essentially all information on the literature is available\\[0.4em]

BI2 & Key literature & 2 &
\textbf{0:} Literature does not fit the topic\\
& & & \textbf{1:} Literature fits the topic\\
& & & \textbf{2:} Literature fits the topic and most of it is highly relevant\\[1em]

\specialrule{1.2pt}{2pt}{2pt}
\multicolumn{4}{l}{\textbf{Rubric:} Additional points (AD)} \\
\multicolumn{4}{>{\raggedright\arraybackslash}p{\dimexpr\textwidth-2\tabcolsep\relax}}{\textit{Allow additional points for very good submissions and major mistakes in the submission.}} \\
\midrule


AD1 & Additional points & 1 &
\textbf{0:} No additional strengths beyond existing criteria.\\
& & & \textbf{1:} Some additional strengths beyond existing criteria.\\
& & & \textbf{2:} Clear exceptional strengths beyond existing criteria.\\[0.4em]

AD2 & Negative points & 1 &
\textbf{0:} No major issues beyond existing criteria.\\
& & & \textbf{-1:} Notable issues beyond existing criteria.\\
& & & \textbf{-2:} Serious issues that strongly reduce overall quality.\\[1em]

\bottomrule
\caption{Guidelines for \expose{} grading. 36 criteria used to grade student \expose{}s. We organize the criteria into nine rubrics, by grouping together criteria that evaluate related aspects of \expose{} writing.} \\
\label{tab:expose_rubric} \\
\end{longtable}
}
\twocolumn

\onecolumn
{
\begin{longtable}{>{\raggedright\arraybackslash}p{1cm}>{\raggedright\arraybackslash}p{3cm}>{\centering\arraybackslash}p{1cm}>{\raggedright\arraybackslash}p{\dimexpr\textwidth-1cm-3cm-1cm-8\tabcolsep\relax}}
\toprule
\textbf{Code} & \textbf{Criterion} & \textbf{Expert} & \textbf{Point descriptors} \\
\midrule
\endfirsthead

\toprule
\textbf{Code} & \textbf{Criterion} & \textbf{Expert} & \textbf{Point descriptors}\\
\midrule
\endhead

\midrule
\multicolumn{4}{r}{\textit{(continued on next page)}} \\
\bottomrule
\endfoot

\endlastfoot

\specialrule{1.2pt}{2pt}{2pt}
\multicolumn{4}{l}{\textbf{Rubric:} Structure and Clarity (SC)} \\
\multicolumn{4}{>{\raggedright\arraybackslash}p{\dimexpr\textwidth-2\tabcolsep\relax}}{\textit{Assessment of overall structure and clarity.}} \\
\midrule


SC1 & Summary available & 1 &
\textbf{0:} No summary available or, the summary does not reflect understanding of the performance\\
& & & \textbf{1:} Simple summary that demonstrates basic understanding of the performance\\
& & & \textbf{2:} Concise and accurate summary that clearly reflects understanding of performance and builds confidence in feedback \\[0.4em]

SC2 & Clarity & 2 &
\textbf{0:} Feedback needs considerable improvement in clarity and comprehensibility, comments are disjointed or difficult to follow\\
& & & \textbf{1:} Feedback is largely clear, but sometimes lacks specific references to particular lines, pages, sections or figures/tables\\
& & & \textbf{2:} Feedback is clear, well-structured and easy to understand, with coherent comments and precise references to specific points in the text\\[1em]

\specialrule{1.2pt}{2pt}{2pt}
\multicolumn{4}{l}{\textbf{Rubric:} Language (LG)} \\
\multicolumn{4}{>{\raggedright\arraybackslash}p{\dimexpr\textwidth-2\tabcolsep\relax}}{\textit{Assessment of language quality and tone.}} \\
\midrule


LG1 & Language & 1 &
\textbf{0:} Many linguistic errors (e.g. wrong tense, wrong personal pronouns)\\
& & & \textbf{1:} Almost no linguistic errors, clear and comprehensible sentences with standardised terminology\\[0.4em]

LG2 & Tone & 1 &
\textbf{0:} The tone is inappropriate, sounds judgemental or reproachful, criticism comes across as destructive or personal\\
& & & \textbf{1:} The tone is generally respectful, avoids personal attacks, but remains unclear or less friendly in places\\
& & & \textbf{2:} The tone is consistently calm, respectful and polite; criticism is constructive, factual and only directed at the text, not the person\\[1em]

\specialrule{1.2pt}{2pt}{2pt}
\multicolumn{4}{l}{\textbf{Rubric:} Feed Up (FU)} \\
\multicolumn{4}{>{\raggedright\arraybackslash}p{\dimexpr\textwidth-2\tabcolsep\relax}}{\textit{Assessment of learning goals and success criteria. }} \\
\midrule


FU1 & Learning Goals & 2 &
\textbf{0:} No learning objectives, unclear what is to be achieved, no orientation\\
& & & \textbf{1:} Partly unclear learning objectives, generally formulated, offer only limited orientation\\
& & & \textbf{2:} Clear learning objectives, specific, challenging, comparable and provide clear direction\\[0.4em]

FU2 & Success Criteria & 1 &
\textbf{0:} No success criteria, success unclear or vaguely formulated, objectives not defined\\
& & & \textbf{1:} Partly unclear success criteria, available but imprecise or general, orientation partly missing\\
& & & \textbf{2:} Clear success criteria, concrete, measurable goals and clear definition of success\\[1em]

\specialrule{1.2pt}{2pt}{2pt}
\multicolumn{4}{l}{\textbf{Rubric:} Feed Back (FB)} \\
\multicolumn{4}{>{\raggedright\arraybackslash}p{\dimexpr\textwidth-2\tabcolsep\relax}}{\textit{Assessment of feedback quality and knowledge transfer.}} \\
\midrule


FB1 & Knowledge of result & 1 &
\textbf{0:} No information as to whether the task has been solved or not\\
& & & \textbf{1:} Information given as to whether the task has been solved or not\\[0.4em]

FB2 & Knowledge of correct response & 2 &
\textbf{0:} No information on the correct answer\\
& & & \textbf{1:} Correct answer is clearly communicated\\[0.4em]

FB3 & Knowledge about task constraints & 1 &
\textbf{0:} No information on rules, requirements or restrictions of the task\\
& & & \textbf{1:} Clear information on rules, requirements or restrictions of the task (e.g. specifications)\\[0.4em]

FB4 & Knowledge about concepts & 1 &
\textbf{0:} No information on conceptual knowledge\\
& & & \textbf{1:} Basic information on relevant concepts is provided\\[0.4em]

FB5 & Knowledge about mistakes & 1 &
\textbf{0:} No information about errors\\
& & & \textbf{1:} Information on the number, location, source or type of errors\\[0.4em]

FB6 & Self-feedback & 1 &
\textbf{0:} Self-feedback is available that has not been combined with task, process or self-regulation (including feedback on the learner's abilities (i.e. intelligence or talent))\\
& & & \textbf{1:} No self-feedback is available or, if available, in combination with task, process or self-regulation\\[1em]

\specialrule{1.2pt}{2pt}{2pt}
\multicolumn{4}{l}{\textbf{Rubric:} Feed Forward (FF)} \\
\multicolumn{4}{>{\raggedright\arraybackslash}p{\dimexpr\textwidth-2\tabcolsep\relax}}{\textit{Assessment of future learning guidance.}} \\
\midrule


FF1 & Self-regulation & 1 &
\textbf{0:} No indication of how to approach the task, no recognisable approaches to self-monitoring or self-assessment\\
& & & \textbf{1:} Either indications of strategies for working on the task and recognising errors (e.g. suggestions for improvement, targeted search for information) or approaches to self-monitoring and self-assessment (e.g. reflection on own understanding, strategies or willingness to seek help)\\[0.4em]

FF2 & Learning Skills & 1 &
\textbf{0:} No information on how to develop or improve learning skills\\
& & & \textbf{1:} Information on how to improve the learning objective and how to deal with challenges\\[1em]

\specialrule{1.2pt}{2pt}{2pt}
\multicolumn{4}{l}{\textbf{Rubric:} Content Quality (CQ)} \\
\multicolumn{4}{>{\raggedright\arraybackslash}p{\dimexpr\textwidth-2\tabcolsep\relax}}{\textit{Assessment of content accuracy and usefulness.}} \\
\midrule


CQ1 & Correctness & 2 &
\textbf{0:} The content has significant deficiencies in terms of correctness\\
& & & \textbf{1:} The content is largely correct\\
& & & \textbf{2:} The content is completely correct\\[0.4em]

CQ2 & Actionability & 1 &
\textbf{0:} No specific instructions available, feedback is difficult to implement\\
& & & \textbf{1:} Some instructions available, but only partially clear or specific\\
& & & \textbf{2:} Clear, specific feedback with actionable instructions, e.g. comparisons with existing methods, application to other tasks, definitions, explanations, discussions, additional analyses or specific suggestions to remove confusion\\[0.4em]

CQ3 & Argumentation & 1 &
\textbf{0:} Statements or conclusions are not supported by evidence or comprehensible explanations. The argumentation remains superficial or unsystematic, making the content unconvincing.\\
& & & \textbf{1:} Statements are supported by comprehensible evidence, examples or explanations. The argumentation shows a clear link between evidence and conclusions, which strengthens the quality of the content.\\[1em]

\specialrule{1.2pt}{2pt}{2pt}
\multicolumn{4}{l}{\textbf{Rubric:} Errors (ER)} \\
\multicolumn{4}{>{\raggedright\arraybackslash}p{\dimexpr\textwidth-2\tabcolsep\relax}}{\textit{Assessment of potential errors and issues in feedback.}} \\
\midrule


ER1 & Neglect & 2 &
\textbf{0:} All important details have been considered and no unnecessary questions are asked\\
& & & \textbf{-1:} Important details have been overlooked, leading to unnecessary questions or criticism\\[0.4em]

ER2 & Vague Critic & 1 &
\textbf{0:} Criticism is specific and clearly states what is missing or can be improved\\
& & & \textbf{-1:} Unspecific criticism, mentions missing components without clearly stating what is missing\\[0.4em]

ER3 & Out-of-Scope & 1 &
\textbf{0:} Proposals remain within the scope of the task and are relevant\\
& & & \textbf{-1:} Suggestions refer to methods or analyses that go beyond the intended scope\\[0.4em]

ER4 & Missing Reference & 1 &
\textbf{0:} Alternative methods, etc. (if available) are supported with justification or references\\
& & & \textbf{-1:} Alternative methods, etc. (if available) are suggested without justification or references\\[0.4em]

ER5 & Contradiction & 1 &
\textbf{0:} Feedback is consistent in itself, no contradictory statements\\
& & & \textbf{-1:} Contradictory feedback, e.g. criticism and praise for the same methods\\[1em]

\specialrule{1.2pt}{2pt}{2pt}
\multicolumn{4}{l}{\textbf{Rubric:} Additional points (AD)} \\
\multicolumn{4}{>{\raggedright\arraybackslash}p{\dimexpr\textwidth-2\tabcolsep\relax}}{\textit{Allow additional points for very good reviews or deduct points for major mistakes.}} \\
\midrule


AD1 & Additional points & 1 &
\textbf{0:} No additional strengths identified.\\
& & & \textbf{1:} Notable additional strengths beyond other criteria.\\[0.4em]

AD2 & Negative points & 1 &
\textbf{0:} No additional major issues identified.\\
& & & \textbf{-1:} Significant issues beyond other criteria.\\[1em]

\bottomrule
\caption{Guidelines for review grading. 24 criteria used to grade student reviews. We organize the criteria into eight rubrics, by grouping together criteria that evaluate related aspects of review.} \\
\label{tab:review_rubric} \\
\end{longtable}
}
\twocolumn

\section{Summary of Collected Sources for Exposé Criteria}
\label{appendix:expose-collection}
\begin{table*}[t]
\centering
\setlength{\tabcolsep}{4pt}
\renewcommand{\arraystretch}{1.15}

\begin{tabular}{@{}p{5.1cm}p{1.2cm}ccccp{0.9cm}@{}}
\toprule
\textbf{Source} & \textbf{Lang} & \textbf{BA} & \textbf{MA} & \textbf{Diss.} & \textbf{Grant/Proj.} & \textbf{Link}\\
\midrule
TU Berlin & EN & \Yes & \Yes & \Yes & \Yes & [1] \\
FU Berlin & EN & \No & \Yes & \Yes & \No & [2] \\
Uni Leipzig & EN & \No & \No & \Yes & \No & [3] \\
Uni Magdeburg & EN & \No & \No & \Yes & \No & [4] \\
Uni Freiburg & DE/EN & \No & \No & \Yes & \No & [5] \\
Hochschule Magdeburg & EN & \No & \No & \Yes & \No & [6] \\
Uni Hannover & EN & \No & \No & \Yes & \No & [7] \\
TU Dresden & EN & \Yes & \Yes & \No & \No & [8] \\
Hochschule für Politik München & EN & \Yes & \Yes & \No & \No & [9] \\
LMU München & EN & \Yes & \Yes & \No & \No & [10] \\
Uni Wien & EN & \No & \Yes & \No & \No & [11] \\
Uni Kassel & EN & \No & \Yes & \No & \No & [12] \\
Uni Mannheim & EN & \No & \Yes & \No & \No & [13] \\
FH Erfurt & EN & \No & \Yes & \No & \No & [14] \\
Scribbr (DE) & DE & \Yes & \No & \No & \No & [15] \\
WU Wien & DE & \Yes & \No & \No & \No & [16] \\
Finito24 & DE & \Yes & \No & \No & \No & [17] \\
Uni Bremen & EN & \No & \No & \No & \No & [18] \\
\bottomrule
\end{tabular}
\caption{University and institutional exposé guidelines with explicit scope for Bachelor, Master, Dissertation, or Grant/Project proposals.}
\label{tab:expose_guidelines}
\end{table*}

Across the full corpus of 110 exposé guidelines we reviewed, we identified materials from universities and higher-education institutions in Germany, Austria, and Switzerland, complemented by student-facing writing portals (e.g., Scribbr, Finito24, Studieren.de), which we further investigated.
In terms of disciplinary scope, our primary sources are published by faculties and units in the humanities and education, the social sciences (including sociology and political science), medicine, economics and business, as well as applied/STEM fields (i.e., architecture, computer science, and civil engineering).
Roughly three-quarters of the sources are published in English, primarily from internationalized programs (e.g., TU Berlin, Universität Wien, TU Dresden, Uni Magdeburg), while the remainder appear in German and serve national undergraduate or graduate audiences. 
About one-fifth of the documents explicitly target Bachelor- or Master-level theses, and a smaller subset focuses on doctoral or dissertation exposés, such as those from the University of Leipzig Graduate Centre and the University of Magdeburg Medical Faculty. 
Only a few pages, most notably the TU Berlin departmental guide, explicitly include project or grant proposal contexts.

Thematically, nearly all guidelines converge on the same structural expectations: a clear \emph{Motivation} or problem statement, a theoretically informed \emph{Approach} or methodology, a feasible \emph{Schedule} or timeline, and a complete \emph{Bibliography}. 
Additional sections, such as \emph{Research Questions}, \emph{Related Work}, and \emph{Formal Aspects} (including format, citation, or layout), occur frequently, especially in English-language handbooks and institutional templates.

Across this heterogeneity, the documents show remarkable consistency in pedagogical intent: \emph{to ensure students articulate a focused question, justify methodological choices, and plan realistic execution.}
These recurrent patterns directly informed the rubric domains of \emph{Motivation}, \emph{Approach}, \emph{Schedule}, and \emph{Bibliography} used for annotation and model benchmarking in our dataset.

\vspace{1em}
\noindent\textbf{Links for Table~\ref{tab:expose_guidelines}}: \\
\setlength{\parskip}{1pt}
\sloppy
{\footnotesize
\begin{enumerate}[
  leftmargin=2.2em,
  labelsep=.6em,
  itemsep=1pt,
  topsep=2pt
]
\item \url{https://www.tu.berlin/en/ak/study-and-teaching/final-theses/writing-an-expose}
\item \url{https://www.oei.fu-berlin.de/soziologie/studiumlehre/handreichungen_zum_studium/How-to-write-an-expose.pdf}
\item \url{https://home.uni-leipzig.de/~gsgas/fileadmin/Recommendations/Expose_Recommendations.pdf}
\item \url{https://expae.med.ovgu.de/unimagdeburg_mm/Downloads/Kliniken/KPAE/EXPAE/Guidelines+for+writing+an+expos%C3%A9+for+a+doctoral+project.pdf}
\item \url{https://www.wvf.uni-freiburg.de/promhabil/merkblatt-erstellung-expose}
\item \url{https://www.h2.de/fileadmin/user_upload/Promotionszentren/PromZ_SGW/4eng_Expose-Hinweise_PromZ_SGW.pdf}
\item \url{https://www.archland.uni-hannover.de/fileadmin/archland/FORSCHUNG/promotionsausbildung/Requirements_admission_and_expose__.pdf}
\item \url{https://tu-dresden.de/ing/informatik/smt/im/studium/theses-and-research-projects/writing-an-expose}
\item \url{https://www.hfp.tum.de/fileadmin/w00cjd/governance/pdf/Instructions_for_Preparing_an_Expose.pdf}
\item \url{https://www.gsi.uni-muenchen.de/personen/wiss_mitarbeiter/kruck/general_guidelines.pdf}
\item \url{https://sts.univie.ac.at/fileadmin/user_upload/i_sts/Studium/Master_STS/05_Services_for_current_students/Master_Thesis/Guidelines_for_Writing_a_Master_Thesis_Expose.pdf}
\item \url{https://www.uni-kassel.de/fb11agrar/index.php?eID=dumpFile&t=f&f=572&token=eb5a0183632b6762477026bbd8f1694660f1b3f6}
\item \url{https://www.bwl.uni-mannheim.de/media/Lehrstuehle/bwl/Schoen/Teaching/Master/Expose_Guide/Master_Thesis_Expose_Guide_v6.pdf}
\item \url{https://www.fh-erfurt.de/fileadmin/Bilder/Seitenbereiche/Fakultaeten/BKR/B/Studium/MA_Sustainable_Engineering_of_Infrastructure/Expose_Recommendations_SEI.pdf}
\item \url{https://www.scribbr.de/anfang-abschlussarbeit/expose-bachelorarbeit/}
\item \url{https://www.wu.ac.at/fileadmin/wu/d/i/ign/IGN_BA-Expos%C3%A9_Requirements_De.pdf}
\item \url{https://www.finito24.de/expose-schreiben/}
\item \url{https://www.uni-bremen.de/fileadmin/user_upload/fachbereiche/fb7/gscm/Dokumente/Structure_of_an_Expose.pdf}
\end{enumerate}
}

\section{\nameDataset Dataset Statistics}
\label{appendix:statistics}
This section provides additional descriptive statistics for \nameDataset{} that complement the summary in Section~\ref{sec:dataset_analysis} of the main paper. To give a more detailed view of the dataset’s composition, we report basic statistics on the exposés, reviews, and assessment scores.

\subsection{Exposé Revision Statistics}
\label{appendix:subsec:expose-revision-stats}
Table~\ref{tab:appendix_expose_lexical} reports section-wise text statistics for \emph{Motivation}, \emph{Approach}, and \emph{Schedule} in draft vs.\ final versions (mean $\pm$ SD). Raw \LaTeX{} is converted to plain text with \texttt{pylatexenc}\footnote{\url{https://github.com/phfaist/pylatexenc}}; sentence/word segmentation uses spaCy. We report \textbf{length/structure} (\#Words, \#Sentences and average\ word length), \textbf{lexical/readability} (type--token ratio (TTR) and Flesch Reading Ease (FRE); \citealp{Flesch1948} via \texttt{textstat}\footnote{\url{https://github.com/textstat/textstat}}), and \textbf{draft--final similarity} (cosine similarity of SentenceTransformer embeddings\footnote{\texttt{sentence-transformers/all-mpnet-base-v2}} and normalized token-level Levenshtein similarity; \citealp{Levenshtein1965Binary}).

Across sections, final \expose{}s are longer in words and sentences, with the largest expansion in \emph{Approach}; average word length changes only marginally. Lexical diversity (TTR) decreases slightly from draft to final in all three sections. The negative FRE values for \emph{Schedule} likely reflect formatting artifacts (fragmentary, table-like content) interacting with plain-text conversion and segmentation.

\paragraph{Similarity and topic structure}
High embedding cosine similarities in Table~\ref{tab:appendix_expose_lexical} indicate strong semantic continuity between draft and final, even where token-level edit similarity is lower (notably for \emph{Motivation} and \emph{Approach}). Table~\ref{tab:expose_topic_pairwise_cosine} reports pairwise embedding cosine similarities within and across topics: within-topic pairs are generally more similar than cross-topic pairs, with the largest within-topic variance for \textit{Generative AI in Healthcare}.

\begin{table*}[ht]
\centering

\begin{tabular}{l | cc | cc | cc}
\toprule
 & \multicolumn{2}{c|}{\textbf{Motivation}} & \multicolumn{2}{c|}{\textbf{Approach}} & \multicolumn{2}{c}{\textbf{Schedule}} \\
\textbf{Metric} & Draft & Final & Draft & Final & Draft & Final \\
\midrule

\multicolumn{7}{l}{\textbf{Length and structure}} \\
\cmidrule(lr){1-7}
\#Words & 402 {\scriptsize ($\pm$133)} & 428 {\scriptsize ($\pm$125)} & 748 {\scriptsize ($\pm$258)} & 851 {\scriptsize ($\pm$221)} & 248 {\scriptsize ($\pm$83)} & 270 {\scriptsize ($\pm$131)}\\
\#Sentences & 21.8 {\scriptsize ($\pm$8.5)} & 24.0 {\scriptsize ($\pm$7.8)} & 40.9 {\scriptsize ($\pm$18.4)} & 47.7 {\scriptsize ($\pm$15.3)} & 7.3 {\scriptsize ($\pm$8.3)} & 8.4 {\scriptsize ($\pm$8.7)}\\
Avg.\ word length & 5.38 {\scriptsize ($\pm$0.45)} & 5.43 {\scriptsize ($\pm$0.44)} & 5.41 {\scriptsize ($\pm$0.49)} & 5.46 {\scriptsize ($\pm$0.50)} & 5.55 {\scriptsize ($\pm$0.37)} & 5.50 {\scriptsize ($\pm$0.36)}\\

\midrule

\multicolumn{7}{l}{\textbf{Lexical and readability}} \\
\cmidrule(lr){1-7}
Lexical diversity & 0.54 {\scriptsize ($\pm$0.06)} & 0.53 {\scriptsize ($\pm$0.06)} & 0.45 {\scriptsize ($\pm$0.07)} & 0.43 {\scriptsize ($\pm$0.06)} & 0.60 {\scriptsize ($\pm$0.06)} & 0.59 {\scriptsize ($\pm$0.06)}\\
Flesch reading ease & 43.0 {\scriptsize ($\pm$14.5)} & 41.6 {\scriptsize ($\pm$13.3)} & 42.1 {\scriptsize ($\pm$15.3)} & 41.0 {\scriptsize ($\pm$15.9)} & -50.4 {\scriptsize ($\pm$90.4)} & -34.2 {\scriptsize ($\pm$90.1)}\\

\midrule

\multicolumn{7}{l}{\textbf{Draft--Final similarity}} \\
\cmidrule(lr){1-7}
Cosine similarity & \multicolumn{2}{c|}{0.96 {\scriptsize ($\pm$0.04)}} & \multicolumn{2}{c|}{0.95 {\scriptsize ($\pm$0.06)}} & \multicolumn{2}{c}{0.93 {\scriptsize ($\pm$0.14)}}\\
Levenshtein similarity & \multicolumn{2}{c|}{0.64 {\scriptsize ($\pm$0.24)}} & \multicolumn{2}{c|}{0.62 {\scriptsize ($\pm$0.26)}} & \multicolumn{2}{c}{0.81 {\scriptsize ($\pm$0.24)}}\\

\bottomrule
\end{tabular}
\caption{Chapter-level textual statistics for expos\'es by version (draft vs.\ final). Entries show mean (±SD) for length, lexical diversity, readability, and draft--final similarity metrics.  }
\label{tab:appendix_expose_lexical}

\end{table*}

\begin{table*}[ht]
\centering
\begingroup

\begin{tabular}{l | cc | cc | cc}
\toprule
 & \multicolumn{2}{c|}{\textbf{Motivation}} & \multicolumn{2}{c|}{\textbf{Approach}} & \multicolumn{2}{c}{\textbf{Schedule}} \\
\textbf{Topic} & Draft & Final & Draft & Final & Draft & Final \\
\midrule
\textbf{T1} & 0.59 {\scriptsize ($\pm$0.13)} & 0.58 {\scriptsize ($\pm$0.13)} & 0.55 {\scriptsize ($\pm$0.14)} & 0.56 {\scriptsize ($\pm$0.14)} & 0.52 {\scriptsize ($\pm$0.16)} & 0.52 {\scriptsize ($\pm$0.16)}\\
\textbf{T2} & 0.60 {\scriptsize ($\pm$0.12)} & 0.66 {\scriptsize ($\pm$0.09)} & 0.60 {\scriptsize ($\pm$0.11)} & 0.62 {\scriptsize ($\pm$0.12)} & 0.59 {\scriptsize ($\pm$0.13)} & 0.61 {\scriptsize ($\pm$0.12)}\\
\textbf{T3} & 0.48 {\scriptsize ($\pm$0.14)} & 0.47 {\scriptsize ($\pm$0.14)} & 0.40 {\scriptsize ($\pm$0.17)} & 0.41 {\scriptsize ($\pm$0.16)} & 0.54 {\scriptsize ($\pm$0.13)} & 0.50 {\scriptsize ($\pm$0.11)}\\
\midrule
\textbf{Overall} & 0.37 {\scriptsize ($\pm$0.18)} & 0.38 {\scriptsize ($\pm$0.18)} & 0.35 {\scriptsize ($\pm$0.18)} & 0.35 {\scriptsize ($\pm$0.19)} & 0.48 {\scriptsize ($\pm$0.16)} & 0.46 {\scriptsize ($\pm$0.15)}\\
\bottomrule
\end{tabular}
\endgroup
\caption{Pairwise cosine similarity between expos\'es (each expos\'e is compared with all others within the same topic and version). Entries show mean (±SD) over all unordered document pairs, computed separately per chapter. Topic mapping: T1=\emph{The Holographic AI Assistant}, T2=\emph{Interactive LLM-based teaching agents}, T3=\emph{Generative AI in Healthcare}.}
\label{tab:expose_topic_pairwise_cosine}
\end{table*}

\paragraph{Bibliography size and stability}
Table~\ref{tab:appendix_bib} reports topic-wise bibliography statistics for draft vs.\ final expos\'es (mean $\pm$ SD), measured as the number of distinct Bib\TeX{} entry keys and their draft--final overlap (Jaccard similarity over key sets). Draft bibliography size differs substantially by topic (T1~$<$~T2~$<$~T3). Despite these initial gaps, revisions show a consistent pattern across topics: authors add about two references on average while removing fewer than one, suggesting modest expansion rather than wholesale replacement. Correspondingly, mean draft--final Jaccard similarity remains relatively high (overall $0.78$), indicating that bibliographies are generally stable across revisions with targeted additions.

\begin{table*}[t]
\centering
\begin{tabular}{l | c | ccc | c}
\toprule
 & \textbf{Draft} & \multicolumn{3}{c|}{\textbf{Final}} & \textbf{Similarity} \\
\textbf{Topic} & $\bar{n}_\text{draft}$ & $\bar{n}_\text{final}$ & $\bar{n}_\text{added}$ & $\bar{n}_\text{removed}$ & $\overline{\text{Jaccard}}$ \\
\midrule

\textbf{T1} & $6.74 \,\pm\, 6.02$ & $8.63 \,\pm\, 5.63$ & $2.16 \,\pm\, 3.86$ & $0.26 \,\pm\, 0.56$ & $0.73 \,\pm\, 0.40$ \\
\textbf{T2} & $9.22 \,\pm\, 6.39$ & $10.78 \,\pm\, 6.31$ & $2.06 \,\pm\, 2.62$ & $0.50 \,\pm\, 1.25$ & $0.76 \,\pm\, 0.27$ \\
\textbf{T3} & $11.83 \,\pm\, 8.01$ & $13.06 \,\pm\, 8.74$ & $1.89 \,\pm\, 2.97$ & $0.67 \,\pm\, 2.83$ & $0.84 \,\pm\, 0.26$ \\
\cmidrule(lr){1-6}
\textbf{Overall} & \textbf{$9.22 \,\pm\, 7.04$} & \textbf{$10.78 \,\pm\, 7.11$} & \textbf{$2.04 \,\pm\, 3.15$} & \textbf{$0.47 \,\pm\, 1.77$} & \textbf{$0.78 \,\pm\, 0.32$} \\
\bottomrule
\end{tabular}

\caption{Bibliography size and draft--final changes by topic. We report mean (±SD) number of entries in draft and final versions, average additions/removals, and mean draft--final entry-set Jaccard similarity computed over Bib\TeX{} keys. Topic mapping: T1=\emph{The Holographic AI Assistant}, T2=\emph{Interactive LLM-based Teaching Agents}, T3=\emph{Generative AI in Healthcare}.}
\label{tab:appendix_bib}
\end{table*}

\paragraph{Draft--final improvements on rubric scores}
Table~\ref{tab:appendix_score_improvements} reports per-rubric statistics for draft and final expos\'es. It complements Section~\ref{sec:dataset_statistics} by additionally showing (i) the remaining gap between the mean final score and the rubric maximum and (ii) the observed score ranges (min--max) for each version. As in the main text (Section~\ref{sec:dataset_statistics}), the largest mean gains occur in conceptually central rubrics such as \emph{Approach} and \emph{Motivation}. In contrast, smaller changes in \emph{Metadata}, \emph{Bibliography}, and \emph{Form/Structure} largely reflect limited headroom: these rubrics have low maxima, and draft scores are already close to the ceiling. Finally, \emph{Additional points} changes little because it represents bounded instructor adjustments rather than points that students can systematically earn through revision (see Appendix~\ref{appendix:subsec:expose-revision-stats:points} for the corresponding scoring rules). Overall, the revision signal is therefore most visible in rubrics with substantial remaining headroom, and is concentrated in substantive content and argumentation.

\begin{table*}[t]
\centering

\begin{tabular}{l|rrr|rr|rr}
\toprule
 & \multicolumn{3}{c|}{\textbf{Mean (±SD)}} & \multicolumn{2}{c|}{\textbf{Gaps}} & \multicolumn{2}{c}{\textbf{Observed bounds}}\\
\textbf{Rubric} & Draft & Final & $\Delta$ & Gap & Max & Draft & Final \\
\midrule
Approach & 7.91 {\scriptsize ($\pm$2.25)} & 9.40 {\scriptsize ($\pm$2.06)} & 1.49 {\scriptsize ($\pm$1.74)} & 1.60 & 11.00 & 1.00--11.00 & 1.00--11.00 \\
Motivation & 6.93 {\scriptsize ($\pm$1.46)} & 8.09 {\scriptsize ($\pm$1.01)} & 1.16 {\scriptsize ($\pm$1.17)} & 0.91 & 9.00 & 4.00--9.00 & 6.00--9.00 \\
Language/Quality & 2.69 {\scriptsize ($\pm$1.02)} & 3.45 {\scriptsize ($\pm$0.78)} & 0.76 {\scriptsize ($\pm$0.81)} & 0.55 & 4.00 & 0.00--4.00 & 1.00--4.00 \\
Methodology & 4.07 {\scriptsize ($\pm$1.32)} & 4.76 {\scriptsize ($\pm$0.57)} & 0.69 {\scriptsize ($\pm$1.08)} & 0.24 & 5.00 & 0.00--5.00 & 3.00--5.00 \\
Schedule & 4.15 {\scriptsize ($\pm$0.84)} & 4.65 {\scriptsize ($\pm$0.55)} & 0.51 {\scriptsize ($\pm$0.83)} & 0.35 & 5.00 & 0.00--5.00 & 3.00--5.00 \\
Bibliography & 2.56 {\scriptsize ($\pm$0.85)} & 2.91 {\scriptsize ($\pm$0.29)} & 0.35 {\scriptsize ($\pm$0.77)} & 0.09 & 3.00 & 0.00--3.00 & 2.00--3.00 \\
Metadata & 2.75 {\scriptsize ($\pm$0.51)} & 2.95 {\scriptsize ($\pm$0.23)} & 0.20 {\scriptsize ($\pm$0.48)} & 0.05 & 3.00 & 1.00--3.00 & 2.00--3.00 \\
Additional points & -0.02 {\scriptsize ($\pm$0.36)} & 0.00 {\scriptsize ($\pm$0.27)} & 0.02 {\scriptsize ($\pm$0.30)} & 2.00 & 2.00 & -2.00--1.00 & -1.00--1.00 \\
Form/Structure & 1.96 {\scriptsize ($\pm$0.19)} & 1.96 {\scriptsize ($\pm$0.19)} & 0.00 {\scriptsize ($\pm$0.19)} & 0.04 & 2.00 & 1.00--2.00 & 1.00--2.00 \\
\bottomrule
\end{tabular}
\caption{Per-rubric statistics for draft and final expos\'e scores. We report mean (±SD), average remaining gap to the rubric maximum, and observed score bounds (min--max) for each version.}
\label{tab:appendix_score_improvements}
\end{table*}

\subsection{Review Statistics}
\label{appendix:subsec:review-stats}
Table~\ref{tab:appendix_review_lexical} reports text statistics for review documents by author role (mean $\pm$ SD), computed analogously to the expos\'e statistics. Student reviews are substantially longer than instructor reviews, while instructors show higher lexical diversity.

\begin{table*}[ht]
\centering
\setlength{\tabcolsep}{4pt}
\renewcommand{\arraystretch}{1.05}

\newcolumntype{C}[1]{>{\centering\arraybackslash}p{#1}}
\begin{tabular}{l | C{0.20\textwidth} | C{0.20\textwidth} | C{0.20\textwidth}}
\toprule
 & \multicolumn{3}{c}{\textbf{Review author role}} \\
\textbf{Metric} & \textbf{Student} & \textbf{Junior Instructor} & \textbf{Senior Instructor} \\
\midrule

\multicolumn{4}{l}{\textbf{Length and structure}} \\
\cmidrule(lr){1-4}
\#Words & 680 ($\pm$455) & 285 ($\pm$134) & 347 ($\pm$116) \\
\#Sentences & 33.7 ($\pm$21.3) & 16.7 ($\pm$8.4) & 18.7 ($\pm$6.3) \\
Avg.\ word\ length & 5.31 ($\pm$0.44) & 5.16 ($\pm$0.28) & 5.18 ($\pm$0.49) \\

\midrule

\multicolumn{4}{l}{\textbf{Lexical and readability}} \\
\cmidrule(lr){1-4}
Lexical diversity & 0.46 ($\pm$0.08) & 0.56 ($\pm$0.09) & 0.55 ($\pm$0.08) \\
Flesch reading ease & 50.2 ($\pm$13.3) & 58.9 ($\pm$9.9) & 54.3 ($\pm$14.2) \\

\bottomrule
\end{tabular}
\caption{Textual statistics of review texts by author role. Entries show mean ($\pm$ SD).}
\label{tab:appendix_review_lexical}

\end{table*}

\paragraph{Instructor roles}
We distinguish between \emph{junior} and \emph{senior} instructors based on their academic stage and experience in reviewing and research. Junior instructors are recruited Master’s students who have completed a Bachelor’s thesis and thus have basic academic writing experience, whereas senior instructors are course staff (PhD researchers and a postdoc) with substantially more experience in academic peer review and supervision; see Appendix~\ref{appendix:iaa-setup} for details on rater composition and training.

\paragraph{Inline comment distributions}
Table~\ref{tab:appendix_comments_by_reviewer_role} summarizes comment volume and length by role: junior instructors contribute the most comments overall, while students and senior instructors write longer comments on average.
Table~\ref{tab:appendix_annotations} reports the distribution of span-level tags. Totals are similar but not identical across the two tables because span annotations are only recorded for comments that are attached to a specific text span; page-level comments have no associated span annotation.
Across roles, \emph{Weakness} is the most frequent tag, followed by \emph{Strength}, \emph{Other}, and \emph{Highlight}.

\begin{table}[ht]
\centering
\setlength{\tabcolsep}{4pt}
\begin{tabular}{l r r r}
\toprule
Role & Comments & Avg chars & Avg words \\
\midrule
St. (n=31) & 640 & 77.0 & 12.6 \\
Jr. (n=7) & 1121 & 53.8 & 9.1 \\
Sr. (n=5) & 492 & 68.8 & 12.0 \\
\midrule
\textbf{$\Sigma$} (n=43) & \textbf{2253} & \textbf{63.6} & \textbf{10.7} \\
\bottomrule
\end{tabular}
\caption{Basic statistics of comments by role, including total counts and average length in characters and words. St.=Students, Jr.=Junior Instructors, Sr.=Senior Instructors.}
\label{tab:appendix_comments_by_reviewer_role}
\end{table}
\begin{table*}[ht]
\centering
\setlength{\tabcolsep}{3.5pt}

\begin{tabular}{l l | r r r | r}
\toprule
\textbf{Tag} & \textbf{Metric} & \textbf{St. (N=31)} & \textbf{Jr. (N=7)} & \textbf{Sr. (N=5)} & \textbf{$\Sigma$} \\
\midrule

Highlight & Raw count & 90 & 28 & 58 & \textbf{176} \\
& \textcolor{black!60}{Avg. length} & \textcolor{black!60}{112.68 ($\pm$188.62)} & \textcolor{black!60}{122.79 ($\pm$149.17)} & \textcolor{black!60}{168.34 ($\pm$227.29)} & \textcolor{black!60}{132.63 ($\pm$198.48)} \\
\cmidrule(lr){1-6}
Other & Raw count & 107 & 145 & 52 & \textbf{304} \\
& \textcolor{black!60}{Avg. length} & \textcolor{black!60}{120.66 ($\pm$236.88)} & \textcolor{black!60}{41.77 ($\pm$61.91)} & \textcolor{black!60}{90.83 ($\pm$132.55)} & \textcolor{black!60}{77.93 ($\pm$160.87)} \\
\cmidrule(lr){1-6}
Strength & Raw count & 131 & 200 & 73 & \textbf{404} \\
& \textcolor{black!60}{Avg. length} & \textcolor{black!60}{254.63 ($\pm$263.11)} & \textcolor{black!60}{178.40 ($\pm$239.87)} & \textcolor{black!60}{196.52 ($\pm$285.21)} & \textcolor{black!60}{206.39 ($\pm$258.44)} \\
\cmidrule(lr){1-6}
Weakness & Raw count & 308 & 691 & 290 & \textbf{1289} \\
& \textcolor{black!60}{Avg. length} & \textcolor{black!60}{79.20 ($\pm$106.17)} & \textcolor{black!60}{59.17 ($\pm$127.02)} & \textcolor{black!60}{97.88 ($\pm$154.42)} & \textcolor{black!60}{72.66 ($\pm$130.22)} \\
\cmidrule(lr){1-6}
\textbf{$\Sigma$} & \textbf{Raw count} & \textbf{636} & \textbf{1064} & \textbf{473} & \textbf{2173} \\
\bottomrule
\end{tabular}
\caption{Distribution of span-level annotation tags by role. We report raw counts and average span length (±SD). St.=Students, Jr.=Junior Instructors, Sr.=Senior Instructors.}
\label{tab:appendix_annotations}
\end{table*}

\subsection{Assessment scores coverage}
\label{subsec:annotation-effort}

Table~\ref{tab:appendix_expose_grading} summarizes the number of assessments that include a complete set of criterion-based scores for exposé drafts and final submissions, by rater group. Draft exposés were graded by instructors during the semester (Group~1) and re-scored after the course by Group~2 to compute inter-annotator agreement (Section~\ref{sec:iaa}). Final exposés, in contrast, were not re-scored post-semester: due to the substantial time required for criterion-based grading, Group~2 focused on a draft subset rather than duplicating the full set of final-submission grades. A similar pattern holds for student review grading (Table~\ref{tab:appendix_review_grading}): reviews were graded during the semester and selectively re-scored for agreement analyses, but criterion-based scores are only available for student-written reviews (not for instructor-written reviews).

\begin{table}[ht]
\centering

\begin{tabular}{l | r r r | r r r}
\toprule
 & \multicolumn{3}{c|}{\textbf{Group 1}} & \multicolumn{3}{c}{\textbf{Group 2}} \\
\textbf{Version / Topic} & Jr. & Sr. & \textbf{$\Sigma$} & Jr. & Sr. & \textbf{$\Sigma$} \\
\midrule

\multicolumn{7}{l}{\textbf{Draft (N = 55)}} \\
\cmidrule(lr){1-7}
T1 (N=19) & 14 & 5 & \textbf{19} & 2 & 17 & \textbf{19} \\
T2 (N=18) & 14 & 4 & \textbf{18} & 4 & 14 & \textbf{18} \\
T3 (N=18) & 15 & 3 & \textbf{18} & 2 & 16 & \textbf{18} \\
\cmidrule(lr){1-7}
\textbf{$\Sigma$ Draft} & \textbf{43} & \textbf{12} & \textbf{55} & \textbf{8} & \textbf{47} & \textbf{55} \\

\midrule

\multicolumn{7}{l}{\textbf{Final (N = 55)}} \\
\cmidrule(lr){1-7}
T1 (N=19) & 15 & 4 & \textbf{19} & 0 & 0 & \textbf{0} \\
T2 (N=18) & 14 & 4 & \textbf{18} & 0 & 0 & \textbf{0} \\
T3 (N=18) & 15 & 3 & \textbf{18} & 0 & 0 & \textbf{0} \\
\cmidrule(lr){1-7}
\textbf{$\Sigma$ Final} & \textbf{44} & \textbf{11} & \textbf{55} & \textbf{0} & \textbf{0} & \textbf{0} \\

\bottomrule
\end{tabular}
\caption{Distribution of criterion-based expos\'e assessments across topics (T1--T3), rater roles (Jr.\ Junior Instructors, Sr.\ Senior Instructors), rater groups (Group~1 in-semester, Group~2 post-semester), and versions (draft vs.\ final). T1: ``The Holographic AI Assistant''; T2: ``Interactive LLM-based teaching agents''; T3: ``Generative AI in Healthcare''.}
\label{tab:appendix_expose_grading}

\end{table}
\begin{table}[ht]
\centering
\setlength{\tabcolsep}{5pt}

\begin{tabular}{l | r r r | r r r}
\toprule
 & \multicolumn{3}{c|}{\textbf{Group 1}} & \multicolumn{3}{c}{\textbf{Group 2}} \\
\textbf{Reviews} & Jr. & Sr. & \textbf{$\Sigma$} & Jr. & Sr. & \textbf{$\Sigma$} \\
\midrule

\multicolumn{7}{l}{\textbf{Student reviews (N = 110)}} \\
\cmidrule(lr){1-7}
$E$ (N = 34) & 28 & 6 & \textbf{34} & 0 & 59 & \textbf{59} \\
$\neg E$ (N = 76) & 67 & 9 & \textbf{76} & 0 & 111 & \textbf{111} \\
\cmidrule(lr){1-7}
\textbf{$\Sigma$} & \textbf{95} & \textbf{15} & \textbf{110} & \textbf{0} & \textbf{170} & \textbf{170} \\

\bottomrule
\end{tabular}
\caption{Distribution of criterion-based student review assessments by instructor role (Jr.\ Junior Instructors, Sr.\ Senior Instructors) and rater group (Group~1 vs.\ Group~2), stratified by whether the reviewed author has an associated expos\'e ($E$) or not ($\neg E$).}
\label{tab:appendix_review_grading}
\end{table}

\paragraph{Annotation effort}
We emphasize that constructing this dataset is unusually labor-intensive. In Group~1, producing a full review with inline comments and criterion-based scores typically required on the order of 2--4 hours early in the semester, decreasing to roughly 1--2 hours as raters gained experience. Criterion-based grading itself is also high effort: grading a single exposé typically takes about 1--2 hours, while grading a single review requires 20--40 minutes. Despite these restrictions, the dataset currently comprises 55 exposés and 306 reviews and includes unusually rich annotation artifacts: inline comments and comment tags, free-text reviews, draft and final exposé versions, and hundreds of instructor assessment scores across 36 \expose and 24 review criteria, which we use throughout our analyses.

\paragraph{Scoring and special rubric behavior}
\label{appendix:subsec:expose-revision-stats:points}
Within each submission (draft expos\'e, final expos\'e, peer feedback), criterion-level scores are summed to yield rubric points, and rubric points are summed to obtain total submission points.
Most rubrics use simple addition; we introduce a small number of rubric-specific rules to maintain consistency across topics and feedback situations.
The \expose{} rubric \emph{Methodology} is capped at 5 points, and the review rubric \emph{Feedback} is capped at 4 points. For example, in the \expose{} rubric, \emph{Existing Material} may be inapplicable when the proposed methodology relies on newly collected primary data rather than reusing an existing dataset; the cap ensures that full credit remains attainable by satisfying an appropriate subset of criteria.
The review rubric \emph{Errors} employs a deduction-only scheme, starting at 4 points and decreasing by 1 point for each identified error, with a minimum score of 0.
Finally, the instructor-only rubric \emph{Additional points} permits bounded adjustments in exceptional cases (Expos\'e: $[-2,+2]$, Review: $[-1,+1]$); these adjustments do not increase the maximum attainable points.

\begin{table}[ht]
\centering
\setlength{\tabcolsep}{2.8pt}
\begin{tabular}{l | r r r r r | r r r r}
\toprule
 & \multicolumn{5}{c|}{\textbf{Group 1}} & \multicolumn{4}{c}{\textbf{Group 2}} \\
 \textbf{\shortstack{Reviews\\per expos\'e}} & \textbf{N} & Jr. & Sr. & St. & \textbf{$\Sigma$} & \textbf{N} & Jr. & Sr. & \textbf{$\Sigma$} \\
\midrule

\multicolumn{10}{l}{\textbf{Expos\'es (total N = 55)}} \\
\cmidrule(lr){1-10}
 0 review(s) & \textbf{0} & 0 & 0 & 0 & \textbf{0} & \textbf{30} & 0 & 0 & \textbf{0} \\
 1 review(s) & \textbf{23} & 17 & 6 & 0 & \textbf{23} & \textbf{24} & 4 & 20 & \textbf{24} \\
 2 review(s) & \textbf{23} & 21 & 7 & 18 & \textbf{46} & \textbf{1} & 0 & 2 & \textbf{2} \\
 3 review(s) & \textbf{9} & 8 & 3 & 16 & \textbf{27} & \textbf{0} & 0 & 0 & \textbf{0} \\
\cmidrule(lr){1-10}
 \textbf{$\Sigma$ totals} & \textbf{55} & \textbf{46} & \textbf{16} & \textbf{34} & \textbf{96} & \textbf{55} & \textbf{4} & \textbf{22} & \textbf{26} \\

\midrule

\multicolumn{10}{l}{\textbf{Orphaned reviews (total N = 184)}} \\
\cmidrule(lr){1-10}
 $\Sigma$ totals & \textbf{184} & 88 & 20 & 76 & \textbf{184} & \textbf{0} & 0 & 0 & \textbf{0} \\
\bottomrule
\end{tabular}

\caption{Distribution of reviews per expos\'e by review group (Group~1 vs.\ Group~2). In the Expos\'es block, \textbf{N} is the number of expos\'es in the bucket; role columns (Jr./Sr./St.) are counts of reviews written by that role (columns that are all-zero for a group are omitted); and $\Sigma$ is the total number of reviews. Orphaned reviews are reported separately.}
\label{tab:appendix:reviews_per_expose}
\end{table}

\section{Inter-annotator agreement}
\label{appendix:iaa}
This section documents the computation of inter-annotator agreement (IAA), complementing the criterion-based scoring description in Section~\ref{sec:iaa} for \nameDataset{}. It also provides the reference point for interpreting LLM--human agreement in the main experiments. Because our criterion scores are discrete and ordinal, we prioritize agreement measures that account for ordered categories and report distributional diagnostics for criteria affected by prevalence skew.

\subsection{Raters, training, and scoring design}
\label{appendix:iaa-setup}

\paragraph{Raters and training}

Twelve raters participated and were organized into two groups. In total, 7 were \emph{junior instructors} (student helpers) and 5 were \emph{senior instructors} (4 PhD researchers and 1 postdoc). One PhD researcher participated in both groups. \\

We refer to instructors who graded during the semester as  \emph{Group 1}. They attended weekly sessions and discussed edge cases. Group 1 raters were instructed to resolve uncertainty in favor of the student, because grades were at stake. \emph{Group 2} includes a different subset of instructors (with one who also participated in Group 1). They received the same guidelines, but they were not instructed to resolve uncertainty in favor of the student. 

\textbf{Group 1 (in-semester):} 6 student helpers, 2 PhD researchers, 1 postdoc; weekly calibration (jointly graded initial exposés, then addressed edge cases). Because grades were at stake, Group~1 was instructed to resolve uncertainty in favor of the student. \\
\textbf{Group 2 (post-semester):} 1 student helper, 3 PhD researchers; scored subsets of draft exposés and reviews using a recorded calibration video prepared by a Group-1 senior instructor.

\paragraph{Rater groups and coverage}
Table~\ref{tab:appendix_expose_grading} indicates that Group~1 grades are predominantly assigned by junior instructors, whereas Group~2 grades are predominantly assigned by senior instructors. This shift mainly reflects post-semester availability: once the course ended, much of the teaching staff was no longer available, so re-grading was carried out primarily by senior instructors; final submissions were graded only during the semester.

\paragraph{Scope of IAA}
Our agreement analyses are necessarily restricted to items (\expose{}s and reviews) that have assessments from two instructors. For exposés, we report agreement only for \emph{draft} submissions, as final submissions were not scored post-semester due to the substantial time required for criterion-based assessment (Appendix~\ref{subsec:annotation-effort}). For reviews, we focus on the subset that has criterion-based grades from Group~1 (student reviews). Finally, for a subset of reviews ($n=60$), we collected duplicate retrospective ratings from Group~2, enabling a within-group agreement comparison in addition to cross-group agreement (Table~\ref{tab:iia-review_criteria_inner_group2}). Beyond criterion-level results, we also report aggregate agreement (i) by rubric domain and (ii) by criterion expertise level (see Appendix~\ref{appendix:criteria}).

\begin{table}[t]
\centering
\setlength{\tabcolsep}{4.2pt}
\renewcommand{\arraystretch}{1.08}
\begin{tabular}{l | c c | c c}
\toprule
& \multicolumn{2}{c|}{\textbf{Exposé} (\#31)} & \multicolumn{2}{c}{\textbf{Review} (\#22)} \\
\textbf{Rater / Model} & All & Ind. & All & Ind. \\
\midrule
Human raters & \multicolumn{2}{c|}{0.76} & \multicolumn{2}{c}{0.88} \\
\midrule
Claude Opus 4.6    & \textbf{0.80} & \textbf{0.79} & \textbf{0.90} & \textbf{0.85} \\
Gemini 3.1 Pro     & 0.78          & \textbf{0.79} & \textbf{0.90} & \textbf{0.89} \\
GPT-5.2            & \textbf{0.80} & 0.78          & 0.87          & 0.79 \\
GPT OSS 120B       & 0.77          & 0.75          & 0.78          & 0.72 \\
GPT OSS 20B        & 0.77          & 0.76          & 0.74          & 0.69 \\
Llama 3.3 70B      & 0.76          & 0.75          & \textbf{0.90} & 0.82 \\
Qwen3 80B          & 0.72          & 0.75          & 0.89          & 0.84 \\
\bottomrule
\end{tabular}
\caption{Quadratic weighted agreement (QWA) for the two tasks, reported over all criteria only. Prompting variants: \emph{all} (combined) and \emph{ind.} (individual). Each LLM value shows human--LLM QWA averaged over both human raters and all criteria. The number of criteria are shown in the title (for exposés and reviews respectively). Bold indicates the best model performance per column.}
\label{tab:qwa-expert-allmodels}
\end{table}

\begin{table*}[t]
\centering
\small
\setlength{\tabcolsep}{5pt}
\renewcommand{\arraystretch}{1.08}
\newcolumntype{Y}{>{\raggedright\arraybackslash}X}
\begin{tabularx}{\textwidth}{l Y Y r r r r}
\toprule
 Code & Criterion & Distribution (\%) & QWA & $\alpha$ & $r$ & $\Delta\mu$ \\
\midrule\midrule
 LA1 & Language quality & \mbox{0 (15), 1 (55), 2 (30)} & 0.82 & 0.15 & 0.17 & 0.13 \\
 LA2 & Common Thread & \mbox{0 (4), 1 (65), 2 (32)} & 0.87 & 0.05 & 0.20 & 0.38 \\
\midrule
 \multicolumn{2}{l}{\textbf{Average~Language}} &  & 0.85 & 0.10 & 0.19 & 0.25 \\
\midrule
 ME1 & Preliminary title & \mbox{0 (3), 1 (42), 2 (55)} & 0.86 & 0.10 & 0.31 & 0.44 \\
 ME2$^{\ast}$ & Metadata available & \mbox{0 (10), 1 (90)} & 0.80 & -0.10 & — & 0.20 \\
\midrule
 \multicolumn{2}{l}{\textbf{Average~Metadata}} &  & 0.83 & 0.00 & 0.31 & 0.32 \\
\midrule
 ST1$^{\ast}$ & Template used & \mbox{0 (2), 1 (98)} & \textbf{0.96} & -0.01 & -0.02 & 0.00 \\
 ST2$^{\ast}$ & Number of pages & \mbox{0 (3), 1 (97)} & \textbf{0.98} & 0.66 & 0.70 & 0.02 \\
\midrule
 \multicolumn{2}{l}{\textbf{Average~Form/Structure}} &  & 0.97 & 0.33 & 0.34 & 0.01 \\
\midrule
 MO1 & Hook existing & \mbox{0 (15), 1 (85)} & 0.76 & 0.10 & 0.10 & 0.05 \\
 MO2 & Anker & \mbox{0 (5), 1 (63), 2 (33)} & 0.85 & -0.00 & 0.06 & 0.31 \\
 MO3$^{\ast}$ & Domain & \mbox{0 (5), 1 (95)} & 0.95 & 0.38 & 0.49 & 0.05 \\
 MO4 & Problem relevance & \mbox{0 (17), 1 (83)} & 0.65 & -0.20 & -0.21 & -0.02 \\
 MO5 & Problem handling & \mbox{0 (35), 1 (65)} & 0.64 & 0.20 & 0.24 & 0.18 \\
 MO6 & Teaser (RQ) & \mbox{0 (7), 1 (56), 2 (36)} & 0.85 & 0.15 & 0.21 & 0.22 \\
 MO7 & RQ Limitation & \mbox{0 (46), 1 (54)} & 0.58 & 0.17 & 0.25 & 0.27 \\
\midrule
 \multicolumn{2}{l}{\textbf{Average~Motivation}} &  & 0.76 & 0.11 & 0.16 & 0.15 \\
\midrule
 AP1 & SOTA & \mbox{0 (13), 1 (87)} & 0.85 & 0.35 & 0.37 & 0.07 \\
 AP2 & SOTA Relevance & \mbox{0 (9), 1 (56), 2 (35)} & 0.86 & 0.28 & 0.39 & 0.36 \\
 AP3 & SOTA Weaknesses & \mbox{0 (23), 1 (55), 2 (22)} & 0.83 & 0.25 & 0.30 & 0.27 \\
 AP4 & SOTA Delimitation & \mbox{0 (30), 1 (70)} & 0.73 & 0.36 & 0.40 & 0.16 \\
 AP5 & SOTA Combination & \mbox{0 (34), 1 (66)} & 0.65 & 0.23 & 0.29 & 0.20 \\
 AP6 & Theoretical Framework & \mbox{0 (20), 1 (47), 2 (33)} & 0.78 & 0.15 & 0.20 & 0.29 \\
 AP7 & Relevance of theoretical framework & \mbox{0 (35), 1 (65)} & 0.64 & 0.20 & 0.20 & 0.00 \\
 AP8$^{\ast}$ & Methodology Availability & \mbox{0 (9), 1 (91)} & 0.82 & -0.09 & -0.08 & 0.11 \\
\midrule
 \multicolumn{2}{l}{\textbf{Average~Approach}} &  & 0.77 & 0.22 & 0.26 & 0.18 \\
\midrule
 MD1 & Methodology Completeness & \mbox{0 (45), 1 (55)} & 0.69 & 0.38 & 0.46 & 0.24 \\
 MD2 & Methodology Relevance & \mbox{0 (26), 1 (74)} & 0.73 & 0.30 & 0.30 & 0.05 \\
 MD3 & Methodology Target group & \mbox{0 (25), 1 (75)} & 0.73 & 0.27 & 0.30 & 0.13 \\
 MD4 & Methodology Existing Material & \mbox{0 (35), 1 (65)} & 0.55 & 0.02 & 0.03 & -0.13 \\
 MD5 & Methodology Difficulties & \mbox{0 (44), 1 (56)} & 0.56 & 0.12 & 0.19 & 0.25 \\
 MD6 & Methodology Possibilities/restrictions & \mbox{0 (56), 1 (44)} & 0.60 & 0.19 & 0.27 & 0.25 \\
 MD7 & Methodology Details & \mbox{0 (53), 1 (47)} & 0.49 & -0.01 & 0.01 & 0.18 \\
\midrule
 \multicolumn{2}{l}{\textbf{Average~Methodology}} &  & 0.62 & 0.18 & 0.22 & 0.14 \\
\midrule
 SC1$^{\ast}$ & Schedule Availability & \mbox{0 (4), 1 (96)} & 0.96 & 0.49 & 0.57 & 0.04 \\
 SC2$^{\ast}$ & Schedule Completeness & \mbox{0 (4), 1 (96)} & 0.96 & 0.49 & 0.57 & 0.04 \\
 SC3$^{\ast}$ & Schedule Block Description & \mbox{0 (6), 1 (94)} & 0.91 & 0.24 & 0.28 & -0.05 \\
 SC4 & Schedule Realistic Relevance & \mbox{0 (7), 1 (71), 2 (22)} & 0.90 & 0.21 & 0.36 & 0.25 \\
\midrule
 \multicolumn{2}{l}{\textbf{Average~Schedule}} &  & 0.93 & 0.36 & 0.44 & 0.07 \\
\midrule
 BI1 & Bibliography Consistency & \mbox{0 (19), 1 (81)} & 0.80 & 0.36 & 0.49 & 0.20 \\
 BI2 & Key literature & \mbox{0 (15), 1 (35), 2 (49)} & 0.76 & 0.03 & 0.35 & 0.64 \\
\midrule
 \multicolumn{2}{l}{\textbf{Average~Bibliography}} &  & 0.78 & 0.19 & 0.42 & 0.42 \\
\midrule
 AD1$^{\ast}$ & Additional points & \mbox{0 (97), 1 (2), 2 (1)} & \textbf{0.99} & 0.67 & 0.70 & 0.00 \\
 AD2$^{\ast}$ & Negative points & \mbox{-2 (2), -1 (1), 0 (97)} & 0.96 & -0.02 & -0.03 & -0.02 \\
\midrule
 \multicolumn{2}{l}{\textbf{Average~Additional points}} &  & 0.98 & 0.32 & 0.34 & -0.01 \\
\midrule
\midrule
 \multicolumn{2}{l}{\textbf{Average~all criteria}} &  & 0.79 & 0.20 & 0.27 & 0.16 \\
\midrule
\bottomrule
\end{tabularx}
\caption{Quadratic weighted agreement (QWA) between Group~1 and Group~2 raters for each expos\'e criterion, with rubric-wise and overall averages. Columns show agreement (Agr), reliability ($\alpha$), correlation ($r$), and mean bias ($\Delta\mu$). Rows marked with $^{\ast}$ have a highly skewed class distribution (\(\max\_c p(c) \ge 90\%\)).}
\label{tab:iia-expose-criteria}
\end{table*}
\begin{table*}[t]
\centering
\small
\setlength{\tabcolsep}{5pt}
\renewcommand{\arraystretch}{1.08}
\newcolumntype{Y}{>{\raggedright\arraybackslash}X}
\begin{tabularx}{\textwidth}{l Y Y r r r r}
\toprule
 Code & Criterion & Distribution (\%) & QWA & $\alpha$ & $r$ & $\Delta\mu$ \\
\midrule\midrule
 SC1 & Summary available & \mbox{0 (14), 1 (30), 2 (56)} & 0.90 & 0.49 & 0.61 & 0.11 \\
 SC2 & Clarity & \mbox{0 (10), 1 (38), 2 (53)} & 0.82 & 0.04 & 0.15 & 0.38 \\
\midrule
 \multicolumn{2}{l}{\textbf{Average~Structure and Clarity}} &  & 0.86 & 0.27 & 0.38 & 0.25 \\
\midrule
 LG1$^{\ast}$ & Language & \mbox{0 (6), 1 (94)} & 0.88 & -0.06 & -0.06 & -0.07 \\
 LG2$^{\ast}$ & Tone & \mbox{0 (1), 1 (7), 2 (92)} & 0.98 & 0.23 & 0.39 & -0.06 \\
\midrule
 \multicolumn{2}{l}{\textbf{Average~Language}} &  & 0.93 & 0.08 & 0.17 & -0.06 \\
\midrule
 FU1 & Learning Goals & \mbox{0 (29), 1 (28), 2 (42)} & 0.83 & 0.51 & 0.50 & -0.08 \\
 FU2 & Success Criteria & \mbox{0 (32), 1 (31), 2 (37)} & 0.79 & 0.35 & 0.36 & 0.05 \\
\midrule
 \multicolumn{2}{l}{\textbf{Average~Feed Up}} &  & 0.81 & 0.43 & 0.43 & -0.01 \\
\midrule
 FB1$^{\ast}$ & Knowledge of result & \mbox{0 (5), 1 (95)} & 0.95 & 0.26 & 0.40 & -0.05 \\
 FB2$^{\ast}$ & Knowledge of correct response & \mbox{0 (8), 1 (92)} & 0.90 & -0.05 & -0.05 & -0.05 \\
 FB3 & Knowledge about task constraints & \mbox{0 (33), 1 (67)} & 0.65 & 0.07 & 0.08 & 0.07 \\
 FB4 & Knowledge about concepts & \mbox{0 (13), 1 (87)} & 0.82 & -0.00 & 0.05 & -0.12 \\
 FB5$^{\ast}$ & Knowledge about mistakes & \mbox{0 (8), 1 (92)} & 0.90 & -0.05 & -0.05 & 0.01 \\
 FB6$^{\ast}$ & Self-feedback & \mbox{0 (3), 1 (97)} & 0.97 & -0.01 & — & 0.03 \\
\midrule
 \multicolumn{2}{l}{\textbf{Average~Feed Back}} &  & 0.86 & 0.04 & 0.08 & -0.02 \\
\midrule
 FF1$^{\ast}$ & Self-regulation & \mbox{0 (6), 1 (94)} & 0.95 & -0.02 & -0.02 & -0.03 \\
 FF2 & Learning Skills & \mbox{0 (39), 1 (61)} & 0.60 & 0.06 & 0.10 & 0.17 \\
\midrule
 \multicolumn{2}{l}{\textbf{Average~Feed Forward}} &  & 0.78 & 0.02 & 0.04 & 0.07 \\
\midrule
 CQ1 & Correctness & \mbox{0 (1), 1 (28), 2 (71)} & 0.91 & 0.14 & 0.16 & 0.12 \\
 CQ2 & Actionability & \mbox{0 (1), 1 (32), 2 (66)} & 0.92 & 0.22 & 0.21 & 0.08 \\
 CQ3$^{\ast}$ & Argumentation & \mbox{0 (10), 1 (90), 2 (0)} & \textbf{0.98} & 0.30 & 0.36 & -0.06 \\
\midrule
 \multicolumn{2}{l}{\textbf{Average~Content Quality}} &  & 0.94 & 0.22 & 0.24 & 0.05 \\
\midrule
 ER1 & Neglect & \mbox{-1 (16), 0 (84)} & 0.86 & 0.40 & 0.41 & -0.05 \\
 ER2$^{\ast}$ & Vague Critic & \mbox{-1 (7), 0 (93)} & 0.91 & 0.12 & 0.12 & 0.00 \\
 ER3$^{\ast}$ & Out-of-Scope & \mbox{-1 (4), 0 (96)} & 0.93 & -0.03 & -0.03 & -0.01 \\
 ER4$^{\ast}$ & Missing Reference & \mbox{-1 (2), 0 (98)} & 0.96 & -0.01 & — & -0.04 \\
 ER5$^{\ast}$ & Contradiction & \mbox{-1 (2), 0 (98)} & \textbf{0.98} & -0.00 & -0.01 & 0.00 \\
\midrule
 \multicolumn{2}{l}{\textbf{Average~Errors}} &  & 0.93 & 0.10 & 0.12 & -0.02 \\
\midrule
 AD1$^{\ast}$ & Additional points & \mbox{0 (100), 1 (0)} & \textbf{1.00} & — & — & 0.00 \\
 AD2$^{\ast}$ & Negative points & \mbox{-1 (2), 0 (98)} & 0.96 & -0.01 & — & -0.04 \\
\midrule
 \multicolumn{2}{l}{\textbf{Average~Additional points}} &  & 0.98 & -0.01 & — & -0.02 \\
\midrule
\midrule
 \multicolumn{2}{l}{\textbf{Average~all criteria}} &  & 0.89 & 0.13 & 0.18 & 0.02 \\
\midrule
\bottomrule
\end{tabularx}
\caption{Quadratic weighted agreement (QWA) between review Group~1 and Group~2 raters for each review criterion, with rubric-wise and overall averages. Columns show agreement (Agr), reliability ($\alpha$), correlation ($r$), and mean bias ($\Delta\mu$). Rows marked with $^{\ast}$ have a highly skewed class distribution (\(\max\_c p(c) \ge 90\%\)).}
\label{tab:iia-review_criteria}
\end{table*}
\begin{table*}[t]
\centering
\small
\setlength{\tabcolsep}{5pt}
\renewcommand{\arraystretch}{1.08}
\newcolumntype{Y}{>{\raggedright\arraybackslash}X}
\begin{tabularx}{\textwidth}{l Y Y r r r r}
\toprule
 Code & Criterion & Distribution (\%) & QWA & $\alpha$ & $r$ & $\Delta\mu$ \\
\midrule\midrule
 SC1 & Summary available & \mbox{0 (13), 1 (43), 2 (43)} & 0.91 & 0.60 & 0.62 & 0.00 \\
 SC2 & Clarity & \mbox{0 (12), 1 (48), 2 (40)} & 0.88 & 0.47 & 0.45 & 0.17 \\
\midrule
 \multicolumn{2}{l}{\textbf{Average~Structure and Clarity}} &  & 0.89 & 0.53 & 0.54 & 0.08 \\
\midrule
 LG1$^{\ast}$ & Language & \mbox{0 (4), 1 (96)} & 0.95 & 0.38 & 0.38 & 0.02 \\
 LG2 & Tone & \mbox{0 (1), 1 (12), 2 (88)} & 0.93 & 0.02 & 0.00 & 0.07 \\
\midrule
 \multicolumn{2}{l}{\textbf{Average~Language}} &  & 0.94 & 0.20 & 0.19 & 0.04 \\
\midrule
 FU1 & Learning Goals & \mbox{0 (28), 1 (28), 2 (44)} & 0.75 & 0.31 & 0.33 & 0.25 \\
 FU2 & Success Criteria & \mbox{0 (34), 1 (35), 2 (31)} & 0.72 & 0.13 & 0.15 & 0.23 \\
\midrule
 \multicolumn{2}{l}{\textbf{Average~Feed Up}} &  & 0.74 & 0.22 & 0.24 & 0.24 \\
\midrule
 FB1$^{\ast}$ & Knowledge of result & \mbox{0 (6), 1 (94)} & 0.88 & -0.05 & -0.06 & -0.02 \\
 FB2 & Knowledge of correct response & \mbox{0 (11), 1 (89)} & 0.85 & 0.23 & 0.22 & -0.02 \\
 FB3 & Knowledge about task constraints & \mbox{0 (37), 1 (63)} & 0.53 & 0.00 & 0.03 & 0.17 \\
 FB4 & Knowledge about concepts & \mbox{0 (13), 1 (87)} & 0.77 & -0.00 & -0.01 & 0.03 \\
 FB5$^{\ast}$ & Knowledge about mistakes & \mbox{0 (10), 1 (90)} & 0.87 & 0.27 & 0.27 & 0.03 \\
 FB6$^{\ast}$ & Self-feedback & \mbox{0 (4), 1 (96)} & 0.92 & -0.03 & -0.04 & 0.02 \\
\midrule
 \multicolumn{2}{l}{\textbf{Average~Feed Back}} &  & 0.80 & 0.07 & 0.07 & 0.04 \\
\midrule
 FF1$^{\ast}$ & Self-regulation & \mbox{0 (9), 1 (91)} & 0.82 & -0.09 & -0.10 & -0.02 \\
 FF2 & Learning Skills & \mbox{0 (48), 1 (52)} & 0.57 & 0.14 & 0.14 & 0.10 \\
\midrule
 \multicolumn{2}{l}{\textbf{Average~Feed Forward}} &  & 0.69 & 0.02 & 0.02 & 0.04 \\
\midrule
 CQ1 & Correctness & \mbox{0 (0), 1 (39), 2 (61)} & 0.91 & 0.27 & 0.27 & 0.02 \\
 CQ2 & Actionability & \mbox{0 (2), 1 (34), 2 (63)} & 0.89 & 0.23 & 0.22 & 0.05 \\
 CQ3 & Argumentation & \mbox{0 (12), 1 (88), 2 (0)} & 0.94 & -0.12 & -0.13 & 0.00 \\
\midrule
 \multicolumn{2}{l}{\textbf{Average~Content Quality}} &  & 0.91 & 0.13 & 0.12 & 0.02 \\
\midrule
 ER1 & Neglect & \mbox{-1 (15), 0 (85)} & 0.80 & 0.22 & 0.22 & -0.03 \\
 ER2$^{\ast}$ & Vague Critic & \mbox{-1 (5), 0 (95)} & 0.90 & -0.04 & -0.05 & 0.00 \\
 ER3$^{\ast}$ & Out-of-Scope & \mbox{-1 (2), 0 (98)} & 0.95 & -0.02 & -0.02 & -0.02 \\
 ER4$^{\ast}$ & Missing Reference & \mbox{-1 (1), 0 (99)} & \textbf{0.98} & 0.00 & — & 0.02 \\
 ER5$^{\ast}$ & Contradiction & \mbox{-1 (2), 0 (98)} & 0.95 & -0.02 & -0.02 & -0.02 \\
\midrule
 \multicolumn{2}{l}{\textbf{Average~Errors}} &  & 0.92 & 0.03 & 0.03 & -0.01 \\
\midrule
 AD1$^{\ast}$ & Additional points & \mbox{0 (100), 1 (0)} & \textbf{1.00} & — & — & 0.00 \\
 AD2$^{\ast}$ & Negative points & \mbox{-1 (1), 0 (99)} & \textbf{0.98} & 0.00 & — & 0.02 \\
\midrule
 \multicolumn{2}{l}{\textbf{Average~Additional points}} &  & 0.99 & 0.00 & — & 0.01 \\
\midrule
\midrule
 \multicolumn{2}{l}{\textbf{Average~all criteria}} &  & 0.86 & 0.13 & 0.14 & 0.04 \\
\midrule
\bottomrule
\end{tabularx}
\caption{Quadratic weighted agreement (QWA) within Group 2 raters for each review criterion, with rubric-wise and overall averages. Columns show agreement (Agr), reliability ($\alpha$), correlation ($r$), and mean bias ($\Delta\mu$). Rows marked with $^{\ast}$ have a highly skewed class distribution (\(\max\_c p(c) \ge 90\%\)).}
\label{tab:iia-review_criteria_inner_group2}
\end{table*}

\subsection{Human--human agreement}
\label{appendix:iaa-human}

\paragraph*{\textbf{Agreement reporting}}
We report raw quadratic weighted agreement (QWA), Krippendorff’s $\alpha$, Pearson correlation coefficient $r$, and simple mean difference (``mean bias'') between groups.
\textbf{QWA} is the observed agreement between two raters on an ordinal scale, where disagreements are weighted by squared distance so that larger score differences are penalized more strongly than adjacent-category mismatches.
Let $x_i,y_i$ be two raters’ scores for item $i$ on an ordinal scale with discrete domain
$\mathcal D=\{d_{\min},\dots,d_{\max}\}$ and span $\Delta=d_{\max}-d_{\min}>0$. We use the quadratic weight
\[
w(x_i,y_i)=\max\!\left(0,\,1-\left(\frac{x_i-y_i}{\Delta}\right)^2\right),
\]
and define raw quadratic weighted agreement as
\[
\mathrm{QWA}=\frac{1}{N}\sum_{i=1}^{N} w(x_i,y_i),
\]
where $N$ is the number of co-rated items.

We use quadratic weighting because it preserves the magnitude of disagreement and remains interpretable under strong prevalence skew, a setting in which chance-corrected coefficients (e.g., $\kappa$ or $\alpha$) can be misleading. 
Accordingly, for highly skewed criteria (e.g., \textit{Number of pages}, \textit{Template used}, \textit{Negative points}), we emphasize QWA.

$\alpha$ values range between -1 and 1, with 1 being perfect agreement, 0 being agreement by chance, and negative values indicating systematic disagreement; $r$ measures the correlation between a system’s predicted
scores and the annotator-assigned scores; it
ranges from -1 to 1. A positive (negative) value
implies that the two sets of predictions are positively (negatively) correlated. 
Inter-annotator agreement results for exposé criteria are reported in Table~\ref{tab:iia-expose-criteria}, while agreement for review criteria is summarized in Tables~\ref{tab:iia-review_criteria} and~\ref{tab:iia-review_criteria_inner_group2}.

\subsection{Failure modes of chance-corrected agreement}
\label{subsec:failure}

This section interprets the agreement patterns observed in the exposé and review analyses (see Section~\ref{sec:iaa} in the main paper). In particular, we focus on cases where chance-corrected reliability or correlation is negative, undefined, or unexpectedly low, despite high raw agreement, and relate these outcomes to prevalence effects, calibration differences between rater groups, and ambiguities in rubrics or training.

\paragraph{Prevalence effects under extreme class imbalance (``rare-event'' criteria)}
For highly skewed binary (or near-binary) criteria, the expected agreement under a chance model is already very high, so a handful of rare, non-overlapping deviations can yield $\alpha<0$ even when raw agreement is high. This explains the slightly negative values for \textit{Template used} (ST1) and \textit{Negative points} (AD2), where almost all submissions receive the default label, but the few exceptional ratings occur on different submissions. In such settings, Pearson’s $r$ is typically close to zero because the rating vectors are nearly constant, and mean bias $\Delta\mu$ helps diagnose whether the deviations are symmetric (ST1: $\Delta\mu \approx 0$) or slightly one-sided (AD2: small negative $\Delta\mu$ consistent with marginally more penalizing scores in one group).

\paragraph{Annotation guideline/interpretation issues (training or rubric ambiguity)}
Some negative/undefined reliability values are attributable to concrete, identifiable rubric failures. For \textit{Metadata available} (ME2), metadata was present for all submissions; all disagreements originated from a single rater in Group~2, indicating a training/interpretation error (i.e., the rater did not apply the intended check). This also leads to an undefined Pearson correlation because one group’s ratings are effectively constant (zero variance). For \textit{Template used} (ST1), the few ``0'' assignments were triggered by superficial deviations (e.g., renamed chapter headings or bibliography compilation issues) despite the underlying template being used, revealing an unclear rubric description about what counts as a template violation. For \textit{Methodology Availability} (AP8), we found edge cases where the methodology is present only as a very short list; these items were inconsistently judged as available vs.\ unavailable across groups, suggesting that the criterion definition is underspecified (i.e., availability vs.\ sufficiency/quality).

\paragraph{Systematic calibration shifts and low signal in subjective criteria}
For \textit{Anker} (MO2), the negative (near-zero) $\alpha$ is consistent with a systematic calibration shift between groups: Group~2 assigns lower scores on average (positive mean bias $\Delta\mu=0.309$), and disagreements are predominantly adjacent-category mismatches (e.g., $2$ vs.\ $1$), which remain relatively benign under quadratic weighting but still depress chance-corrected reliability. For \textit{Problem relevance} (MO4), negative coefficients are driven by sparse and subjective ``0'' decisions: the negative label is rare and tends to be applied to different submissions across groups.

\paragraph{Prevalence effects under extreme class imbalance in review scoring}
A closely related prevalence-driven failure mode appears in review scoring for error and penalty criteria (Tables~\ref{tab:iia-review_criteria} and~\ref{tab:iia-review_criteria_inner_group2}). These criteria are effectively \emph{rare-event detection} tasks on a highly skewed label space: the non-penalizing default (typically $0$) dominates, while penalizing assignments (e.g., $-1$ in \textit{Errors (ER)}) occur only rarely. As a result, chance-corrected reliability and Pearson correlation become unstable or uninformative. A small number of non-overlapping penalizations can yield $\alpha \le 0$ despite high raw agreement, and near-constant rating vectors drive $r$ toward zero or render it undefined when one rater shows zero variance.

Overall, we emphasize quadratic-weighted raw agreement as our primary agreement signal and use $\alpha$, $r$, $\Delta\mu$, and the pooled distributions to (i) flag prevalence-induced artifacts, (ii) diagnose calibration differences, and (iii) identify rubric/training issues that can be addressed in the future.

\begin{table*}[t]
\centering
\small
\setlength{\tabcolsep}{4.3pt}
\renewcommand{\arraystretch}{1.08}
\begin{tabular}{l | r r r r | r r r r | r r r r | r r r r}
\toprule
 & \multicolumn{4}{c}{Qwen3 (80B)} & \multicolumn{4}{c}{Llama 3.3 (70B)} & \multicolumn{4}{c}{GPT OSS (120B)} & \multicolumn{4}{c}{GPT OSS (20B)} \\
Code & $A^{\text{all}}_{\text{G1}}$ & $A^{\text{ind}}_{\text{G1}}$ & $A^{\text{all}}_{\text{G2}}$ & $A^{\text{ind}}_{\text{G2}}$ & $A^{\text{all}}_{\text{G1}}$ & $A^{\text{ind}}_{\text{G1}}$ & $A^{\text{all}}_{\text{G2}}$ & $A^{\text{ind}}_{\text{G2}}$ & $A^{\text{all}}_{\text{G1}}$ & $A^{\text{ind}}_{\text{G1}}$ & $A^{\text{all}}_{\text{G2}}$ & $A^{\text{ind}}_{\text{G2}}$ & $A^{\text{all}}_{\text{G1}}$ & $A^{\text{ind}}_{\text{G1}}$ & $A^{\text{all}}_{\text{G2}}$ & $A^{\text{ind}}_{\text{G2}}$ \\
\midrule
LA1 & 0.83 & 0.85 & 0.85 & 0.91 & 0.78 & 0.84 & 0.77 & 0.83 & 0.68 & \underline{0.59} & 0.73 & 0.67 & 0.72 & \underline{0.57} & 0.77 & 0.64 \\
LA2 & 0.87 & 0.88 & 0.76 & 0.85 & 0.87 & 0.87 & 0.75 & 0.75 & 0.88 & 0.90 & 0.89 & 0.86 & 0.82 & 0.87 & 0.91 & 0.81 \\
\midrule
\textbf{Language} & 0.85 & 0.87 & 0.80 & 0.88 & 0.82 & 0.85 & 0.76 & 0.79 & 0.78 & 0.75 & 0.81 & 0.76 & 0.77 & 0.72 & 0.84 & 0.73 \\
\midrule
ME1 & 0.91 & 0.92 & 0.82 & 0.86 & 0.91 & 0.90 & 0.84 & 0.87 & 0.91 & 0.84 & 0.87 & 0.84 & 0.82 & 0.81 & 0.90 & 0.86 \\
\midrule
\textbf{Metadata} & 0.91 & 0.92 & 0.82 & 0.86 & 0.91 & 0.90 & 0.84 & 0.87 & 0.91 & 0.84 & 0.87 & 0.84 & 0.82 & 0.81 & 0.90 & 0.86 \\
\midrule
MO1 & 0.87 & 0.87 & 0.82 & 0.82 & 0.87 & 0.84 & 0.82 & 0.82 & 0.84 & 0.73 & 0.78 & 0.60 & 0.84 & 0.71 & 0.78 & 0.69 \\
MO2 & 0.86 & 0.88 & 0.74 & 0.76 & 0.86 & 0.86 & 0.74 & 0.74 & 0.92 & 0.87 & 0.82 & 0.76 & 0.90 & 0.86 & 0.85 & 0.74 \\
MO3 & \textbf{0.96} & \textbf{0.98} & \textbf{0.91} & \textbf{0.93} & \textbf{0.98} & \textbf{0.98} & 0.93 & \textbf{0.93} & \textbf{0.98} & \textbf{0.98} & \textbf{0.93} & \textbf{0.93} & \textbf{0.98} & \textbf{0.98} & \textbf{0.93} & \textbf{0.93} \\
MO4 & \underline{0.38} & 0.82 & 0.42 & 0.84 & 0.75 & 0.82 & 0.76 & 0.84 & 0.82 & 0.82 & 0.84 & 0.84 & 0.82 & 0.82 & 0.84 & 0.84 \\
MO5 & 0.75 & 0.75 & 0.56 & 0.56 & 0.75 & 0.75 & 0.56 & 0.56 & 0.76 & 0.75 & 0.55 & \underline{0.56} & 0.73 & 0.75 & 0.55 & 0.56 \\
MO6 & 0.87 & 0.85 & 0.79 & 0.78 & 0.86 & 0.87 & 0.78 & 0.79 & 0.92 & 0.89 & 0.85 & 0.85 & 0.90 & 0.87 & 0.81 & 0.81 \\
MO7 & 0.67 & 0.69 & 0.40 & \underline{0.42} & 0.67 & 0.67 & \underline{0.40} & \underline{0.40} & 0.75 & 0.65 & 0.47 & 0.71 & 0.67 & 0.76 & \underline{0.40} & \underline{0.49} \\
\midrule
\textbf{Motivation} & 0.77 & 0.83 & 0.66 & 0.73 & 0.82 & 0.83 & 0.71 & 0.72 & 0.85 & 0.81 & 0.75 & 0.75 & 0.83 & 0.82 & 0.74 & 0.72 \\
\midrule
AP1 & 0.89 & 0.91 & 0.87 & 0.84 & 0.91 & 0.91 & 0.84 & 0.84 & 0.93 & 0.85 & 0.85 & 0.82 & 0.93 & 0.89 & 0.89 & 0.85 \\
AP2 & 0.86 & 0.85 & 0.76 & 0.71 & 0.87 & 0.84 & 0.79 & 0.70 & 0.89 & 0.88 & 0.81 & 0.77 & 0.86 & 0.86 & 0.91 & 0.74 \\
AP3 & 0.75 & 0.70 & 0.64 & 0.60 & 0.75 & 0.72 & 0.72 & 0.65 & 0.78 & 0.77 & 0.73 & 0.66 & 0.85 & 0.75 & 0.87 & 0.67 \\
AP4 & \underline{0.15} & 0.78 & \underline{0.11} & 0.62 & 0.80 & 0.78 & 0.64 & 0.62 & 0.82 & 0.76 & 0.65 & 0.67 & 0.84 & 0.82 & 0.67 & 0.65 \\
AP5 & 0.62 & 0.76 & 0.49 & 0.56 & 0.80 & 0.76 & 0.60 & 0.56 & 0.84 & 0.75 & 0.56 & 0.62 & 0.82 & 0.84 & 0.62 & 0.67 \\
AP6 & 0.83 & 0.77 & 0.76 & 0.66 & 0.80 & 0.76 & 0.74 & 0.63 & 0.81 & 0.84 & 0.88 & 0.80 & 0.78 & 0.83 & 0.88 & 0.82 \\
AP7 & \underline{0.16} & \underline{0.65} & \underline{0.15} & 0.65 & \underline{0.64} & \underline{0.65} & 0.64 & 0.65 & \underline{0.67} & 0.71 & 0.71 & 0.60 & \underline{0.56} & 0.75 & 0.65 & 0.71 \\
AP8 & \textbf{0.96} & \textbf{0.96} & 0.85 & 0.85 & \textbf{0.96} & \textbf{0.96} & 0.85 & 0.85 & \textbf{0.96} & \textbf{0.96} & 0.85 & 0.85 & \textbf{0.96} & \textbf{0.96} & 0.85 & 0.85 \\
\midrule
\textbf{Approach} & 0.65 & 0.80 & 0.58 & 0.69 & 0.82 & 0.80 & 0.73 & 0.69 & 0.84 & 0.82 & 0.76 & 0.72 & 0.82 & 0.84 & 0.79 & 0.75 \\
\midrule
MD1 & 0.67 & 0.67 & 0.44 & \underline{0.44} & 0.67 & 0.67 & \underline{0.44} & \underline{0.44} & 0.69 & \underline{0.64} & \underline{0.45} & 0.58 & 0.65 & 0.78 & \underline{0.49} & 0.58 \\
MD2 & 0.73 & 0.76 & 0.69 & 0.71 & 0.75 & 0.76 & 0.69 & 0.71 & 0.76 & 0.75 & 0.71 & 0.69 & 0.76 & 0.76 & 0.71 & 0.71 \\
MD3 & 0.80 & 0.82 & 0.67 & 0.69 & 0.80 & 0.82 & 0.67 & 0.69 & 0.85 & 0.78 & 0.73 & 0.69 & 0.87 & 0.78 & 0.75 & 0.73 \\
MD4 & \underline{0.56} & \underline{0.58} & 0.73 & 0.71 & \underline{0.56} & \underline{0.58} & 0.73 & 0.71 & \underline{0.62} & 0.64 & 0.71 & 0.73 & \underline{0.60} & \underline{0.58} & 0.73 & 0.67 \\
MD5 & 0.71 & 0.69 & 0.45 & 0.44 & 0.71 & 0.71 & 0.49 & 0.45 & 0.78 & 0.64 & 0.56 & 0.64 & 0.82 & 0.70 & 0.71 & 0.70 \\
MD6 & 0.58 & \underline{0.56} & \underline{0.33} & \underline{0.31} & \underline{0.58} & \underline{0.56} & \underline{0.36} & \underline{0.31} & 0.67 & 0.73 & \underline{0.45} & \underline{0.55} & 0.64 & 0.65 & 0.56 & \underline{0.55} \\
MD7 & 0.56 & \underline{0.56} & \underline{0.38} & \underline{0.38} & \underline{0.56} & \underline{0.56} & \underline{0.38} & \underline{0.38} & \underline{0.58} & \underline{0.60} & \underline{0.44} & \underline{0.45} & \underline{0.60} & \underline{0.60} & \underline{0.42} & \underline{0.42} \\
\midrule
\textbf{Methodology} & 0.66 & 0.66 & 0.53 & 0.52 & 0.66 & 0.67 & 0.54 & 0.53 & 0.71 & 0.68 & 0.58 & 0.62 & 0.71 & 0.70 & 0.62 & 0.62 \\
\midrule
SC1 & \textbf{0.98} & \textbf{0.98} & \textbf{0.95} & \textbf{0.95} & \textbf{0.98} & \textbf{0.98} & \textbf{0.95} & \textbf{0.95} & \textbf{0.96} & \textbf{0.96} & \textbf{0.96} & \textbf{0.96} & \textbf{0.98} & \textbf{0.96} & \textbf{0.95} & \textbf{0.96} \\
SC2 & \textbf{0.98} & \textbf{0.98} & \textbf{0.95} & \textbf{0.95} & \textbf{0.98} & \textbf{0.98} & \textbf{0.95} & \textbf{0.95} & \textbf{0.98} & \textbf{0.95} & \textbf{0.95} & \textbf{0.91} & \textbf{0.98} & \textbf{0.98} & \textbf{0.95} & \textbf{0.95} \\
SC3 & 0.91 & 0.91 & \textbf{0.96} & \textbf{0.96} & 0.91 & 0.91 & \textbf{0.96} & \textbf{0.96} & 0.91 & 0.93 & \textbf{0.96} & \textbf{0.98} & 0.91 & 0.91 & \textbf{0.96} & \textbf{0.96} \\
SC4 & 0.85 & 0.82 & 0.85 & 0.78 & 0.90 & 0.78 & \textbf{0.95} & 0.72 & 0.85 & 0.83 & 0.88 & 0.81 & 0.85 & 0.77 & 0.88 & 0.74 \\
\midrule
\textbf{Schedule} & 0.93 & 0.92 & 0.93 & 0.91 & 0.94 & 0.91 & 0.95 & 0.89 & 0.93 & 0.92 & 0.94 & 0.92 & 0.93 & 0.91 & 0.93 & 0.90 \\
\midrule
BI1 & 0.93 & 0.72 & 0.72 & 0.59 & 0.91 & 0.91 & 0.71 & 0.71 & \underline{0.27} & \underline{0.13} & \underline{0.40} & \underline{0.33} & \underline{0.20} & \underline{0.49} & \underline{0.40} & \underline{0.55} \\
BI2 & 0.90 & 0.92 & 0.68 & 0.73 & 0.90 & 0.88 & 0.71 & 0.66 & 0.86 & 0.91 & 0.80 & 0.72 & 0.91 & 0.88 & 0.75 & 0.70 \\
\midrule
\textbf{Bibliography} & 0.91 & 0.82 & 0.70 & 0.66 & 0.91 & 0.90 & 0.71 & 0.68 & 0.57 & 0.52 & 0.60 & 0.52 & 0.56 & 0.69 & 0.57 & 0.62 \\
\midrule
\midrule
\textbf{All criteria} & 0.75 & 0.80 & 0.66 & 0.70 & 0.81 & 0.80 & 0.71 & 0.70 & 0.81 & 0.77 & 0.74 & 0.72 & 0.79 & 0.79 & 0.75 & 0.73 \\
\midrule
\bottomrule
\end{tabular}
\caption{Quadratic-weighted agreement between open-weight LLM scores and human ratings for \expose{} criteria. Columns report $A_{G1}$ (Group~1 vs.~LLM) and $A_{G2}$ (Group~2 vs.~LLM) for the prompting variants shown (all, ind). Bold/underlined cells mark column-wise maxima/minima. Average rows summarize agreement within rubrics and across all criteria.}
\label{tab:model-compare-metric-expose-criteria}
\end{table*}
\begin{table*}[t]
\centering
\small
\setlength{\tabcolsep}{4.3pt}
\renewcommand{\arraystretch}{1.08}
\begin{tabular}{l | r r r r | r r r r | r r r r}
\toprule
 & \multicolumn{4}{c}{Claude Opus 4-6} & \multicolumn{4}{c}{Gemini 3.1 Pro} & \multicolumn{4}{c}{GPT-5.2} \\
Code & $A^{\text{all}}_{\text{G1}}$ & $A^{\text{ind}}_{\text{G1}}$ & $A^{\text{all}}_{\text{G2}}$ & $A^{\text{ind}}_{\text{G2}}$ & $A^{\text{all}}_{\text{G1}}$ & $A^{\text{ind}}_{\text{G1}}$ & $A^{\text{all}}_{\text{G2}}$ & $A^{\text{ind}}_{\text{G2}}$ & $A^{\text{all}}_{\text{G1}}$ & $A^{\text{ind}}_{\text{G1}}$ & $A^{\text{all}}_{\text{G2}}$ & $A^{\text{ind}}_{\text{G2}}$ \\
\midrule
LA1 & 0.84 & 0.83 & 0.87 & 0.86 & 0.85 & 0.85 & 0.80 & 0.85 & 0.86 & 0.83 & 0.91 & 0.88 \\
LA2 & 0.92 & 0.92 & 0.87 & 0.86 & 0.89 & 0.85 & 0.76 & 0.76 & 0.90 & 0.91 & 0.80 & 0.86 \\
\midrule
\textbf{Language} & 0.88 & 0.87 & 0.87 & 0.86 & 0.87 & 0.85 & 0.78 & 0.81 & 0.88 & 0.87 & 0.85 & 0.87 \\
\midrule
ME1 & 0.93 & 0.92 & 0.88 & 0.89 & 0.92 & 0.94 & 0.84 & 0.83 & 0.90 & 0.89 & 0.89 & 0.88 \\
\midrule
\textbf{Metadata} & 0.93 & 0.92 & 0.88 & 0.89 & 0.92 & 0.94 & 0.84 & 0.83 & 0.90 & 0.89 & 0.89 & 0.88 \\
\midrule
MO1 & 0.87 & 0.87 & 0.82 & 0.82 & 0.87 & 0.84 & 0.82 & 0.81 & 0.87 & 0.80 & 0.82 & 0.71 \\
MO2 & 0.89 & 0.91 & 0.78 & 0.81 & 0.88 & 0.87 & 0.77 & 0.77 & 0.89 & 0.87 & 0.84 & 0.75 \\
MO3 & \textbf{0.98} & \textbf{0.98} & \textbf{0.93} & 0.93 & \textbf{0.98} & \textbf{0.98} & \textbf{0.93} & \textbf{0.90} & \textbf{0.98} & \textbf{0.98} & 0.93 & \textbf{0.93} \\
MO4 & 0.82 & 0.82 & 0.84 & 0.84 & 0.82 & 0.81 & 0.84 & 0.81 & 0.82 & 0.82 & 0.84 & 0.84 \\
MO5 & 0.76 & 0.76 & \underline{0.58} & \underline{0.58} & 0.76 & 0.71 & \underline{0.58} & \underline{0.54} & 0.75 & 0.75 & \underline{0.56} & \underline{0.56} \\
MO6 & 0.93 & 0.92 & 0.90 & 0.91 & 0.88 & 0.92 & 0.82 & 0.84 & 0.91 & 0.90 & 0.89 & 0.90 \\
MO7 & \underline{0.60} & \underline{0.56} & 0.69 & 0.65 & 0.71 & \underline{0.55} & \underline{0.58} & 0.67 & \underline{0.60} & \underline{0.45} & 0.69 & 0.65 \\
\midrule
\textbf{Motivation} & 0.84 & 0.83 & 0.79 & 0.79 & 0.84 & 0.81 & 0.76 & 0.76 & 0.83 & 0.80 & 0.79 & 0.76 \\
\midrule
AP1 & 0.91 & 0.93 & 0.87 & \textbf{0.93} & 0.91 & 0.95 & 0.87 & 0.83 & 0.91 & 0.89 & 0.87 & 0.85 \\
AP2 & 0.91 & 0.90 & 0.85 & 0.89 & 0.88 & 0.87 & 0.79 & 0.74 & 0.92 & 0.89 & 0.89 & 0.78 \\
AP3 & 0.80 & 0.84 & 0.77 & 0.85 & 0.75 & 0.80 & 0.67 & 0.70 & 0.88 & 0.85 & 0.84 & 0.79 \\
AP4 & 0.84 & 0.82 & 0.67 & 0.69 & 0.80 & 0.78 & 0.64 & 0.66 & 0.80 & 0.84 & 0.67 & 0.67 \\
AP5 & 0.76 & \underline{0.62} & 0.67 & 0.67 & 0.78 & 0.68 & 0.65 & 0.66 & 0.80 & 0.73 & \underline{0.64} & 0.71 \\
AP6 & 0.85 & 0.85 & 0.85 & 0.82 & 0.80 & 0.79 & 0.77 & 0.81 & 0.87 & 0.88 & 0.87 & 0.84 \\
AP7 & 0.67 & 0.67 & 0.67 & \underline{0.64} & \underline{0.69} & 0.66 & 0.69 & 0.80 & 0.73 & 0.71 & 0.69 & 0.64 \\
AP8 & \textbf{0.96} & \textbf{0.96} & 0.85 & 0.85 & \textbf{0.96} & \textbf{0.98} & 0.85 & 0.84 & \textbf{0.96} & \textbf{0.96} & 0.85 & 0.85 \\
\midrule
\textbf{Approach} & 0.84 & 0.82 & 0.78 & 0.79 & 0.82 & 0.81 & 0.74 & 0.75 & 0.86 & 0.84 & 0.79 & 0.77 \\
\midrule
MD1 & 0.73 & 0.76 & \underline{0.64} & 0.64 & 0.71 & 0.68 & \underline{0.55} & \underline{0.56} & \underline{0.65} & 0.69 & \underline{0.64} & 0.64 \\
MD2 & 0.78 & 0.78 & 0.73 & 0.69 & 0.76 & 0.71 & 0.71 & 0.71 & 0.78 & 0.75 & 0.73 & 0.69 \\
MD3 & 0.85 & 0.73 & 0.65 & 0.64 & 0.84 & 0.79 & 0.71 & 0.67 & 0.85 & 0.80 & 0.69 & 0.71 \\
MD4 & \underline{0.65} & \underline{0.62} & 0.71 & 0.75 & \underline{0.62} & \underline{0.54} & 0.75 & 0.83 & 0.69 & \underline{0.58} & 0.71 & 0.75 \\
MD5 & 0.84 & 0.80 & 0.69 & 0.73 & 0.85 & \underline{0.66} & 0.67 & \underline{0.59} & 0.71 & 0.65 & 0.67 & 0.76 \\
MD6 & 0.69 & 0.71 & 0.76 & 0.75 & 0.71 & 0.68 & 0.60 & 0.66 & 0.71 & 0.73 & 0.75 & \underline{0.55} \\
MD7 & \underline{0.65} & 0.62 & \underline{0.47} & \underline{0.44} & \underline{0.62} & 0.71 & \underline{0.44} & \underline{0.49} & \underline{0.64} & \underline{0.64} & 0.67 & \underline{0.49} \\
\midrule
\textbf{Methodology} & 0.74 & 0.72 & 0.66 & 0.66 & 0.73 & 0.68 & 0.63 & 0.64 & 0.72 & 0.69 & 0.69 & 0.65 \\
\midrule
SC1 & \textbf{0.96} & \textbf{0.98} & \textbf{0.96} & \textbf{0.98} & \textbf{0.96} & \textbf{0.97} & \textbf{0.96} & \textbf{0.93} & \textbf{0.96} & \textbf{0.96} & \textbf{0.96} & \textbf{0.96} \\
SC2 & \textbf{0.98} & \textbf{1.00} & \textbf{0.95} & \textbf{0.96} & \textbf{0.98} & \textbf{0.97} & \textbf{0.95} & \textbf{0.93} & \textbf{0.98} & \textbf{0.98} & \textbf{0.95} & \textbf{0.95} \\
SC3 & 0.91 & 0.91 & \textbf{0.96} & \textbf{0.96} & 0.91 & 0.93 & \textbf{0.96} & \textbf{1.00} & 0.91 & 0.93 & \textbf{0.96} & \textbf{0.98} \\
SC4 & 0.84 & 0.90 & 0.91 & 0.89 & 0.80 & 0.85 & 0.78 & 0.85 & 0.89 & 0.89 & \textbf{0.95} & 0.91 \\
\midrule
\textbf{Schedule} & 0.92 & 0.95 & 0.95 & 0.95 & 0.91 & 0.93 & 0.91 & 0.92 & 0.94 & 0.94 & 0.95 & 0.95 \\
\midrule
BI1 & \underline{0.58} & \underline{0.25} & \underline{0.56} & \underline{0.35} & \underline{0.62} & \underline{0.62} & 0.67 & 0.62 & \underline{0.27} & \underline{0.15} & \underline{0.40} & \underline{0.35} \\
BI2 & 0.87 & 0.83 & 0.86 & 0.90 & 0.92 & 0.90 & 0.72 & 0.72 & 0.86 & 0.89 & 0.87 & 0.84 \\
\midrule
\textbf{Bibliography} & 0.72 & 0.54 & 0.71 & 0.62 & 0.77 & 0.76 & 0.70 & 0.67 & 0.57 & 0.52 & 0.63 & 0.59 \\
\midrule
\midrule
\textbf{All criteria} & 0.83 & 0.81 & 0.78 & 0.78 & 0.82 & 0.80 & 0.75 & 0.75 & 0.82 & 0.79 & 0.79 & 0.76 \\
\midrule
\bottomrule
\end{tabular}
\caption{Quadratic-weighted agreement between closed-source LLM scores and human ratings for \expose{} criteria. Columns report $A_{G1}$ (Group~1 vs.~LLM) and $A_{G2}$ (Group~2 vs.~LLM) for the prompting variants shown (all, ind). Bold/underlined cells mark column-wise maxima/minima. Average rows summarize agreement within rubrics and across all criteria.}
\label{tab:model-compare-metric-expose-criteria-closed}
\end{table*}

\begin{table*}[t]
\centering
\small
\setlength{\tabcolsep}{3.3pt}
\renewcommand{\arraystretch}{1.08}
\begin{tabular}{l | r r r r | r r r r | r r r r | r r r r}
\toprule
 & \multicolumn{4}{c}{Qwen3 (80B)} & \multicolumn{4}{c}{Llama 3.3 (70B)} & \multicolumn{4}{c}{GPT OSS (120B)} & \multicolumn{4}{c}{GPT OSS (20B)} \\
Code & $A^{\text{all}}_{\text{G1}}$ & $A^{\text{ind}}_{\text{G1}}$ & $A^{\text{all}}_{\text{G2}}$ & $A^{\text{ind}}_{\text{G2}}$ & $A^{\text{all}}_{\text{G1}}$ & $A^{\text{ind}}_{\text{G1}}$ & $A^{\text{all}}_{\text{G2}}$ & $A^{\text{ind}}_{\text{G2}}$ & $A^{\text{all}}_{\text{G1}}$ & $A^{\text{ind}}_{\text{G1}}$ & $A^{\text{all}}_{\text{G2}}$ & $A^{\text{ind}}_{\text{G2}}$ & $A^{\text{all}}_{\text{G1}}$ & $A^{\text{ind}}_{\text{G1}}$ & $A^{\text{all}}_{\text{G2}}$ & $A^{\text{ind}}_{\text{G2}}$ \\
\midrule
SC1 & 0.83 & 0.83 & 0.85 & 0.84 & 0.84 & 0.85 & 0.88 & 0.89 & 0.85 & 0.83 & 0.87 & 0.87 & 0.86 & 0.88 & 0.88 & 0.91 \\
SC2 & 0.92 & 0.92 & 0.80 & 0.81 & 0.90 & 0.86 & 0.81 & 0.80 & 0.91 & 0.86 & 0.83 & 0.85 & 0.84 & 0.85 & 0.85 & 0.83 \\
\midrule
\textbf{Structure and Clarity} & 0.88 & 0.87 & 0.82 & 0.83 & 0.87 & 0.85 & 0.84 & 0.85 & 0.88 & 0.84 & 0.85 & 0.86 & 0.85 & 0.87 & 0.87 & 0.87 \\
\midrule
LG1 & 0.90 & 0.91 & 0.98 & 0.97 & 0.91 & 0.88 & 0.97 & 0.94 & 0.86 & 0.67 & 0.85 & 0.62 & 0.75 & 0.69 & 0.75 & 0.64 \\
LG2 & \textbf{0.98} & \textbf{0.98} & \textbf{0.99} & \textbf{0.98} & \textbf{0.98} & \textbf{0.98} & 0.99 & \textbf{0.99} & \textbf{0.98} & \textbf{0.99} & \textbf{0.99} & \textbf{1.00} & \textbf{0.96} & \textbf{0.97} & \textbf{0.98} & \textbf{1.00} \\
\midrule
\textbf{Language} & 0.94 & 0.94 & 0.99 & 0.97 & 0.94 & 0.93 & 0.98 & 0.97 & 0.92 & 0.83 & 0.92 & 0.81 & 0.85 & 0.83 & 0.87 & 0.82 \\
\midrule
FU1 & \underline{0.75} & \underline{0.70} & \underline{0.74} & 0.69 & \underline{0.74} & 0.78 & \underline{0.72} & 0.73 & \underline{0.64} & 0.63 & \underline{0.57} & 0.55 & 0.66 & 0.55 & 0.60 & 0.48 \\
FU2 & \underline{0.75} & \underline{0.71} & \underline{0.73} & \underline{0.69} & \underline{0.76} & \underline{0.78} & \underline{0.72} & 0.75 & 0.64 & 0.66 & 0.62 & 0.65 & 0.58 & 0.61 & 0.59 & 0.60 \\
\midrule
\textbf{Feed Up} & 0.75 & 0.71 & 0.73 & 0.69 & 0.75 & 0.78 & 0.72 & 0.74 & 0.64 & 0.64 & 0.59 & 0.60 & 0.62 & 0.58 & 0.59 & 0.54 \\
\midrule
FB1 & 0.94 & 0.94 & \textbf{0.99} & \textbf{0.99} & 0.94 & 0.92 & \textbf{0.99} & 0.96 & \underline{0.23} & \underline{0.10} & \underline{0.20} & \underline{0.06} & \underline{0.25} & \underline{0.09} & \underline{0.22} & \underline{0.05} \\
FB2 & 0.92 & 0.93 & 0.97 & 0.96 & 0.91 & 0.93 & 0.95 & 0.96 & \underline{0.24} & \underline{0.09} & \underline{0.19} & \underline{0.05} & \underline{0.08} & \underline{0.10} & \underline{0.04} & \underline{0.04} \\
FB3 & 0.79 & 0.79 & \underline{0.73} & 0.73 & \underline{0.79} & 0.78 & \underline{0.73} & 0.74 & 0.77 & 0.62 & 0.73 & 0.63 & 0.75 & 0.59 & 0.69 & 0.58 \\
FB4 & 0.84 & 0.84 & 0.96 & 0.96 & 0.84 & 0.84 & 0.96 & 0.96 & 0.83 & 0.74 & 0.93 & 0.78 & 0.84 & 0.84 & 0.96 & 0.91 \\
FB5 & 0.95 & 0.95 & 0.94 & 0.94 & 0.95 & 0.94 & 0.94 & 0.92 & 0.95 & 0.92 & 0.94 & 0.92 & 0.95 & 0.76 & 0.94 & 0.79 \\
FB6 & \textbf{1.00} & \textbf{0.99} & 0.97 & 0.96 & \textbf{0.99} & \textbf{0.99} & 0.98 & \textbf{0.98} & 0.96 & 0.83 & 0.95 & 0.82 & \underline{0.34} & 0.41 & \underline{0.37} & \underline{0.42} \\
\midrule
\textbf{Feed Back} & 0.91 & 0.91 & 0.93 & 0.92 & 0.90 & 0.90 & 0.93 & 0.92 & 0.66 & 0.55 & 0.66 & 0.54 & 0.54 & 0.47 & 0.54 & 0.46 \\
\midrule
FF1 & 0.96 & \textbf{0.96} & \textbf{0.99} & \textbf{0.99} & 0.96 & \textbf{0.96} & \textbf{0.99} & \textbf{0.99} & 0.96 & 0.90 & \textbf{0.99} & 0.92 & 0.91 & 0.71 & 0.90 & 0.70 \\
FF2 & \underline{0.78} & 0.78 & \underline{0.61} & \underline{0.61} & \underline{0.78} & 0.78 & \underline{0.61} & \underline{0.61} & 0.75 & \underline{0.37} & 0.62 & \underline{0.50} & 0.78 & \underline{0.36} & 0.71 & 0.51 \\
\midrule
\textbf{Feed Forward} & 0.87 & 0.87 & 0.80 & 0.80 & 0.87 & 0.87 & 0.80 & 0.80 & 0.86 & 0.63 & 0.81 & 0.71 & 0.84 & 0.53 & 0.80 & 0.60 \\
\midrule
CQ1 & 0.96 & 0.94 & 0.92 & 0.92 & 0.94 & 0.88 & 0.93 & 0.86 & 0.92 & \textbf{0.95} & 0.90 & 0.92 & 0.84 & 0.90 & 0.84 & 0.90 \\
CQ2 & 0.94 & 0.94 & 0.92 & 0.92 & 0.93 & 0.94 & 0.93 & 0.91 & 0.94 & 0.94 & 0.92 & 0.92 & 0.92 & 0.93 & 0.93 & 0.92 \\
CQ3 & \underline{0.77} & \textbf{0.98} & 0.80 & \textbf{0.99} & \textbf{0.98} & \textbf{0.98} & \textbf{0.99} & \textbf{0.99} & \textbf{0.98} & \textbf{0.96} & \textbf{0.99} & \textbf{0.96} & \textbf{0.98} & \textbf{0.96} & \textbf{0.99} & \textbf{0.97} \\
\midrule
\textbf{Content Quality} & 0.89 & 0.95 & 0.88 & 0.94 & 0.95 & 0.93 & 0.95 & 0.92 & 0.95 & 0.95 & 0.94 & 0.93 & 0.91 & 0.93 & 0.92 & 0.93 \\
\midrule
ER1 & 0.83 & \underline{0.63} & 0.88 & \underline{0.68} & 0.83 & \underline{0.44} & 0.88 & \underline{0.45} & 0.75 & 0.65 & 0.77 & 0.70 & 0.80 & 0.82 & 0.83 & 0.87 \\
ER2 & 0.92 & 0.74 & 0.92 & 0.75 & 0.93 & \underline{0.65} & 0.95 & \underline{0.61} & 0.89 & 0.91 & 0.89 & 0.91 & 0.86 & \textbf{0.94} & 0.88 & \textbf{0.94} \\
ER3 & \textbf{0.96} & 0.91 & 0.97 & 0.92 & 0.96 & 0.93 & 0.97 & 0.94 & \textbf{0.96} & 0.92 & 0.97 & \textbf{0.93} & \textbf{0.96} & 0.92 & \textbf{0.97} & 0.93 \\
ER4 & 0.80 & \underline{0.25} & 0.84 & \underline{0.26} & 0.94 & \underline{0.17} & 0.98 & \underline{0.13} & \underline{0.42} & \underline{0.32} & \underline{0.38} & \underline{0.32} & \underline{0.36} & \underline{0.36} & \underline{0.38} & \underline{0.34} \\
ER5 & \textbf{0.99} & 0.95 & \textbf{0.99} & 0.95 & \textbf{0.99} & 0.94 & \textbf{0.99} & 0.94 & \textbf{0.99} & \textbf{0.97} & \textbf{0.99} & \textbf{0.97} & \textbf{0.99} & \textbf{0.95} & \textbf{0.99} & \textbf{0.95} \\
\midrule
\textbf{Errors} & 0.90 & 0.69 & 0.92 & 0.71 & 0.93 & 0.63 & 0.95 & 0.62 & 0.80 & 0.75 & 0.80 & 0.77 & 0.79 & 0.80 & 0.81 & 0.81 \\
\midrule
\midrule
\textbf{All criteria} & 0.89 & 0.84 & 0.89 & 0.84 & 0.90 & 0.83 & 0.90 & 0.82 & 0.79 & 0.72 & 0.78 & 0.72 & 0.74 & 0.69 & 0.74 & 0.69 \\
\midrule
\bottomrule
\end{tabular}
\caption{Quadratic-weighted agreement between open-weight LLM scores and human ratings for review criteria. Columns report $A_{G1}$ (Group~1 vs.~LLM) and $A_{G2}$ (Group~2 vs.~LLM) for the prompting variants shown (all, ind). Bold/underlined cells mark column-wise maxima/minima. Average rows summarize agreement within rubrics and across all criteria.}
\label{tab:model-compare-metric-review-criteria}
\end{table*}
\begin{table*}[t]
\centering
\small
\setlength{\tabcolsep}{4.3pt}
\renewcommand{\arraystretch}{1.08}
\begin{tabular}{l | r r r r | r r r r | r r r r}
\toprule
 & \multicolumn{4}{c}{Claude Opus 4-6} & \multicolumn{4}{c}{Gemini 3.1 Pro} & \multicolumn{4}{c}{GPT-5.2} \\
Code & $A^{\text{all}}_{\text{G1}}$ & $A^{\text{ind}}_{\text{G1}}$ & $A^{\text{all}}_{\text{G2}}$ & $A^{\text{ind}}_{\text{G2}}$ & $A^{\text{all}}_{\text{G1}}$ & $A^{\text{ind}}_{\text{G1}}$ & $A^{\text{all}}_{\text{G2}}$ & $A^{\text{ind}}_{\text{G2}}$ & $A^{\text{all}}_{\text{G1}}$ & $A^{\text{ind}}_{\text{G1}}$ & $A^{\text{all}}_{\text{G2}}$ & $A^{\text{ind}}_{\text{G2}}$ \\
\midrule
SC1 & 0.85 & 0.84 & 0.89 & 0.87 & 0.86 & 0.87 & 0.86 & 0.87 & 0.84 & 0.85 & 0.86 & 0.86 \\
SC2 & 0.90 & 0.85 & 0.88 & 0.88 & 0.90 & 0.90 & 0.84 & 0.89 & 0.83 & 0.87 & 0.85 & 0.85 \\
\midrule
\textbf{Structure and Clarity} & 0.87 & 0.85 & 0.88 & 0.87 & 0.88 & 0.88 & 0.85 & 0.88 & 0.84 & 0.86 & 0.86 & 0.85 \\
\midrule
LG1 & \underline{0.81} & 0.80 & \underline{0.80} & 0.83 & 0.85 & 0.81 & 0.93 & 0.91 & 0.91 & 0.83 & 0.92 & 0.89 \\
LG2 & \textbf{0.98} & \textbf{0.99} & \textbf{0.99} & \textbf{0.99} & \textbf{0.98} & \textbf{0.99} & 0.99 & \textbf{0.99} & \textbf{0.98} & \textbf{0.98} & \textbf{0.99} & \textbf{0.98} \\
\midrule
\textbf{Language} & 0.90 & 0.90 & 0.90 & 0.91 & 0.92 & 0.90 & 0.96 & 0.95 & 0.94 & 0.91 & 0.95 & 0.94 \\
\midrule
FU1 & 0.87 & 0.87 & 0.87 & 0.83 & \underline{0.78} & 0.83 & \underline{0.76} & 0.77 & 0.83 & 0.85 & 0.80 & 0.79 \\
FU2 & 0.87 & 0.84 & \underline{0.85} & 0.82 & \underline{0.78} & \underline{0.80} & \underline{0.75} & \underline{0.76} & 0.83 & 0.84 & \underline{0.79} & 0.79 \\
\midrule
\textbf{Feed Up} & 0.87 & 0.86 & 0.86 & 0.83 & 0.78 & 0.81 & 0.76 & 0.77 & 0.83 & 0.84 & 0.79 & 0.79 \\
\midrule
FB1 & 0.94 & \underline{0.75} & \textbf{0.99} & 0.75 & 0.94 & \underline{0.75} & \textbf{0.99} & \underline{0.76} & 0.94 & \underline{0.65} & \textbf{0.99} & 0.64 \\
FB2 & 0.93 & \underline{0.75} & 0.94 & \underline{0.74} & 0.94 & 0.90 & 0.95 & 0.96 & 0.93 & \underline{0.28} & 0.92 & \underline{0.22} \\
FB3 & \underline{0.78} & \underline{0.75} & \underline{0.74} & \underline{0.71} & \underline{0.74} & \underline{0.60} & \underline{0.71} & \underline{0.62} & \underline{0.62} & \underline{0.42} & \underline{0.65} & \underline{0.47} \\
FB4 & \underline{0.84} & 0.85 & 0.96 & 0.97 & 0.84 & 0.83 & 0.96 & \textbf{0.99} & 0.84 & 0.84 & 0.96 & 0.96 \\
FB5 & 0.95 & 0.95 & 0.94 & 0.94 & 0.95 & 0.96 & 0.94 & 0.94 & 0.94 & 0.95 & 0.93 & 0.94 \\
FB6 & 0.95 & \textbf{0.97} & 0.96 & 0.98 & \textbf{0.99} & \textbf{0.99} & 0.96 & 0.97 & \textbf{0.97} & 0.85 & 0.94 & 0.84 \\
\midrule
\textbf{Feed Back} & 0.90 & 0.84 & 0.92 & 0.85 & 0.90 & 0.84 & 0.92 & 0.87 & 0.88 & 0.66 & 0.90 & 0.68 \\
\midrule
FF1 & \textbf{0.96} & 0.96 & \textbf{0.99} & \textbf{0.99} & 0.96 & 0.97 & \textbf{0.99} & \textbf{0.99} & 0.95 & \textbf{0.96} & 0.98 & \textbf{0.99} \\
FF2 & \underline{0.81} & 0.79 & \underline{0.68} & \underline{0.61} & \underline{0.81} & \underline{0.73} & \underline{0.70} & \underline{0.69} & \underline{0.79} & 0.79 & \underline{0.68} & \underline{0.63} \\
\midrule
\textbf{Feed Forward} & 0.89 & 0.88 & 0.83 & 0.80 & 0.89 & 0.85 & 0.84 & 0.84 & 0.87 & 0.88 & 0.83 & 0.81 \\
\midrule
CQ1 & 0.92 & 0.86 & 0.92 & 0.89 & 0.96 & 0.96 & 0.92 & 0.92 & 0.80 & 0.80 & 0.83 & 0.83 \\
CQ2 & 0.92 & 0.94 & 0.94 & 0.93 & 0.94 & 0.93 & 0.92 & 0.91 & 0.94 & 0.95 & 0.92 & 0.92 \\
CQ3 & \textbf{0.97} & \textbf{0.98} & 0.99 & \textbf{0.99} & \textbf{0.98} & \textbf{0.98} & \textbf{0.99} & \textbf{0.99} & \textbf{0.98} & \textbf{0.98} & \textbf{0.99} & \textbf{0.99} \\
\midrule
\textbf{Content Quality} & 0.94 & 0.92 & 0.95 & 0.93 & 0.96 & 0.96 & 0.94 & 0.94 & 0.91 & 0.91 & 0.92 & 0.91 \\
\midrule
ER1 & 0.86 & 0.84 & 0.91 & 0.85 & 0.84 & 0.84 & 0.89 & 0.88 & \underline{0.75} & 0.79 & \underline{0.79} & 0.83 \\
ER2 & 0.85 & 0.84 & 0.85 & 0.82 & 0.93 & \textbf{0.97} & 0.93 & 0.92 & 0.85 & 0.90 & 0.83 & 0.88 \\
ER3 & 0.94 & 0.84 & 0.92 & 0.79 & 0.95 & 0.85 & 0.96 & 0.84 & 0.86 & 0.81 & 0.85 & 0.78 \\
ER4 & 0.85 & \underline{0.42} & 0.87 & \underline{0.38} & 0.94 & 0.93 & 0.98 & 0.98 & \underline{0.79} & \underline{0.22} & 0.81 & \underline{0.18} \\
ER5 & \textbf{0.99} & \textbf{0.99} & \textbf{0.99} & \textbf{0.99} & \textbf{0.99} & 0.91 & \textbf{0.99} & 0.91 & \textbf{0.99} & \textbf{0.96} & \textbf{0.99} & \textbf{0.96} \\
\midrule
\textbf{Errors} & 0.90 & 0.78 & 0.91 & 0.77 & 0.93 & 0.90 & 0.95 & 0.91 & 0.85 & 0.73 & 0.85 & 0.73 \\
\midrule
\midrule
\textbf{All criteria} & 0.90 & 0.85 & 0.90 & 0.84 & 0.90 & 0.88 & 0.91 & 0.88 & 0.87 & 0.79 & 0.88 & 0.78 \\
\midrule
\bottomrule
\end{tabular}
\caption{Quadratic-weighted agreement between closed-source LLM scores and human ratings for review criteria. Columns report $A_{G1}$ (Group~1 vs.~LLM) and $A_{G2}$ (Group~2 vs.~LLM) for the prompting variants shown (all, ind). Bold/underlined cells mark column-wise maxima/minima. Average rows summarize agreement within rubrics and across all criteria.}
\label{tab:model-compare-metric-review-criteria-closed}
\end{table*}

\section{Illustrative Examples}
\label{appendix:examples}
This chapter presents two representative examples from \nameDataset. The examples illustrate how expos\'e drafts, reviewer comments, and revised final versions are connected in the dataset. Each example is shown in three parts: an annotated draft excerpt, a short summary of the reviewer feedback, and the corresponding final excerpt. Passages highlighted in red indicate text spans that received inline comments from the reviewer, in this case, marking weaknesses that were addressed during revision.

\paragraph{Example 1: Anchor}

This example illustrates how reviewers ask students to strengthen the \emph{anchor} of a proposal by making the motivation more concrete and evidence-based. In the draft, the problem context is understandable, but still too general: the text claims that museums have experienced a major decline in visitor numbers and need more engaging approaches, yet the argument is not substantiated with references or concrete examples. This is reflected not only in the human assessment, but also in the automatic scoring of the draft, where models differed in whether they interpreted the problem description as already sufficiently detailed or still somewhat too general. The automatic assessments of the draft show a clear split, which helps explain why the human evaluation was not entirely uniform either. First, most models already viewed the problem statement as sufficiently detailed. For example, \emph{Claude Opus 4.6} assigned a score of 2 and argued that ``the problem is described in detail,'' explicitly pointing to the post-COVID decline in visitor numbers, the reliance on traditional text-panel-based interaction, and the limited current use of holograms. Second, a smaller group of models judged the same passage more critically because the motivation remained somewhat general. For instance, \emph{GPT OSS (20B)} assigned a score of 1 and summarized the draft as ``Problem stated with some detail but missing deeper specifics.''

The student review is especially informative when viewed through \feeduphl{\emph{Feed Up}}, \feedbackhl{\emph{Feed Back}}, and \feedforwardhl{\emph{Feed Forward}}. It clearly identifies the goal of the passage, namely to motivate the topic convincingly, which is reflected in the score for \emph{Learning Goals} (Score 2). It also explains what is missing in the current text by pointing out the unsupported claim and the need for a citation, corresponding to positive scores across the \emph{Feed Back} criteria. Finally, it provides forward-looking guidance by recommending that important factual claims should be supported with evidence and statistics in future revisions, which is reflected in the scores for \emph{Self-regulation} and \emph{Learning Skills} (Score 1 each).

\begin{wideexamplefloat}{Example 1: Excerpt (Criteria: Anker)}
\textbf{Draft Exposé}

\begin{quote}
"The modern museum has multiple purposes - to curate and preserve, to research, and to reach out to the public. They challenge us and ask us to question our assumptions about the past or the world around us." [1] [...] But espacially the part about reaching out to the public doesn't hold truth nowadays. \reviewhl{Since the COVID-19 pandemic, museums worldwide have faced a dramatic decline in visitor numbers, highlighting the need for more engaging and innovative approaches to attract audiences.}\reviewnote{\textsc{Weakness:} Where does this information come from? Cite.} Despite \reviewhl{their}\reviewnote{\textsc{Weakness:} Unclear reference; use a precise noun phrase such as ``the museums''.} efforts to adapt many institutions still rely on traditional methods of \reviewhl{interaction}\reviewnote{\textsc{Weakness:} Underline why improving human interaction is important.}, by simply showcasing an object and additional information written on text panels.\mbox{[...]}
\end{quote}

\textbf{Review 1 (Junior Instructor)}

\begin{quote}
\mbox{[...]} The anchor section provides a general context for the problem. However, the problem lacks detailed evidence or why it is important to research on it; for instance, while it mentions a "dramatic decline" since COVID-19, no quantitative data or case studies are included to support the claim. To strengthen the anchor, it is recommended to reference studies on the impact of the pandemic on cultural institutions and to underline its relevance/importance.\mbox{[...]}   
\end{quote}

\textbf{Review 2 (Student)}

\begin{quote}
\mbox{[...]} \feeduphl{In your motivation, you wrote that since the COVID-19 pandemic, museums faced a dramatic decline in visitor numbers. You aim to motivate your topic by providing an up-to-date example and an alarming development. This is important to make the reader clear why you think that this topic is relevant.}[\textsc{Feed Up}]
\feedbackhl{To ensure you provide valid and important facts, you should always provide a citation or source of such claims. This ensures that the reader will trust your judgment and understand the importance of your work. In this case, you do not provide any source or citation for the claim. For a sophisticated paper, you should avoid having unproofed claims in your work.}[\textsc{Feed Back}]
\feedforwardhl{I would suggest citing some paper that proves your claim with statistics. This will lead to a better, more understandable paper and avoid having false claims in your work, which could be an undesired situation.}[\textsc{Feed Forward}] \mbox{[...]}
\end{quote}

\textbf{Final Exposé}

\begin{quote}
    \mbox{[...]} But especially the part about reaching out to the public doesn't hold truth nowadays. Since the COVID-19 pandemic, museums worldwide have faced a dramatic decline in visitor numbers [2], highlighting the need for more engaging and innovative approaches to attract audiences.  For example, the National Gallery in London still had a 55\% drop in visitors two years after the pandemic [3]. To counter this decline, museums need to innovate and create more engaging ways to interact with their content, because meaningful interaction is the key driver for museum visits. Despite the efforts to adapt, many institutions still rely on traditional methods of interaction\mbox{[...]}

\end{quote}

\end{wideexamplefloat}

\paragraph{Example 2: Teaser (Research Question)}
This example shows how reviewers use the \emph{teaser} criterion to assess whether a proposal states a clear research question. In the draft, the overall research direction is understandable, but the teaser remains too broad because it describes goals and anticipated contributions rather than a distinct research question. This is reflected especially clearly in the human reviews: one reviewer assigned a score of 1 and noted that a clearly formulated research question would give the expos\'e more structure, while another reviewer assigned a score of 0 and criticized the absence of a proper research question altogether. The automatic scoring of the draft points to the same central issue. First, several models took a strict view and penalized the text because no explicit research question was stated. For example, \emph{GPT 5.2} assigned a score of 0 and wrote that ``No explicit research question is stated.'' Second, some models interpreted the passage more leniently and treated it as an implicit research direction, while still criticizing the lack of precision. For instance, \emph{Claude Opus 4.6} assigned a score of 1 and argued that the work was described in broad terms without a clearly articulated research question. Third, a smaller group of models inferred a sufficiently clear question from the surrounding context. For example, \emph{Llama 3.3} assigned a score of 2 and described the research question as clearly formulated and aligned with the problem description.

\begin{wideexamplefloat}{Example 2: Excerpt (Criteria: Teaser (RQ))}
\textbf{Draft Exposé}

\begin{quote}
\mbox{[...]} \reviewhl{This research addresses the critical gap in current AI-driven cancer research by integrating the impact of the tumor microenvironment into data analysis workflows for personalized treatment planning.}\reviewnote{\textsc{Weakness}: This states a research aim rather than an explicit research question.} \reviewhl{We will discuss the landscape of adapting current AI-enhanced data analysis to the importance of TME, address the limitations of current AI models in incorporating TME data and explore the anticipated improvement of personalized treatment prediction.}\reviewnote{\textsc{Weakness}: The goals remain broad; the main research question is not clearly stated.}\mbox{[...]}
\end{quote}

\textbf{Review 1 (Junior Instructor)} \textit{Score: 1}

\begin{quote}
\mbox{[...]} Having an explicit, clearly defined research question gives a lot of structure to the overall expose, especially to the methodology section as your methods should tackle all details of all RQs. As you describe your work instead, the exact goals and your main focus are not entirely clear to me. When making your RQ more explicit, also keep in mind to check whether your methodology addresses all relevant sub-parts of your RQ(s). \mbox{[...]}
\end{quote}

\textbf{Review 2 (Senior Instructor) }\textit{Score: 0}

\begin{quote}
\mbox{[...]} Also, a major issue is the lack of a distinct research question. The Motivation raises a question right at the end of the first paragraph. However, this question is rather a rhetorical question than a research question. Authors should include at least one sound research question that is answered in the work. \mbox{[...]}
\end{quote}

\textbf{Final Exposé} \textit{Score: 2}

\begin{quote}
\mbox{[...]} The research will explore the following key questions:
\begin{itemize}
    \item How does the heterogeneity of the tumor microenvironment impact treatment response predictions?
    \item What are the limitations of current AI-driven models in incorporating TME data, and how can these be addressed?
    \item How can integrating TME characteristics improve the predictive accuracy of personalized cancer treatments? \mbox{[...]}
\end{itemize}
\end{quote}

\end{wideexamplefloat}

\section{Experimental Details}
\label{appendix:experimental-details}
\paragraph{Models}
\label{appendix:performance_models}
We evaluate a diverse set of recent instruction-tuned LLMs spanning different architectures, scales, and deployment regimes. Our open-weight on-premise selection includes
(i) a large dense decoder-only Transformer from the Llama~3.3 family (meta-llama/Llama-3.3-70B-Instruct; 70B parameters) \cite{grattafiori2024llama3herdmodels},
(ii) a high-sparsity Mixture-of-Experts (MoE) model with long-context support (Qwen/Qwen3-Next-80B-A3B-Instruct; 80B total parameters with 3B active) \cite{qwen3technicalreport},
as well as an independent MoE reasoning model in two variants,
(iii) a high-capacity reasoning model (openai/gpt-oss-120b; 117B total parameters with 5.1B active) and
(iv) a smaller low-latency variant (openai/gpt-oss-20b; 21B total parameters with 3.6B active) \cite{openai2025gptoss120bgptoss20bmodel}.
These open-weight models can be executed locally on two NVIDIA~A100~80GB GPUs, which is important for classroom deployment where data protection and limited consent often motivate on-premise inference rather than cloud-based APIs.

To contextualize results against strong proprietary baselines, we additionally include three closed-source, API-accessed models: OpenAI GPT 5.2 (\texttt{openai/gpt-5.2}), Anthropic Claude Opus~4.6 (\texttt{anthropic/claude-opus-4-6}), and Google Gemini~3.1 Pro Preview (\texttt{gemini/gemini-3.1-pro-preview}). Closed-source models are queried via the LiteLLM\footnote{\url{https://www.litellm.ai}} interface, enabling a unified prompting and logging pipeline across providers.

Inference for open-weight models is performed using the Hugging Face \textsc{Transformers} library\footnote{\url{https://huggingface.co/docs/transformers}} with deterministic greedy decoding (\texttt{do\_sample=False}). For API-based models, we likewise use greedy decoding by default (temperature $=0$; and where supported, $\texttt{top\_p}=1$).

\paragraph{Prompts}
 For both tasks, we use a unified prompting strategy (Appendix~\ref{appendix:prompt_templates}) with a system message that (i) assigns the model the role of an expert evaluator and (ii) enumerates the relevant criteria, including their names, short descriptions, allowed integer score ranges, and the verbal descriptions associated with each score level (Tables~\ref{tab:expose_rubric} and~\ref{tab:review_rubric}).  
The model is required to return a JSON list with one object per criterion, where each object contains the fields \texttt{criterion\_name}, \texttt{assigned\_score}, and \texttt{justification}.
The additional justifications are requested to explain the decision-making process, as it is crucial in an environment where people are affected by the LLM outcome~\cite{Kolov2026}. 

\paragraph{Criteria excluded from the LLM experiments} The exposé and review rubrics contain structural or format-related checks (e.g., LaTeX template usage, page limits) and content-related criteria.
We exclude the first one, as the LLM cannot successfully assess it because the generated PDF is needed for correct evaluation.
We also exclude additional points that can be given by the human evaluator, as these occur only in isolated cases and are intended for exceptional circumstances. The following \expose{} criteria are excluded: Metadata available; Template used; Number of pages; Additional points; Negative points. The following review criteria are excluded: Additional points; Negative points.

\section{Prompt Templates}
\label{appendix:prompt_templates}
This appendix provides the exact prompt formats used in all baseline experiments.
The rubric text included in the templates is automatically generated from the released JSON rubric files, ensuring that the prompts align with the annotation scheme used in \nameDataset{}.

For both tasks, we use a unified prompt structure. The system message (i) assigns the model the role of an expert evaluator and (ii) enumerates the relevant criteria, including their names, short descriptions, allowed integer score ranges, and the verbal descriptions associated with each score level (Tables~\ref{tab:expose_rubric} and~\ref{tab:review_rubric}). The model is required to return a JSON \emph{list} with one object per criterion, containing the fields \texttt{criterion\_name}, \texttt{assigned\_score}, and a short \texttt{justification}. The additional justifications are requested to explain the decision-making process, as it is crucial in an environment where people are affected by the LLM outcome~\cite{Kolov2026}.

\subsection{Expose Scoring Prompt}
\label{app:prompt-expose}
The user message provides the topic, the full exposé text, and the bibliography.

\begin{systempanel}{System message}
\begin{lstlisting}[language={},basicstyle=\ttfamily\small]
You are an expert evaluator specializing in assessing student exposes.
Your task is to carefully evaluate the given expose based on the specified
grading criteria.

The grading criteria are as follows:

## <Rubric Name>
<Rubric description>

### Criterion: <Criterion Name>
Description: <Criterion description>
Allowed scores:
- <Points> points: <Description>
...

## <Next Rubric>
...

You must:
  1. Evaluate the expose **criterion by criterion**.
  2. For **every** listed criterion, assign exactly **one** score using **only** the scoring options and point values provided.
  3. Justify each score with a clear, concise and **short** explanation.

Your response MUST be a list of dictionaries in **valid JSON** schema:

[
  {
    "criterion_name": "<the EXACT criterion name from the criteria list, without any other words>",
    "assigned_score": "<an INTEGER, one of the allowed point values for this criterion>",
    "justification": "<SHORT justification of why you decided to give this score>"
  },
  ...
]

Constraints:
- Return **one object per criterion**; do not skip any criteria.
- Use only the criteria and scoring options that are explicitly provided.
- Return a response that consists solely of valid JSON, with no additional explanatory text before or after it.
- Make sure the JSON is syntactically valid (no trailing commas, correct quotation marks, etc.).
\end{lstlisting}
\end{systempanel}

\begin{userpanel}{User message}
\begin{lstlisting}[language={},basicstyle=\ttfamily\small]
The topic is: "<TOPIC>".

Here is the expose you have to grade:

<EXPOSE TEXT>

Bibliography:
<BIBLIOGRAPHY>
\end{lstlisting}
\end{userpanel}

\subsection{Review Scoring Prompt}
\label{app:prompt-review}
The user message provides the full written review text.

{
\begin{systempanel}{System message}

\begin{lstlisting}[language={},basicstyle=\ttfamily\small]
You are an expert evaluator specializing in assessing the quality of student peer reviews.
Your task is to carefully evaluate the reviewer's performance based on the specified grading criteria.

Grading criteria:

## <Rubric Name>
<Rubric description>

### Criterion: <Criterion Name>
Description: <Criterion description>
Allowed scores:
- <Points> points: <Description>
...

## <Next Rubric>
...

You must:
1. Evaluate the peer review **criterion by criterion**.
2. For **every** listed criterion, assign exactly **one** score using **only** the scoring options and point values provided for that criterion.
3. Base your evaluation strictly on the quality of the full review text.
4. Justify each score with a clear, concise and **short** explanation.

Output format:
Return a list of objects following this **valid JSON** schema:

[
  {
    "criterion_name": "<EXACT criterion name from the criteria list>",
    "assigned_score": <INTEGER, one of the allowed point values for this criterion>,
    "justification": "<SHORT justification of why you decided to give this score>"
  },
  ...
]

Constraints:
- Return **one object per criterion**; do not skip any criteria.
- Use only the criteria and scoring options that are explicitly provided.
- The response must consist solely of valid JSON, with no additional explanatory text.
- Make sure the JSON is syntactically valid (no trailing commas, correct quotation marks, etc.).
\end{lstlisting}
\end{systempanel}
}

\begin{userpanel}{User message}

\begin{lstlisting}[language={},basicstyle=\ttfamily\small]
Here are the materials of the peer review you have to grade.

=== Full written text of the review ===
<FULL_REVIEW_TEXT>
\end{lstlisting}
\end{userpanel}

\section{Additional Experimental Results}
\label{appendix:results}
\begin{figure*}[t]
  \centering
  \includegraphics[width=\linewidth]{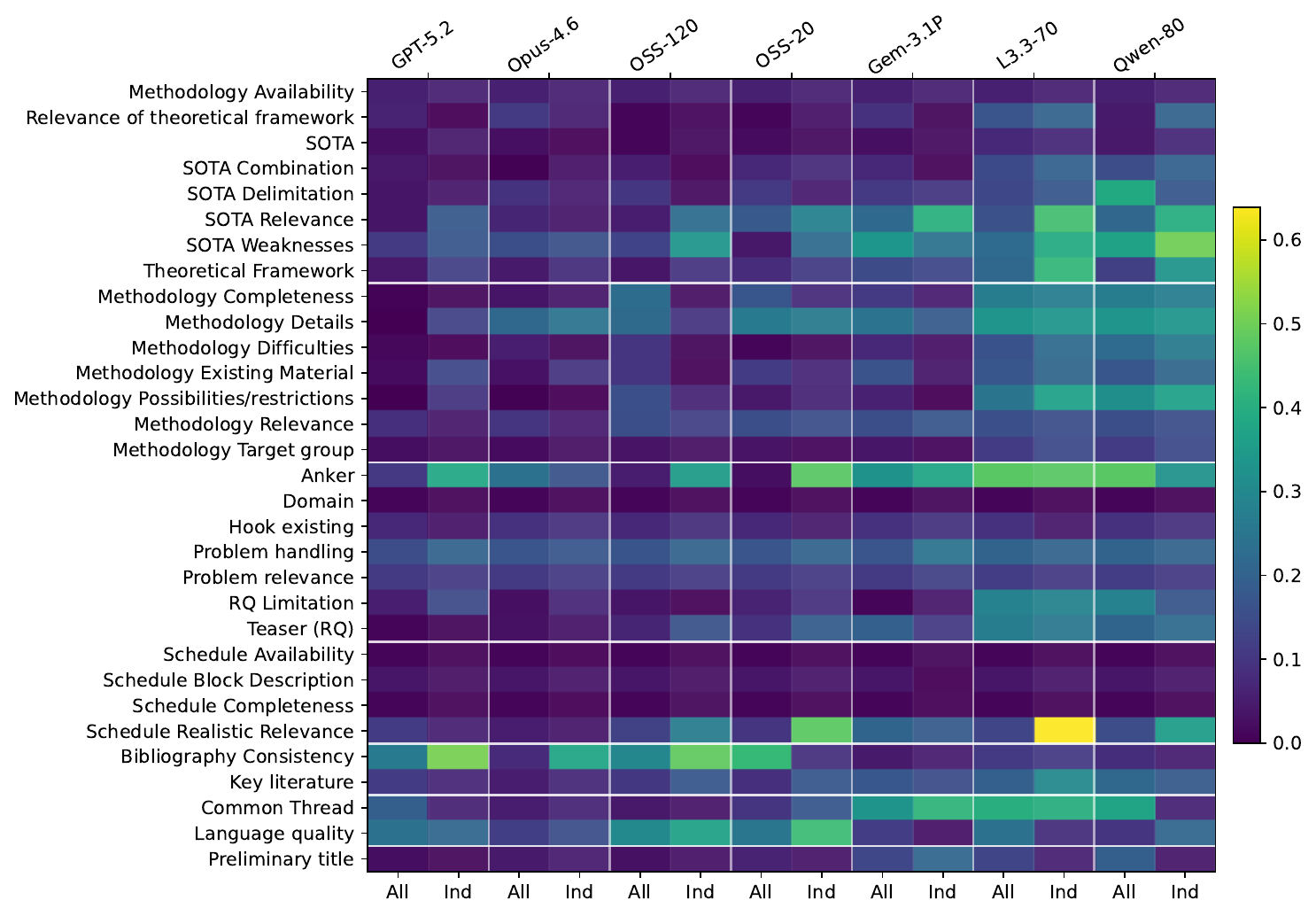}
  \caption{\textbf{Exposé: Criterion $\times$ model heatmap (JSD).}
  Cell values show weighted mean JSD per criterion and model (weights: items\_in\_bin). Lower is better. Model abbreviations: \textit{Opus-4.6}=Claude Opus 4.6; \textit{Gem-3.1P}=Gemini 3.1 Pro; \textit{GPT-5.2}=GPT 5.2; \textit{OSS-120}=GPT OSS (120B); \textit{OSS-20}=GPT OSS (20B); \textit{L3.3-70}=Llama 3.3 (70B); \textit{Qwen-80}=Qwen 3 (80B).}
  \label{fig:expose-criterion-model-jsd}
\end{figure*}

\begin{figure*}[t]
  \centering
  \includegraphics[width=\linewidth]{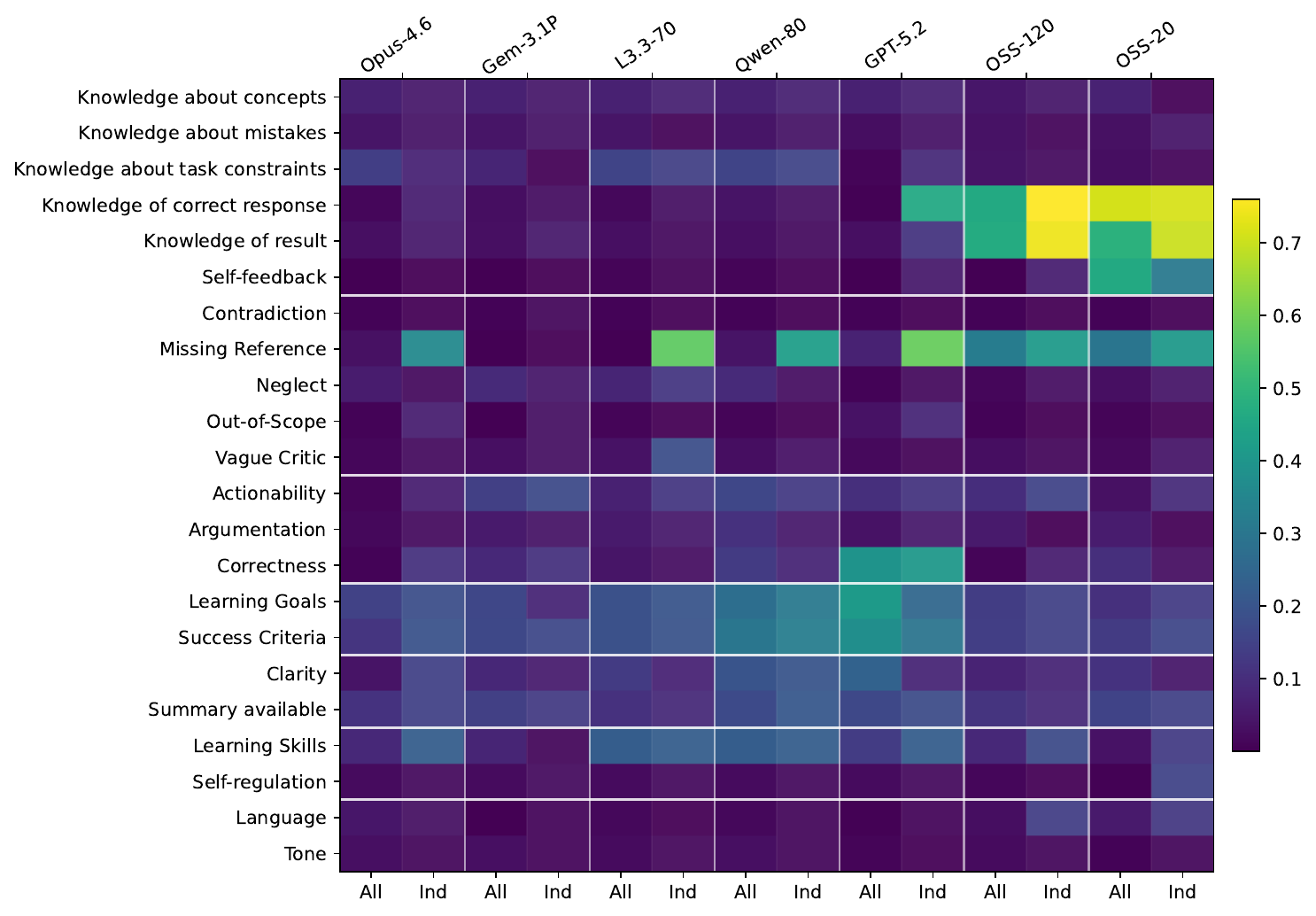}
  \caption{\textbf{Reviews: Criterion $\times$ model heatmap (JSD).}
  Cell values show weighted mean JSD per criterion and model (weights: items\_in\_bin). Lower is better. Model abbreviations: \textit{Opus-4.6}=Claude Opus 4.6; \textit{Gem-3.1P}=Gemini 3.1 Pro; \textit{GPT-5.2}=GPT 5.2; \textit{OSS-120}=GPT OSS (120B); \textit{OSS-20}=GPT OSS (20B); \textit{L3.3-70}=Llama 3.3 (70B); \textit{Qwen-80}=Qwen 3 (80B).}
  \label{fig:review-criterion-model-jsd}
\end{figure*}

\paragraph{Criterion-wise alignment} Figures~\ref{fig:expose-criterion-model-jsd} and~\ref{fig:review-criterion-model-jsd} report weighted mean binned JSD at the criterion level (darker = lower/better), complementing the domain-level averages in Table~\ref{tab:jsd-domain-allmodels}. Across both tasks, criteria on the 0--1--2 scale show a consistently higher JSD baseline than 0--1 criteria (Table~\ref{tab:jsd-domain-allmodels}), reflecting the increased degrees of freedom and typically higher annotator dispersion in three-way ratings, which makes distributional matching more challenging. For \expose{}, most criteria exhibit low divergence, but mismatches concentrate in the more interpretive rubric components—especially SOTA- and methodology-related criteria—with \textit{Anker} emerging as uniformly difficult and \textit{Schedule Realistic Relevance} showing a clear spike for some model/prompt settings. Reviews are more homogeneous overall: stylistic dimensions (language, tone) remain low across models, whereas divergence is driven primarily by epistemic “knowledge” judgments and reference-related criteria (notably \textit{Missing Reference}).

\begin{table*}[t]
\centering
\setlength{\tabcolsep}{4.1pt}
\renewcommand{\arraystretch}{1.08}
\begin{tabular}{c | c c | c c | c c | c c | c c | c c | c c}
\toprule
  & \multicolumn{2}{c|}{{\small \textbf{Opus-4.6}}} & \multicolumn{2}{c|}{{\small \textbf{Gem-3.1P}}} & \multicolumn{2}{c|}{{\small \textbf{GPT-5.2}}} & \multicolumn{2}{c|}{{\small \textbf{OSS-120}}} & \multicolumn{2}{c|}{{\small \textbf{OSS-20}}} & \multicolumn{2}{c|}{{\small \textbf{L3.3-70}}} & \multicolumn{2}{c}{{\small \textbf{Qwen-80}}} \\
\textbf{Domain} & \textbf{All} & \textbf{Ind} & \textbf{All} & \textbf{Ind} & \textbf{All} & \textbf{Ind} & \textbf{All} & \textbf{Ind} & \textbf{All} & \textbf{Ind} & \textbf{All} & \textbf{Ind} & \textbf{All} & \textbf{Ind} \\
\midrule
\addlinespace[0.35em]
\multicolumn{15}{c}{\textbf{Exposé criterion-based scoring (over 31 criteria)}} \\
\addlinespace[0.25em]
 0-1 (\#21) & 0.06 & 0.07 & 0.08 & 0.06 & 0.05 & 0.08 & 0.09 & 0.07 & 0.10 & 0.07 & 0.14 & 0.16 & 0.15 & 0.15 \\
\midrule
 0-1-2 (\#10) & 0.08 & 0.09 & 0.22 & 0.23 & 0.10 & 0.13 & 0.09 & 0.21 & 0.10 & 0.26 & 0.24 & 0.35 & 0.24 & 0.26 \\
\midrule
\specialrule{1.2pt}{0.8pt}{0.8pt}
 \textbf{All (\#31)} & \textbf{0.07} & \textbf{0.08} & \textbf{0.13} & \textbf{0.12} & \textbf{0.07} & \textbf{0.10} & \textbf{0.09} & \textbf{0.11} & \textbf{0.10} & \textbf{0.13} & \textbf{0.17} & \textbf{0.22} & \textbf{0.18} & \textbf{0.19} \\
\specialrule{1.2pt}{0.8pt}{0.6pt}
\addlinespace[0.35em]
\multicolumn{15}{c}{\textbf{Review criterion-based scoring (over 22 criteria)}} \\
\addlinespace[0.25em]
 0-1 (\#10) & 0.05 & 0.06 & 0.04 & 0.03 & 0.04 & 0.11 & 0.12 & 0.19 & 0.19 & 0.22 & 0.07 & 0.06 & 0.07 & 0.07 \\
\midrule
 0-1-2 (\#7) & 0.07 & 0.12 & 0.12 & 0.10 & 0.24 & 0.19 & 0.09 & 0.10 & 0.08 & 0.09 & 0.11 & 0.11 & 0.21 & 0.21 \\
\midrule
 -1-0 (\#5) & 0.03 & 0.10 & 0.03 & 0.03 & 0.03 & 0.14 & 0.07 & 0.11 & 0.06 & 0.11 & 0.03 & 0.16 & 0.04 & 0.08 \\
\midrule
\specialrule{1.2pt}{0.8pt}{0.8pt}
 \textbf{All (\#22)} & \textbf{0.05} & \textbf{0.09} & \textbf{0.06} & \textbf{0.06} & \textbf{0.10} & \textbf{0.14} & \textbf{0.10} & \textbf{0.14} & \textbf{0.13} & \textbf{0.15} & \textbf{0.08} & \textbf{0.10} & \textbf{0.11} & \textbf{0.12} \\
\bottomrule
\end{tabular}
\caption{Weighted-mean Jensen--Shannon divergence (JSD) between human-vote and LLM-vote distributions, grouped by domain. Prompting variants: All, Ind. Lower is better. Model abbreviations: \textit{Opus-4.6}=Claude Opus 4.6; \textit{Gem-3.1P}=Gemini 3.1 Pro; \textit{GPT-5.2}=GPT 5.2; \textit{OSS-120}=GPT OSS (120B); \textit{OSS-20}=GPT OSS (20B); \textit{L3.3-70}=Llama 3.3 (70B); \textit{Qwen-80}=Qwen 3 (80B).}
\label{tab:jsd-domain-allmodels}
\end{table*}

\begin{figure}[t]
  \centering
  \includegraphics[width=\columnwidth]{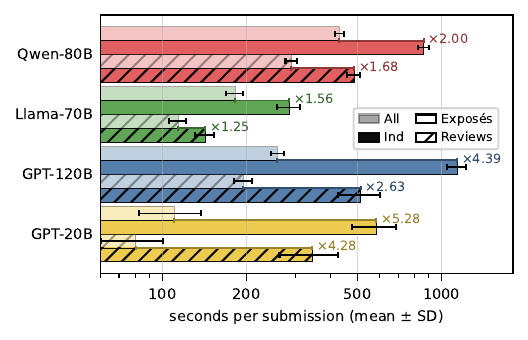}
  \caption{Runtime per submission for criterion-based assessments across open-weight models. Bars show mean seconds per submission with standard deviation bars. Within each task, \textbf{All} denotes combined prompting (light fill) and \textbf{Ind} denotes single-criterion prompting (dark fill). Review runs are hatched; Exposé runs are unhatched. Ratios show $\times(\mathrm{Ind}/\mathrm{All})$, highlighting the additional latency introduced by single-criterion prompting.}
  \label{fig:runtime-expose-review}
\end{figure}

\paragraph{Scalability and classroom deployment} Across all models and both tasks, combined prompting is more efficient in most cases than single-criterion prompting: it reduces mean runtime per submission by about $1.3\times$--$5.3\times$ and avoids the linear cost growth from issuing one model call per criterion (Figure~\ref{fig:runtime-expose-review}).
Single-criterion prompting also substantially increases total token processing because the full input is repeated across prompts, with particularly large overhead for long exposés and the review setting.
These efficiency differences become especially consequential under the \emph{observed course usage dynamics}.
As shown by the behavioural interaction logs (Appendix Figure~\ref{fig:appendix:behaviour:daily_activity}), student activity is strongly deadline-driven, with a pronounced spike in actions immediately before the review deadline, whereas instructor activity is relatively steady.
This implies that in practice, LLM-based scoring must be provisioned for short periods of peak demand rather than average-day usage.
As such, the overhead of single-criterion prompting is particularly problematic, while combined prompting better supports batching and throughput during deadline-induced bursts.
Consequently, \emph{combined prompting is the recommended operating point for classroom deployment under constrained on-premise resources, with model choice primarily determining the remaining latency--quality trade-off}. Runtime and token usage are reported in Appendix~\ref{appendix:performance} (Tables~\ref{tab:llm-runtime-tokens-expose}--\ref{tab:llm-runtime-tokens-review}).

\paragraph{Operational latency under peak-demand grading windows}
In classroom deployment under constrained on-premise resources, using LLM scores as instructor support (e.g., batch pre-scoring or rubric-aligned decision support) introduces a practical scheduling choice: scores can either be \emph{pre-computed} ahead of grading, or compute capacity must be provisioned to serve requests \emph{on demand} throughout the grading period.
Under deadline-driven bursts (Appendix Figure~\ref{fig:appendix:behaviour:daily_activity}), it can translate directly into waiting times, because even combined prompting incurs minute-level latency per submission and shows substantial min--max spread (Tables~\ref{tab:llm-runtime-tokens-expose}--\ref{tab:llm-runtime-tokens-review}).
To illustrate, consider a cohort of $N{=}600$ expos\'es scored on the same two-GPU setup used in our experiments. 
For expos\'es, the fastest combined configuration in Table~\ref{tab:llm-runtime-tokens-expose} (GPT OSS 20B, All) yields $600 \times 110.2\text{s} \approx 18.4$ hours on average; larger models further increase this budget (e.g., Qwen3 80B, All: $\approx 71.7$ hours).
For student reviews, the volume is larger ($2N$ submissions); in the worst-case combined configuration, runtimes of $\approx 289.6$s per review (Table~\ref{tab:llm-runtime-tokens-review}) translate into $\approx 96.5$ hours of wall-clock time, making peak-time live scoring particularly susceptible to queueing delays.

Finally, beyond latency, we observed occasional out-of-memory failures for unusually long inputs (e.g., exposés with many and very long comments), reinforcing the need for robust deployment to plan around worst-case input lengths (e.g., advance computation, batching, and input-length safeguards) rather than relying on unconstrained live scoring near deadlines.

\paragraph{Is review text enough for review assessment?}
Overall, we did not find an indication that review scoring requires context beyond the review text. In supplementary experiments, we augmented the review input jointly with additional context from the \expose{} and inline comments, but did not observe consistent improvements over the review-only setting.

\section{Runtime and Token Consumption}
\label{appendix:performance}

In this appendix, we focus exclusively on inference-time cost (runtime and token usage) under these fixed conditions.
Prompting variants (combined vs.\ single-criterion) are defined in Section \ref{subsec:baseline-setup}. 
For each model/condition, we measure wall-clock runtime both per full submission (\texttt{s/assessment}) and per model call (\texttt{s/prompt}).
Token counts are computed with the respective model tokenizer on the fully formatted prompt (system + user messages) and the generated JSON response, and reported as \texttt{tokens/prompt} split into \texttt{tokens/input} and \texttt{tokens/output}.
All values are aggregated across evaluation instances and reported as mean (SD) and [min--max], capturing variation in input length.

\paragraph{LLM scoring throughput: combined prompting is essential}
Tables~\ref{tab:llm-runtime-tokens-expose} and~\ref{tab:llm-runtime-tokens-review} report runtime and token usage for exposé and review scoring, respectively. Across both tasks, combined prompting is essential for throughput: it requires a single forward pass per submission and therefore yields the lowest latency (110--430s on average across models for exposé scoring). In contrast, single-criterion prompting multiplies calls per submission, increasing total runtime to 284--1133s on average for exposés, with worst cases up to 23 minutes. The same multiplicative effect appears in token processing: although single-criterion outputs are shorter per call, repeating the full input across criteria yields an order-of-magnitude increase in total processed tokens per submission (approximately $13\times$ in our setup). Review scoring exhibits the same trend (Table~\ref{tab:llm-runtime-tokens-review}), with combined prompting consistently yielding lower latency and substantially fewer processed tokens per submission than single-criterion prompting.

\begin{table*}[t]
\centering
\small
\setlength{\tabcolsep}{1.0pt}

\begin{tabular}{l c r r r r r}
\toprule
\textbf{Model} & \textbf{Mode} & \textbf{s/assessment} & \textbf{s/prompt} & \textbf{tokens/prompt} & \textbf{tokens/input} & \textbf{tokens/output} \\
\midrule
GPT OSS (20B) & All & 110.20 (\(\pm\)27.69) & 110.20 (\(\pm\)27.69) & 8,010.67 (\(\pm\)1,018.78) & 5,587.85 (\(\pm\)932.59) & 2,422.82 (\(\pm\)629.58) \\
\multicolumn{2}{r}{\textcolor{black!70}{[min--max]}} & \textcolor{black!70}{[84.88--289.55]} & \textcolor{black!70}{[84.88--289.55]} & \textcolor{black!70}{[6,384.00--11,178.00]} & \textcolor{black!70}{[4,092.00--8,765.00]} & \textcolor{black!70}{[1,874.00--6,489.00]} \\
\cmidrule(lr){1-7}
Llama 3.3 (70B) & All & 182.10 (\(\pm\)12.79) & 182.10 (\(\pm\)12.79) & 7,181.78 (\(\pm\)973.95) & 5,606.69 (\(\pm\)947.16) & 1,575.09 (\(\pm\)105.46) \\
\multicolumn{2}{r}{\textcolor{black!70}{[min--max]}} & \textcolor{black!70}{[163.18--212.11]} & \textcolor{black!70}{[163.18--212.11]} & \textcolor{black!70}{[5,594.00--10,311.00]} & \textcolor{black!70}{[4,091.00--8,839.00]} & \textcolor{black!70}{[1,399.00--1,858.00]} \\
\cmidrule(lr){1-7}
GPT OSS (120B) & All & 258.01 (\(\pm\)13.74) & 258.01 (\(\pm\)13.74) & 8,227.11 (\(\pm\)915.79) & 5,679.85 (\(\pm\)932.59) & 2,547.25 (\(\pm\)148.73) \\
\multicolumn{2}{r}{\textcolor{black!70}{[min--max]}} & \textcolor{black!70}{[227.52--304.39]} & \textcolor{black!70}{[227.52--304.39]} & \textcolor{black!70}{[6,692.00--11,292.00]} & \textcolor{black!70}{[4,184.00--8,857.00]} & \textcolor{black!70}{[2,265.00--3,067.00]} \\
\cmidrule(lr){1-7}
Llama 3.3 (70B) & Ind & 284.30 (\(\pm\)26.54) & 9.17 (\(\pm\)1.37) & 3,228.31 (\(\pm\)948.58) & 3,162.76 (\(\pm\)947.61) & 65.55 (\(\pm\)9.79) \\
\multicolumn{2}{r}{\textcolor{black!70}{[min--max]}} & \textcolor{black!70}{[237.56--370.23]} & \textcolor{black!70}{[5.43--14.43]} & \textcolor{black!70}{[1,656.00--6,555.00]} & \textcolor{black!70}{[1,613.00--6,490.00]} & \textcolor{black!70}{[42.00--117.00]} \\
\cmidrule(lr){1-7}
Qwen3 (80B) & All & 430.37 (\(\pm\)15.13) & 430.37 (\(\pm\)15.13) & 7,473.75 (\(\pm\)1,046.99) & 5,719.84 (\(\pm\)1,024.62) & 1,753.91 (\(\pm\)64.05) \\
\multicolumn{2}{r}{\textcolor{black!70}{[min--max]}} & \textcolor{black!70}{[394.29--460.62]} & \textcolor{black!70}{[394.29--460.62]} & \textcolor{black!70}{[5,775.00--11,079.00]} & \textcolor{black!70}{[4,126.00--9,301.00]} & \textcolor{black!70}{[1,602.00--1,874.00]} \\
\cmidrule(lr){1-7}
GPT OSS (20B) & Ind & 582.36 (\(\pm\)103.30) & 18.79 (\(\pm\)20.85) & 3,553.87 (\(\pm\)1,047.56) & 3,162.31 (\(\pm\)933.03) & 391.56 (\(\pm\)478.39) \\
\multicolumn{2}{r}{\textcolor{black!70}{[min--max]}} & \textcolor{black!70}{[452.01--878.16]} & \textcolor{black!70}{[7.34--357.11]} & \textcolor{black!70}{[1,821.00--11,982.00]} & \textcolor{black!70}{[1,633.00--6,431.00]} & \textcolor{black!70}{[123.00--8,091.00]} \\
\cmidrule(lr){1-7}
Qwen3 (80B) & Ind & 862.02 (\(\pm\)36.97) & 27.81 (\(\pm\)3.81) & 3,359.56 (\(\pm\)1,027.13) & 3,275.90 (\(\pm\)1,025.04) & 83.66 (\(\pm\)15.20) \\
\multicolumn{2}{r}{\textcolor{black!70}{[min--max]}} & \textcolor{black!70}{[789.80--946.85]} & \textcolor{black!70}{[19.10--41.08]} & \textcolor{black!70}{[1,705.00--7,050.00]} & \textcolor{black!70}{[1,648.00--6,952.00]} & \textcolor{black!70}{[51.00--138.00]} \\
\cmidrule(lr){1-7}
GPT OSS (120B) & Ind & 1133.41 (\(\pm\)89.76) & 34.35 (\(\pm\)6.90) & 3,383.58 (\(\pm\)932.15) & 3,160.07 (\(\pm\)933.04) & 223.51 (\(\pm\)67.47) \\
\multicolumn{2}{r}{\textcolor{black!70}{[min--max]}} & \textcolor{black!70}{[869.61--1398.68]} & \textcolor{black!70}{[19.20--97.15]} & \textcolor{black!70}{[1,771.00--6,902.00]} & \textcolor{black!70}{[1,627.00--6,431.00]} & \textcolor{black!70}{[100.00--854.00]} \\
\bottomrule
\end{tabular}
\caption{Runtime and token usage for LLM scoring. Entries report mean (\(\pm\) SD) and a second row with [min--max]. Mode: All = combined prompting; Ind = single-criterion prompting. Token statistics are reported per prompt (total, input, output).}
\label{tab:llm-runtime-tokens-expose}
\end{table*}

\begin{table*}[t]
\centering
\small
\setlength{\tabcolsep}{1.0pt}

\begin{tabular}{l c r r r r r}
\toprule
\textbf{Model} & \textbf{Mode} & \textbf{s/assessment} & \textbf{s/prompt} & \textbf{tokens/prompt} & \textbf{tokens/input} & \textbf{tokens/output} \\
\midrule
GPT OSS (20B) & All & 80.25 (\(\pm\)20.05) & 80.25 (\(\pm\)20.05) & 4,465.78 (\(\pm\)606.62) & 2,539.31 (\(\pm\)402.22) & 1,926.47 (\(\pm\)472.59) \\
\multicolumn{2}{r}{\textcolor{black!70}{[min--max]}} & \textcolor{black!70}{[60.56--345.62]} & \textcolor{black!70}{[60.56--345.62]} & \textcolor{black!70}{[3,641.00--10,957.00]} & \textcolor{black!70}{[2,117.00--6,330.00]} & \textcolor{black!70}{[1,449.00--8,091.00]} \\
\cmidrule(lr){1-7}
Llama 3.3 (70B) & All & 113.76 (\(\pm\)8.10) & 113.76 (\(\pm\)8.10) & 3,574.05 (\(\pm\)434.77) & 2,513.88 (\(\pm\)404.74) & 1,060.17 (\(\pm\)73.19) \\
\multicolumn{2}{r}{\textcolor{black!70}{[min--max]}} & \textcolor{black!70}{[98.39--135.10]} & \textcolor{black!70}{[98.39--135.10]} & \textcolor{black!70}{[3,041.00--7,464.00]} & \textcolor{black!70}{[2,088.00--6,329.00]} & \textcolor{black!70}{[920.00--1,264.00]} \\
\cmidrule(lr){1-7}
Llama 3.3 (70B) & Ind & 142.31 (\(\pm\)11.27) & 6.47 (\(\pm\)0.95) & 982.85 (\(\pm\)406.97) & 924.60 (\(\pm\)405.24) & 58.25 (\(\pm\)8.26) \\
\multicolumn{2}{r}{\textcolor{black!70}{[min--max]}} & \textcolor{black!70}{[121.03--210.90]} & \textcolor{black!70}{[4.26--11.25]} & \textcolor{black!70}{[518.00--4,845.00]} & \textcolor{black!70}{[471.00--4,774.00]} & \textcolor{black!70}{[40.00--93.00]} \\
\cmidrule(lr){1-7}
GPT OSS (120B) & All & 194.64 (\(\pm\)14.13) & 194.64 (\(\pm\)14.13) & 4,485.63 (\(\pm\)447.10) & 2,540.12 (\(\pm\)403.33) & 1,945.51 (\(\pm\)146.74) \\
\multicolumn{2}{r}{\textcolor{black!70}{[min--max]}} & \textcolor{black!70}{[167.25--252.35]} & \textcolor{black!70}{[167.25--252.35]} & \textcolor{black!70}{[3,823.00--8,242.00]} & \textcolor{black!70}{[2,117.00--6,330.00]} & \textcolor{black!70}{[1,670.00--2,555.00]} \\
\cmidrule(lr){1-7}
Qwen3 (80B) & All & 289.58 (\(\pm\)13.70) & 289.58 (\(\pm\)13.70) & 3,705.31 (\(\pm\)483.11) & 2,567.16 (\(\pm\)460.99) & 1,138.15 (\(\pm\)50.89) \\
\multicolumn{2}{r}{\textcolor{black!70}{[min--max]}} & \textcolor{black!70}{[254.40--359.69]} & \textcolor{black!70}{[254.40--359.69]} & \textcolor{black!70}{[3,087.00--7,550.00]} & \textcolor{black!70}{[2,066.00--6,295.00]} & \textcolor{black!70}{[1,004.00--1,361.00]} \\
\cmidrule(lr){1-7}
GPT OSS (20B) & Ind & 343.64 (\(\pm\)81.03) & 15.62 (\(\pm\)15.37) & 1,316.56 (\(\pm\)561.36) & 955.08 (\(\pm\)402.72) & 361.47 (\(\pm\)363.39) \\
\multicolumn{2}{r}{\textcolor{black!70}{[min--max]}} & \textcolor{black!70}{[240.61--792.63]} & \textcolor{black!70}{[4.91--344.67]} & \textcolor{black!70}{[687.00--9,596.00]} & \textcolor{black!70}{[505.00--4,779.00]} & \textcolor{black!70}{[108.00--8,091.00]} \\
\cmidrule(lr){1-7}
Qwen3 (80B) & Ind & 485.12 (\(\pm\)25.59) & 22.05 (\(\pm\)2.33) & 970.92 (\(\pm\)407.79) & 902.12 (\(\pm\)405.52) & 68.80 (\(\pm\)9.00) \\
\multicolumn{2}{r}{\textcolor{black!70}{[min--max]}} & \textcolor{black!70}{[421.38--576.16]} & \textcolor{black!70}{[15.73--40.33]} & \textcolor{black!70}{[516.00--4,822.00]} & \textcolor{black!70}{[449.00--4,740.00]} & \textcolor{black!70}{[45.00--145.00]} \\
\cmidrule(lr){1-7}
GPT OSS (120B) & Ind & 512.82 (\(\pm\)88.18) & 23.31 (\(\pm\)9.87) & 1,171.90 (\(\pm\)420.31) & 955.08 (\(\pm\)402.72) & 216.81 (\(\pm\)96.63) \\
\multicolumn{2}{r}{\textcolor{black!70}{[min--max]}} & \textcolor{black!70}{[375.80--1048.64]} & \textcolor{black!70}{[8.89--465.17]} & \textcolor{black!70}{[634.00--5,883.00]} & \textcolor{black!70}{[505.00--4,779.00]} & \textcolor{black!70}{[85.00--4,827.00]} \\
\bottomrule
\end{tabular}
\caption{Runtime and token usage for LLM scoring. Entries report mean (\(\pm\) SD) and a second row with [min--max]. Mode: All = combined prompting; Ind = single-criterion prompting. Token statistics are reported per prompt (total, input, output).}
\label{tab:llm-runtime-tokens-review}
\end{table*}

\section{Behaviour Data: Interaction Logs and Semester Dynamics}
\label{appendix:behavior}
To complement the textual artefacts in \nameDataset, we also record fine-grained \emph{behavioural interaction logs} from the review platform used in the course. 

\paragraph{Temporal dynamics across the semester}
Figure~\ref{fig:appendix:behaviour:daily_activity} shows daily action counts for students and instructors. Two qualitative patterns stand out.

First, student activity is strongly deadline-driven: actions rise sharply in the days immediately preceding the review deadline (red dashed line). In contrast, instructor activity is comparatively smoother and distributed across the semester, consistent with instructional work that is scheduled and paced over time rather than concentrated in the final days.

Second, activity drops to near-zero during the winter break (hatched region), including the Christmas/New Year period, and then resumes afterward.

\paragraph{Implications for LLM-assisted assessment workflows}
The pronounced pre-deadline student spike (Figure~\ref{fig:appendix:behaviour:daily_activity}) suggests that any LLM-based support integrated into the reviewing/assessment pipeline (e.g., rubric guidance, feedback scaffolding, or automated checks) will face \emph{highly non-uniform demand}. Compute budgets and latency constraints should be evaluated under peak load conditions rather than on average-day usage. 

\paragraph{Limitations}
These behavioural patterns require \emph{particular} caution because they reflect only \emph{partial observation}. Not all students consented to behavioural logging, so the curves describe only the consenting subset and may not generalize to the full cohort. The instructor group is also small ($n{=}9$), limiting statistical power and making aggregates sensitive to individual work styles. Moreover, instructor activity is not directly comparable to student activity because instructors carried a substantially higher workload: while students completed only two reviews without grading, instructors collectively reviewed across the cohort and additionally assigned criterion-based grades. Finally, logs capture only on-platform actions; substantial reviewing work may occur offline (e.g., reading drafts externally or drafting feedback elsewhere). Thus, peaks and troughs should be interpreted as \emph{platform activity} rather than complete measures of effort or time-on-task.

\end{document}